\documentclass[12pt]{article}

\usepackage[T1]{fontenc}
\usepackage{newtxtext,newtxmath}
\renewcommand{\textsubscript}[1]{\raisebox{-0.4ex}{\scriptsize #1}}
\usepackage[letterpaper,margin=1in]{geometry}
\usepackage{setspace}
\usepackage{graphicx}
\usepackage{float}
\usepackage{xcolor}

\usepackage{booktabs}
\usepackage{longtable}
\usepackage{multirow}
\usepackage{array}
\usepackage{calc}
\usepackage[normalem]{ulem}
\usepackage{caption}
\usepackage{pdflscape}
\usepackage[natbibapa]{apacite}
\makeatletter
\AtBeginDocument{\NAT@longnamesfalse}
\makeatother
\AtBeginDocument{\renewcommand{\doi}[1]{\url{https://doi.org/#1}}}
\usepackage{url}
\usepackage{newunicodechar}
\usepackage{eso-pic}
\usepackage[hidelinks]{hyperref}
\AtBeginDocument{\urlstyle{same}}
\usepackage{placeins}
\usepackage{etoolbox}
\AtBeginEnvironment{longtable}{\singlespacing}
\usepackage{titlesec}
\titleformat{\section}{\normalfont\fontsize{14}{16}\selectfont\bfseries\raggedright}{\thesection.}{0.5em}{}
\titleformat{\subsection}{\normalfont\fontsize{12}{14}\selectfont\bfseries\raggedright}{\thesubsection.}{0.5em}{}
\titleformat{\subsubsection}[runin]{\normalfont\fontsize{12}{14}\selectfont\bfseries}{\thesubsubsection.}{0.5em}{}
\titlespacing*{\subsubsection}{0pt}{1.6ex plus .4ex minus .2ex}{0.6em}

\newunicodechar{α}{\ensuremath{\alpha}}
\newunicodechar{β}{\ensuremath{\beta}}
\newunicodechar{γ}{\ensuremath{\gamma}}
\newunicodechar{η}{\ensuremath{\eta}}
\newunicodechar{λ}{\ensuremath{\lambda}}
\newunicodechar{μ}{\ensuremath{\mu}}
\newunicodechar{σ}{\ensuremath{\sigma}}
\newunicodechar{χ}{\ensuremath{\chi}}
\newunicodechar{²}{\textsuperscript{2}}
\newunicodechar{×}{\ensuremath{\times}}
\newunicodechar{→}{\ensuremath{\rightarrow}}
\newunicodechar{≈}{\ensuremath{\approx}}
\newunicodechar{≤}{\ensuremath{\leq}}
\newunicodechar{≥}{\ensuremath{\geq}}
\newunicodechar{−}{\ensuremath{-}}
\newunicodechar{ï}{\"{\i}}
\newunicodechar{†}{\textsuperscript{\textdagger}}

\graphicspath{{figures/}}
\hypersetup{
  pdftitle={Seeing Is Not Perceiving: When Synthetic Consumers Can and Cannot Pretest Visual Marketing},
  pdfauthor={Yi-Lin Tsai; Yung-Hsiu (Arvin) Lai}}

\begin{document}
\singlespacing

\renewcommand{\thefootnote}{\fnsymbol{footnote}}

\begin{center}
  {\Large\bfseries Seeing Is Not Perceiving: \\ When Synthetic Consumers Can and Cannot Pretest Visual Marketing \par}
  \vspace{2.5ex}
  {\large Yi-Lin Tsai\footnotemark[1] \qquad Yung-Hsiu (Arvin) Lai\footnotemark[1] \par}
  \vspace{1.2ex}
  {\normalsize Department of Management and Marketing, The University of Melbourne \par}
  \vspace{0.8ex}
  {\small \href{mailto:yilin.tsai@unimelb.edu.au}{yilin.tsai@unimelb.edu.au} \quad
          \href{mailto:yunghsiu.lai@unimelb.edu.au}{yunghsiu.lai@unimelb.edu.au} \par}
\end{center}

\footnotetext[1]{The authors contributed equally to this work.}

\renewcommand{\thefootnote}{\arabic{footnote}}

\vspace{2ex}

{\centering\textbf{Abstract}\par}\smallskip
\noindent Marketers now deploy generative AI agents as synthetic consumers to pretest visual assets such as logos, packaging, and advertising at a fraction of human-panel cost. However, this procedure assumes that a model seeing a visual cue can also perceive its consumer meaning, which is largely untested. We stress-test the assumption using six canonical visual marketing experiments, varying the two levers managers control: model generation (GPT-4o-mini vs. GPT-5.4-mini) and input format (plain text vs. JSON). Every resulting configuration passed the manipulation checks; however, none of the configurations reproduced more than two of the six human effects, and the remainder were nonsignificant. The one exception was a significant reversal of the human pattern. Providing conceptual or empirical evidence through in-context learning steers average responses toward the human effect. Yet steering has a limit: even when it succeeds, a configuration reproduces less than half of the natural spread of human responses and so understates consumer heterogeneity. We integrate these results into an AI governance protocol (Calibrate, Intervene, Deploy) that delineates when synthetic consumers can responsibly screen creatives and when human panels remain necessary.


\newpage

\section{Introduction}\label{introduction}
Marketing persuades through the eye. Virtually every advertisement contains visual stimuli, and much of marketing communication works through visuals alone: a logo's shape, a package's color, a product image. The pictorial element of an ad captures attention independent of its size \citep{pieters2004}, and the mere presence of an image lifts consumer engagement with brand posts (Li \& Xie, \citeyear{li2020}). Generative AI has lowered the cost of creating these assets: 37\% of marketers already use generative tools \citep{thormundsson2024}, and the output increasingly rivals professional creative work \citep{hartmann2025}. The binding constraint has therefore shifted downstream, from creating variants to evaluating them. For instance, which of fifty logo drafts will consumers read as authentic or exciting? Getting that read wrong is costly. The same visual choice can raise engagement for one brand and depress it for another, which is why recent work urges firms to pretest visual assets before deployment \citep{adam2026}. Nowadays vendors also sell an answer: synthetic consumers. These AI agents, prompted to respond to a stimulus as a human respondent would, promise creative pretesting at a fraction of the cost and time of a human panel \citep[e.g.,][]{toluna2025}.

On text-based tasks, the record appears to justify the promise. Large language models (LLMs) have reproduced human behavior in economic games and market scenarios \citep{horton2026}, matched the response patterns of opinion polls \citep{argyle2023}, and generated willingness-to-pay estimates close to human benchmarks \citep{brand2023}. In the strongest demonstration to date, an LLM predicted the outcomes of 70 preregistered survey experiments about as accurately as pooled human forecasters \citep{ashokkumar2026}. This practice has a name, silicon sampling: prompting an LLM to answer as a specified human respondent would \citep{sarstedt2024}. A manager who has watched these models predicting human survey data may reasonably assume that evaluating a logo or ad copy is a short step away.

However, this extrapolation to the visual domain, where creative budgets are primarily allocated, remains largely untested, even as generative AI spreads through the marketing-research toolkit faster than its validity is established \citep{joerling2026}. A visual cue influences consumer judgment through implicit semantic associations rather than the explicit reasoning that text benchmarks reward. We therefore distinguish seeing, the literal extraction of an image's physical features, from perceiving, the mapping of those features onto latent consumer meaning. Consumers implicitly associate faded, less-saturated colors with temporally distant events (Lee et al., \citeyear{lee2017}); a model can report an image's saturation level exactly and still miss what the fading signals about time. Our data show that the gap is not hypothetical. With no intervention, a widely used commercial model judged a product bearing an angular logo significantly more comfortable than the same product with a circular logo, reversing a canonical finding in visual marketing \citep{jiang2016}. A manager who trusted that pretest would select the wrong design. Deployed without validation, a synthetic consumer is not a shortcut. Instead, it becomes a governance risk \citep{sarstedt2024,gao2025}.

A reversal of this kind changes the question a firm faces. Before trusting a synthetic consumer with a visual asset, a manager needs to know where the agent's judgment fails, whether it can be repaired, and what the repaired agent still cannot deliver. We therefore ask: (1) Where and why does a synthetic consumer's judgment of a visual asset diverge from the human benchmark, even when it sees the cue? (2) Can evidence from the human benchmark steer a misaligned synthetic consumer back toward it, and does success depend on how the agent is configured? and (3) When steering succeeds, does the agent reproduce the full distribution of human responses?

To answer these questions, we stress-test synthetic consumers in six canonical visual marketing experiments, spanning logo descriptiveness, shape, symmetry, and stability as well as color saturation in the image, with each panel matched to the original study's human sample (69 to 220 respondents per study; more than 800 synthetic respondents per configuration). All six benchmark experiments were preregistered. Each runs in every cell of a two (model generation: GPT-4o-mini vs. GPT-5.4-mini) by two (input format: plain text vs. JSON) design, the two levers a marketer controls in practice. To test steering, we then supply in-context evidence in two doses: conceptual evidence states the original study's theory, and empirical evidence adds its statistics, the upper bound of steering a firm could attempt. Each synthetic respondent returns a numerical rating and a textual rationale. Topic-modeling the rationales allows us to separate seeing from perceiving.

This research makes three contributions. First, in response to calls to validate generative AI outputs before marketers rely on them \citep{sarstedt2024,grewal2025}, we extend silicon sampling to the visual domain and show that synthetic consumers see reliably but perceive inconsistently. All four configurations passed every visual manipulation check, yet none reproduced more than two of the six benchmark effects, and the rationales trace the failures to missing or competing semantic associations. Unlike in the text domain, this is a gap to which marketers should be alert.

Second, responding to calls to evaluate silicon samples on distributions rather than averages alone \citep{bisbee2024,peng2025}, we document where steering succeeds and where it stops. Evidence from the human benchmark pulls average responses back toward the human effect, far more effectively for the newer model generation. However, recovering consumer heterogeneity is a different matter: no configuration reproduces even half of the natural spread of human responses, and varying persona demographics does not restore it. Calibrated synthetic consumers therefore suit average-effect questions and remain unreliable for segmentation or targeting.

Third, we integrate the diagnosis and this boundary into a three-step governance protocol (Calibrate, Intervene, Deploy), answering calls for control mechanisms in responsible generative AI use \citep{arora2025,blanchard2025}. The protocol delineates the decisions a calibrated agent can support; however, only a human panel can reveal how consumers will differ.

\section{\texorpdfstring{Theoretical background}{Theoretical background}}

\subsection{\texorpdfstring{Synthetic consumers as simulators of human behavior}{Synthetic consumers as simulators of human behavior}}\label{llms-as-simulators-of-human-behavior}

Research on silicon sampling answers recent calls to validate generative AI outputs \citep{grewal2025} by testing whether LLMs can serve as proxies for human subjects. However, as summarized in \hyperref[tab:prior-studies]{Table~\ref*{tab:prior-studies}}, the extent to which these silicon samples can faithfully simulate human behavior remains contested \citep{blythe2025}. \citet{gao2025} show that models diverge from human participants in economic games and that their answers are sensitive to the wording of the prompt. More importantly, previous replications have relied predominantly on unimodal, textual inputs \citep{argyle2023,brand2023,bisbee2024,horton2026}. The closest exception is \citet{yoo2025}, which provides a field guide that replicates the research stages of 35 published articles, including image stimuli from original studies. Their goal, however, is informal demonstration rather than a test of how models perceive visual cues. Moreover, few of these studies test whether human-like outcomes rest on the semantic associations that underlie the human pathways. \citet{goli2024} and \citet{arora2025} come closest by pairing numerical outcomes with topic models of the language models' text. Our research extends this paradigm to the visual domain and pairs mediation analysis with topic modeling, allowing us to ask not only whether a model reproduces a human outcome but also whether it reproduces the semantic association underlying the human pathway.

{\footnotesize\begin{longtable}[]{@{}
  >{\raggedright\arraybackslash}p{(\linewidth - 10\tabcolsep) * \real{0.1562}}
  >{\raggedright\arraybackslash}p{(\linewidth - 10\tabcolsep) * \real{0.2684}}
  >{\centering\arraybackslash}p{(\linewidth - 10\tabcolsep) * \real{0.0844}}
  >{\centering\arraybackslash}p{(\linewidth - 10\tabcolsep) * \real{0.1138}}
  >{\centering\arraybackslash}p{(\linewidth - 10\tabcolsep) * \real{0.1564}}
  >{\centering\arraybackslash}p{(\linewidth - 10\tabcolsep) * \real{0.2208}}@{}}
\caption{\label{tab:prior-studies}Comparison of prior silicon-sample studies}\tabularnewline
\toprule\noalign{}
\begin{minipage}[b]{\linewidth}\centering
\textbf{Study}
\end{minipage} & \begin{minipage}[b]{\linewidth}\centering
\textbf{Target Domain}
\end{minipage} & \begin{minipage}[b]{\linewidth}\centering
\textbf{Input}
\end{minipage} & \begin{minipage}[b]{\linewidth}\centering
\textbf{Theory Testing}
\end{minipage} & \begin{minipage}[b]{\linewidth}\centering
\textbf{Mitigation}
\end{minipage} & \begin{minipage}[b]{\linewidth}\centering
\textbf{Mechanism Analysis}
\end{minipage} \\
\midrule\noalign{}
\endfirsthead
\toprule\noalign{}
\begin{minipage}[b]{\linewidth}\centering
\textbf{Study}
\end{minipage} & \begin{minipage}[b]{\linewidth}\centering
\textbf{Target Domain}
\end{minipage} & \begin{minipage}[b]{\linewidth}\centering
\textbf{Input}
\end{minipage} & \begin{minipage}[b]{\linewidth}\centering
\textbf{Theory Testing}
\end{minipage} & \begin{minipage}[b]{\linewidth}\centering
\textbf{Mitigation}
\end{minipage} & \begin{minipage}[b]{\linewidth}\centering
\textbf{Mechanism Analysis}
\end{minipage} \\
\midrule\noalign{}
\endhead
\bottomrule\noalign{}
\endlastfoot
\citet{argyle2023} & Political survey responses & Text & No & None &
None \\
\citet{bisbee2024} & Public opinion polling & Text & No & ICL &
None \\
\citet{brand2023} & Willingness-to-pay & Text & Yes & Fine-tuning
\textsuperscript{a} & None \\
Li et al. (\citeyear{li2024}) & Perceptual brand mapping & Text & Yes & ICL
(few-shot) & None \\
\citet{goli2024} & Intertemporal preferences & Text & Yes & CoT &
Topic Model \\
\citet{toubia2025} & Personal traits and cognition & Text & Yes & ICL
(zero-shot) & None \\
\citet{peng2025} & 19 behavioral domains & Text & Yes & None & None \\
\citet{arora2025} & Qualitative in-depth interviews & Text & No & ICL
(few-shot) + RAG & Topic Model \\
\citet{yoo2025}\textsuperscript{b} & Research-process demonstrations & Image + text & Yes & None & None \\
\citet{ashokkumar2026} & Preregistered survey experiments & Text & Yes &
None & None \\
\citet{horton2026} & Behavioral Economics, fairness, status quo & Text & Yes
& Persona calibration & None \\
This research & Perception of visual design &
Image + text & Yes & ICL (evidence-based) & Mediation analysis +

topic model \\
\end{longtable}}

{\footnotesize\setstretch{1}\raggedright\textsuperscript{a} Human-response data were used to fine-tune the model's weights.
\textsuperscript{b} A field guide for research process; image stimuli were uploaded where the replicated article used them, and the authors note their comparisons are not exact, high-powered replications.\par}

\subsection{\texorpdfstring{Technical priors versus psychological priors in visual perception}{Technical priors versus psychological priors in visual perception}}\label{the-psychology-of-visual-marketing}

Visual marketing relies heavily on implicit inferences, where subtle design elements (e.g., logo shape) can trigger latent psychological associations (e.g., perceived comfort) that extend beyond literal visual recognition \citep{labrecque2012}. The true consumer meaning of a visual cue is rarely discernible from its surface alone. A picture of a scented object improves product evaluations by evoking olfactory imagery \citep{sharma2024}, and presenting products jointly rather than separately lifts evaluations through consumption imagery and psychological ownership (Zhao \& Xia, \citeyear{zhaoxia2021}). In both cases, the effect runs through a simulation in the consumer's mind, not through anything a feature detector could read off the pixels. The same surface cue can also mean opposite things in different brand contexts \citep{adam2026,chen2025}, or move behavior only when its form matches consumers' internal representations \citep{kim2023}. Consequently, evaluating whether a synthetic consumer reproduces these associations requires distinguishing between what the model technically sees and what it actually perceives. A model trained predominantly on descriptive image--text pairs may read the geometry without the association it carries for a consumer. Existing AI frameworks in marketing already separate the processing of objective data from the response to emotion \citep{huang2021}. Consumers share a similar belief, viewing AI as strong in logic but weaker in emotional connection \citep{xu2022}. The dissociation is visible in field data: an image model trained on objective style labels predicts them near-perfectly (96.9\% accuracy) yet collapses to 62.5\% when retrained to predict human familiarity perceptions of the same images (Xie et al., \citeyear{xie2025}).

Drawing on Bayesian cognitive modeling, which formalizes human perception and learning as probabilistic updates \citep{griffiths2008}, we conceptualize the split between seeing and perceiving as two learned priors that a synthetic consumer brings to a visual asset: a technical prior and a psychological prior. We define a prior not as a conscious cognitive state or a parameter we directly estimate, but as a baseline tendency to interpret a visual cue one way rather than another. The technical prior is the baseline tendency to extract physical features, such as a logo's geometry or an image's color saturation. The psychological prior is the learned statistical association that maps those physical features onto latent consumer meaning, such as authenticity or excitement. This split mirrors how human visual perception separates an image's physical features from its semantic meaning \citep{sample2020}. Image--text training data may describe what an image shows more often than how it makes a consumer feel, so a model's technical prior can sometimes overshadow its psychological prior. A single visual cue can also support several learned associations, so a model may express a competing psychological prior that is coherent but differs from the human benchmark. A synthetic consumer may therefore pass all visual recognition tests while missing the implicit association that humans would form. Whether these two priors dissociate in practice, and how far the gap can be closed, are the empirical questions our six benchmark experiments quantify.

\subsection{\texorpdfstring{In-context evidence as a steering mechanism}{In-context evidence as a steering mechanism}}\label{bridging-the-gap-with-in-context-learning}

To better activate the target psychological prior, we use In-Context Learning (ICL), a prompt-based method that does not modify any model weights \citep{brown2020}. While prior studies show Chain-of-Thought (CoT) prompting can improve performance on explicit logical tasks \citep{goli2024,kojima2022} and theory development \citep{yoo2025}, these step-by-step rationalizations target explicit reasoning rather than the implicit associations that drive visual marketing. Similarly, while Retrieval-Augmented Generation (RAG) is robust for fact-based queries \citep{arora2025}, retrieval targets facts rather than implicit psychological associations. Methods that update the model's weights, such as fine-tuning on human responses \citep{brand2023} or structured computer-vision models \citep{dew2022}, may steer a model further, but they need training data and model access that most practitioners do not have. We therefore select ICL, treating it as a mechanism for implicit Bayesian inference (Xie et al., \citeyear{xie2022}). Embedding expert conceptual or empirical knowledge directly into the prompt acts as contextual evidence, which allows us to shift the associations the model expresses at a low computational cost. A language model's tendency to echo its prompt is what makes steering possible in the first place, but how far that steering extends when the judgment target is an image rather than text has not been tested. In practice, a marketer is unlikely to retrain a closed commercial model; she chooses which model generation to license and how to structure the stimulus input, and text-domain work shows model output is sensitive to both \citep{bisbee2024,he2024,toubia2025}. We therefore evaluate the robustness of ICL across model generations and input formats. Beyond mean accuracy, given that LLM-generated outputs consistently suffer from variance shrinkage \citep{peng2025}, we evaluate how effectively this in-context evidence can restore the natural heterogeneity of human responses.

The two priors, the steering mechanism, and the variance concern map one-to-one onto our research questions posed in Section~\ref{introduction}: perceptual misalignment despite accurate seeing (RQ1), evidence-based steering and its dependence on model generation and input format (RQ2), and recovery of the full distribution of human responses rather than the mean alone (RQ3). The answers are integrated into the governance protocol developed in Section~\ref{general-discussion}.

\section{Methodology}\label{methodology}

To compare synthetic consumers' visual judgments with established human benchmarks, we re-ran six visual marketing experiments with synthetic respondents. The visual stimuli for these experiments were drawn directly from six marketing studies. All of our benchmark experiments were preregistered before data collection (\hyperref[web-appendix-a-pre-registration-for-studies]{Web Appendix A}). For each experiment, two types of data were collected: (1) a numerical rating mirroring the original study's dependent variable, and (2) a textual rationale explaining the basis for that rating. This approach allows us to assess both the statistical alignment of the output and the semantic associations behind it.

\subsection{\texorpdfstring{Overview of the studies}{Overview of the studies}}\label{overview-of-the-studies}

To test the boundary between seeing and perceiving, we selected six canonical studies based on four criteria: (1) publication in a premier marketing journal (e.g., \emph{Journal of Consumer Research}; \emph{Journal of Marketing Research}), (2) publicly available visual stimuli, (3) a focus on perceptual variation rather than objective performance measures (e.g., measuring response time of an LLM prediction is not meaningful), and (4) an established citation record prior to 2020, so that the findings constitute recognizable benchmarks. We also prioritized studies that specified a psychological pathway, which allows us to compare an outcome with the pathway behind it. We treat the published results from these six studies as an operational benchmark for comparison with LLM outputs, not as absolute ground truth. \hyperref[tab:benchmark-studies]{Table~\ref*{tab:benchmark-studies}} summarizes these studies. Full details on hypotheses and stimuli are provided in \hyperref[web-appendix-b-description-of-studies-and-visual-stimuli]{Web Appendix B}.

\begin{table}[H]
\singlespacing\footnotesize
\caption{\label{tab:benchmark-studies}Summary of the six benchmark studies}
\setlength{\tabcolsep}{3.5pt}\renewcommand{\arraystretch}{1.2}
\begin{tabular}{@{}c >{\raggedright\arraybackslash}p{2.9cm} c c c c >{\raggedright\arraybackslash}p{4.5cm}@{}}
\toprule
\# & Paper & Journal & Stimuli Source\textsuperscript{a} & Sample Size & Citation Count\textsuperscript{b} & Human Perception \\
\midrule
1 & \citet{luffarelli2019a} & JMR & 1 & 180 & 156 & High logo descriptiveness is associated with high brand authenticity. \\[2pt]
2 & Lee et al. (\citeyear{lee2017}) & JCR & 2A & 108 & 113 & Consumers select less-saturated (more-saturated) images for temporally distant (near) events. \\[2pt]
3 & \citet{jiang2016} & JCR & 2 & 69 & 380 & A circular (angular) logo is associated with higher (lower) comfort and lower (higher) durability. \\[2pt]
4 & \citet{luffarelli2019} & JMR & 1B & 220 & 215 & Logo asymmetry is associated with enhanced excitement. \\[2pt]
5 & \citet{hagtvedt2017} & JCR & 3 & 80 & 237 & High color saturation is associated with a bigger product size. \\[2pt]
6 & \citet{rahinel2016} & JCR & 2A & 154 & 67 & An unstable logo is associated with greater utility for safety-oriented products. \\
\bottomrule
\end{tabular}

\smallskip
{\footnotesize\setstretch{1}\raggedright
\textsuperscript{a} Stimuli Source gives the study (and stimulus set) in the original article from which the stimuli were taken (e.g., 2A $=$ Study 2A). \textsuperscript{b} The numbers of citations were obtained in April 2025 from Google Scholar, before our data collection began.\par}
\end{table}

\subsection{Research design using AI participants}\label{research-design-using-ai-participants}

\begin{figure}[H]
\centering
\includegraphics[width=\textwidth]{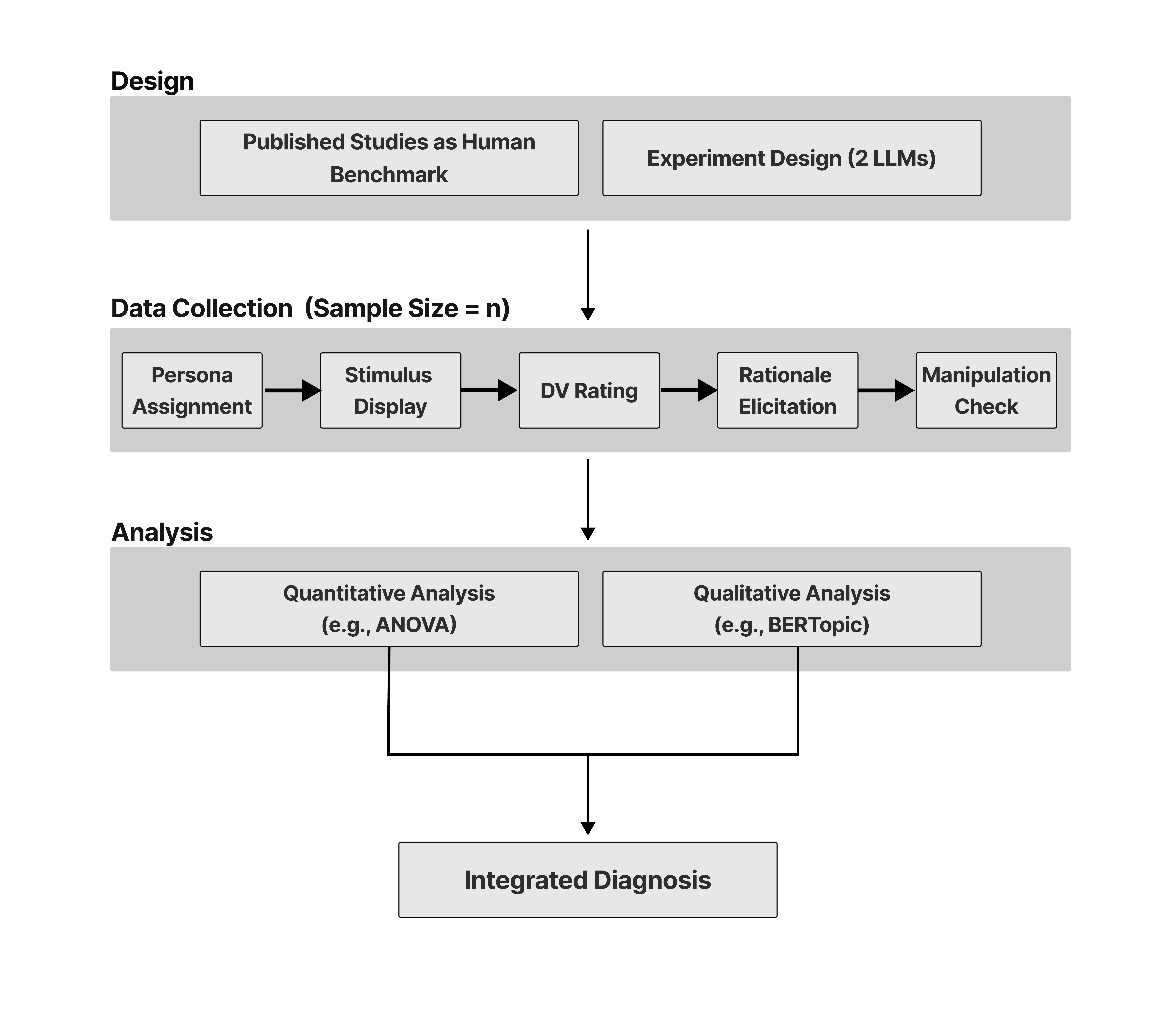}
\caption{\label{fig:research-design}Research design for the AI participants.}
\end{figure}

We used a four-stage workflow that includes Design, Data Collection, Analysis, and Diagnosis (\hyperref[fig:research-design]{Figure~\ref*{fig:research-design}}). This workflow generates the data for the zero-shot baselines that address RQ1.

\subsection{Design and data collection}\label{design-and-data-collection}

During the Design phase, we selected two GPT models to serve as synthetic consumers (hereafter, AI participants): GPT-4o-mini (released in July 2024) and the newer GPT-5.4-mini (released in March 2026), two versions of the same model line.\footnote{API model identifiers: gpt-4o-mini and gpt-5.4-mini-2026-03-17.} We also ran each model on two input formats: plain text (TXT) and JSON. This allows us to assess whether perceptual alignment varies between the two model generations or simply with structuring the prompt interface. Unless otherwise noted, the main text reports the plain-text runs of each model, GPT-4o-mini (TXT) and GPT-5.4-mini (TXT).

To ensure the validity of our synthetic-consumer sampling, we made the following configuration decisions on the parameters the API exposed at the time. First, we set the model temperature to 1.0, the API default and the midpoint of its 0 to 2 range, so that responses vary across participants \citep{arora2025}. Second, we kept nucleus sampling (top\_p) at its default value of 1. Lowering this value would cut off the model's less likely answers and artificially narrow the range of responses it can produce \citep{openai2024}. Third, we did not specify a seed parameter due to the realities of commercial APIs. Although seed-fixing is a customary practice in simulations, the API does not return the same output even when identical seeds are applied \citep{openai2024}. Finally, to support the statistical independence of our synthetic sample, we followed \citet{blanchard2025} by initiating a separate API call for every individual AI participant, rather than batch-generating multiple personas within a single context window, so no context carries over between participants.

\begin{figure}[H]
\centering
\includegraphics[width=0.9\textwidth]{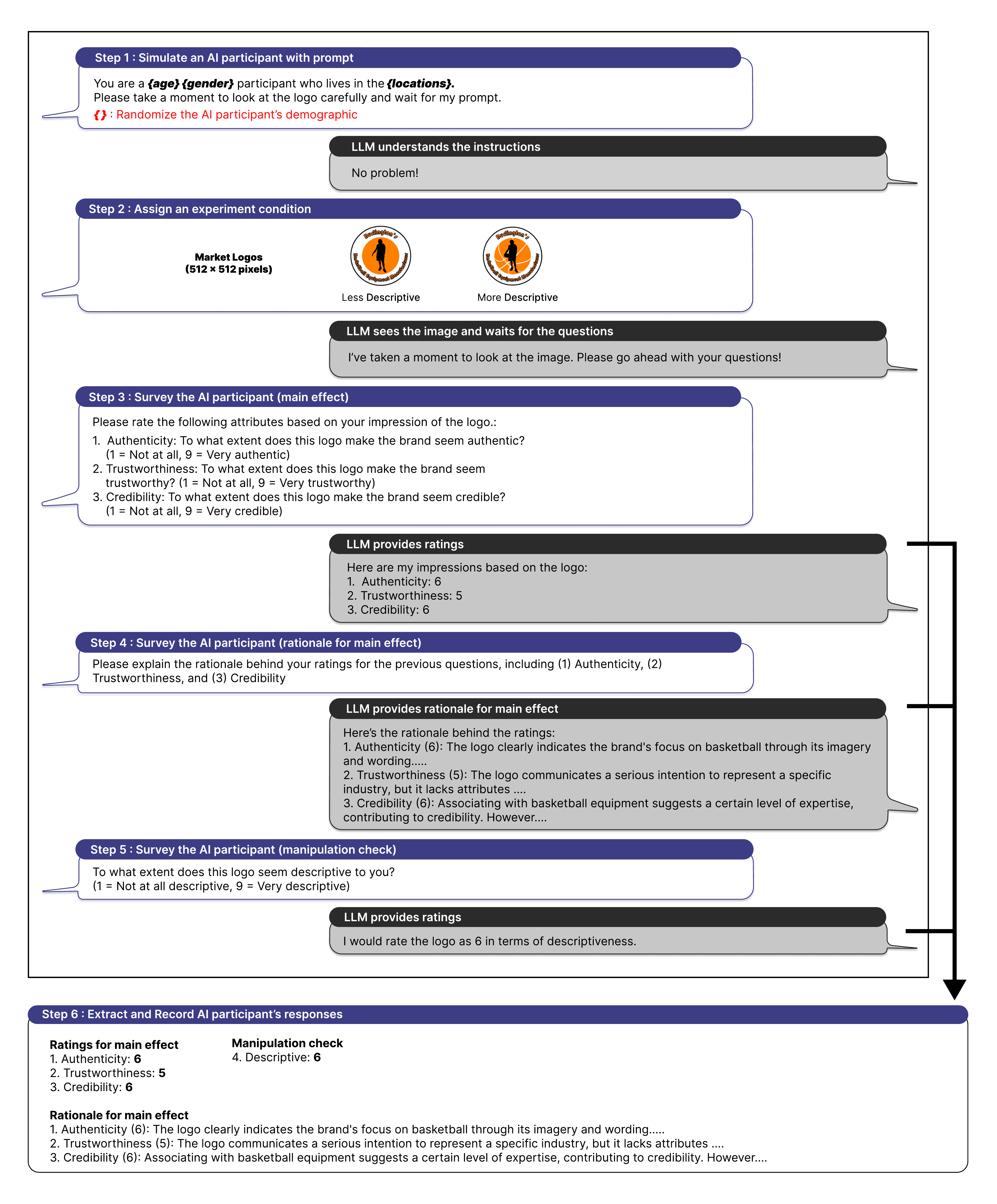}
\caption{\label{fig:api-example}An example of surveying AI participants using the visual stimuli from Study 1 in \citet{luffarelli2019a}, implemented through the API.}
\end{figure}

The Data Collection phase includes multiple conversations (i.e., API calls) to mirror human information processing. See \hyperref[fig:api-example]{Figure~\ref*{fig:api-example}} for an illustration. In Step 1, we instantiated the participant's persona by providing a demographic profile (age, gender, location) and foundational instructions. Each AI participant was assigned a demographic profile consistent with the sample used in the original study. For the studies drawn from U.S. adult samples (Studies 1, 2, and 4), age was sampled uniformly from 18 to 65, and gender was balanced between male and female. For the studies drawn from undergraduate samples (Studies 3, 5, and 6), the persona was specified as an undergraduate student with an age sampled uniformly from an undergraduate range. The demographic distributions for each study, including the range used in each, are reported in \hyperref[web-appendix-c-heterogeneity]{Web Appendix C}. In Step 2, the AI participant was randomly assigned a visual stimulus presented as a 512×512-pixel PNG file. This resolution was chosen to preserve sufficient visual detail while maintaining computational efficiency \citep{hartmann2025}. In Step 3, we prompted the AI to provide zero-shot numerical ratings for the key dependent variables from the original study. Zero-shot prompting entails instructing an LLM to perform a task without prior examples of the desired output \citep{brown2020}, so that the AI's ratings reflect the associations the model expresses without any benchmark information. This was followed by a request for a textual rationale for each rating in Step 4. Finally, we administered a numerical manipulation check to ensure the stimulus was perceived as intended (Step 5). We adopted this sequence to mitigate demand effects: inquiries about specific visual features could otherwise cue the model to the manipulated feature before the focal dependent-variable ratings.

We kept each call stateful by passing the entire prior conversation history (e.g., the assigned persona, the stimulus, and all previous answers) as input to the next step, keeping responses consistent with the assigned persona and prior ratings. As shown in \hyperref[fig:api-example]{Figure~\ref*{fig:api-example}}, the API call for the Step 4 rationale was explicitly conditioned on the context from Steps 1 to 3.
This entire process was repeated to match the original study's sample size, rounded up where an odd total could not be split evenly across conditions.

\subsection{Analysis and diagnosis}\label{analysis-and-diagnosis}

We replicated the original statistical analyses (e.g., ANOVA) to estimate the main effects using the AI-generated data. A configuration reproduces a human effect when the AI participants show an association between the independent and dependent variables that is
both statistically significant and directionally consistent with the original human subjects' findings. For studies that proposed mediators, we followed the original PROCESS models \citep{hayes2022} using the AI's numerical ratings, with the same mediator and outcome measures as the original studies.

The textual rationales (Steps 4 and 6, \hyperref[fig:api-example]{Figure~\ref*{fig:api-example}}) provide another source of diagnostic data. Our analysis serves two key purposes. First, it verifies that the models identify the target visual features (e.g., shape or color saturation). Second, the analysis captures nuances that fixed-point rating scales cannot flexibly represent \citep{zhang2024}. In instances where an AI fails to reproduce a human main effect, this diagnostic layer allows us to distinguish the sources of the misalignment. For example, we can tell if the model defaults to a literal description of image features, or if it identifies the visual cue but expresses an alternative semantic association (e.g.,
a more descriptive logo elicits ``professionalism'' instead of the intended ``ease of processing''). While this procedure mimics human think-aloud protocols \citep{ericsson1980}, LLM-generated explanations can be plausible yet systematically unfaithful to the model's actual decision process \citep{turpin2023}. We therefore treat rationales with the caution recommended for AI-generated outputs \citep{tomaino2025}, as observable artifacts rather than the model's actual thought process.

To analyze these short textual rationales (typically 200--2,000 characters), we needed an NLP pipeline that could overcome high semantic sparsity. Traditional bag-of-words models like Latent Dirichlet Allocation (LDA; \citealp{blei2003}) often perform poorly on short texts because they ignore syntax and contextual meaning (Zhao et al., \citeyear{zhao2011}). Although Bidirectional Encoder Representations from Transformers (BERT; \citealp{devlin2019}) resolves this issue, its native architecture is not optimized for measuring direct sentence-level similarity. Therefore, we encoded the documents using Sentence-BERT (SBERT; \citealp{reimers2019}) and clustered them using BERTopic \citep{grootendorst2022}. Moreover, unlike embedding-based alternatives that require researchers to pre-specify the number of topics, BERTopic uses a density-based clustering algorithm to determine the number of distinct topics. Given our per-condition sample sizes, this pipeline typically produced two to ten topics per condition (technical details are available in \hyperref[web-appendix-d-bertopic-and-llm-assisted-topic-models]{Web Appendix D}).

\section{Results}\label{results-and-analysis}

\hyperref[tab:zeroshot-main]{Table~\ref*{tab:zeroshot-main}} summarizes the zero-shot results of six studies for the plain-text configurations; results for all four configurations appear in \hyperref[tab:intervention-main]{Table~\ref*{tab:intervention-main}} and \hyperref[web-appendix-e-prompt-intervention-templates-and-results-after-interventions]{Web Appendix E}. First, at zero-shot, all configurations passed the manipulation checks (e.g., correctly identifying circular logos). Second, we find that none of the four configurations successfully reproduced more than two of the six canonical effects. Both GPT-4o-mini configurations reproduced Study 1's effect, and both GPT-5.4-mini configurations reproduced the same two: Studies 3 and 5. Notably, 4o-mini (TXT) also generated a statistically significant directional reversal of a human finding in Study 3. Predicting the opposite of human judgment poses a greater operational risk than a null effect, as the synthetic agent outputs can lead to a flawed managerial recommendation. By synthesizing the textual rationales across these failures, we identify two sources of AI perceptual misalignment. The first is that LLMs successfully recognize visual features but restrict their evaluation to objective aesthetic properties without triggering the downstream semantic associations that carry the human effect (e.g., Study 2). The second is that LLMs recognize the visual cue but activate a competing psychological prior different from the human benchmark (e.g., Studies 1 and 3). This zero-shot performance thus answers RQ1. A model that sees a marketing-relevant visual cue does not reliably perceive its consumer meaning as humans do.

\begin{table}[H]
\singlespacing\footnotesize
\caption{\label{tab:zeroshot-main}Zero-shot main effects and manipulation checks by model generation (plain-text configurations)}
\setlength{\tabcolsep}{6pt}\renewcommand{\arraystretch}{1.2}
\centering
\begin{tabular}{@{}lcccc@{}}
\toprule
& \multicolumn{2}{c}{4o-mini (TXT)} & \multicolumn{2}{c}{5.4-mini (TXT)} \\
\cmidrule(lr){2-3}\cmidrule(lr){4-5}
Study & Main effect & Manipulation\ check & Main effect & Manipulation\ check \\
\midrule
1 (Authenticity) & \textbf{$+$ ($<$.001)} & \textbf{$+$ ($<$.001)} & $-$ (.586) & \textbf{$+$ ($<$.001)} \\
2 (Selected saturation) & $+$ (.631) & \textbf{$+$ ($<$.001)} & $+$ (.288) & \textbf{$+$ ($<$.001)} \\
3 (Comfort) & \textbf{$-$ (.012)} & \textbf{$+$ ($<$.001)} & \textbf{$+$ (.022)} & \textbf{$+$ ($<$.001)} \\
3 (Durability) & $-$ (.177) & \textbf{$+$ ($<$.001)} & $-$ (.321) & \textbf{$+$ ($<$.001)} \\
4 (Excitement) & $-$ (.736) & \textbf{$+$ ($<$.001)} & $-$ (.383) & \textbf{$+$ ($<$.001)} \\
5 (Size estimation) & $+$ (.052) & \textbf{$+$ ($<$.001)} & \textbf{$+$ (.010)} & \textbf{$+$ ($<$.001)} \\
6 (Utility) & $-$ (.136) & \textbf{$+$ (.004)} & $-$ (.844) & \textbf{$+$ (.017)} \\
\bottomrule
\end{tabular}

\smallskip
{\footnotesize\setstretch{1}\raggedright\emph{Note}: Parentheses after study numbers give the dependent variable. ``$+$/$-$'' indicates a direction consistent (inconsistent) with the human finding; for manipulation checks, $+$ indicates the manipulated feature was rated in the intended direction (e.g., angular logos rated as more angular). Bold marks significant effects ($p<.05$). Study 6's main effect is driven by its simple effect within safety-oriented products. JSON results were statistically indistinguishable at zero-shot; complete results for all four configurations appear in \hyperref[tab:intervention-main]{Table~\ref*{tab:intervention-main}} and \hyperref[web-appendix-e-prompt-intervention-templates-and-results-after-interventions]{Web Appendix E}.\par}
\end{table}

One concern is that reproduction success might simply reflect the model's prior exposure to well-known studies.\footnote{GPT-4o-mini training cutoff: October 2023; GPT-5.4-mini training cutoff: August 2025.} If the models were simply recalling these studies, the most-cited and oldest effects should reproduce best, and none should reverse. We observe the opposite. Most studies fail at zero-shot and one reverses (Study 3). An exploratory logistic regression predicting reproduction success from citation count, paper age, sample size, and model likewise finds no significant predictors, though with only six studies this check has limited power. Given the small differences in zero-shot results across input formats within each model, we report the findings on the plain-text configurations. For parsimony, we abbreviate these runs as 4o-mini and 5.4-mini.

\subsection{Detailed findings by study}

\FloatBarrier
\subsubsection{Study 1: Logo descriptiveness and brand authenticity}\label{study-1-logo-descriptiveness-and-brand-authenticity}

\begin{figure}[H]
\centering
\includegraphics[width=6in,height=3in]{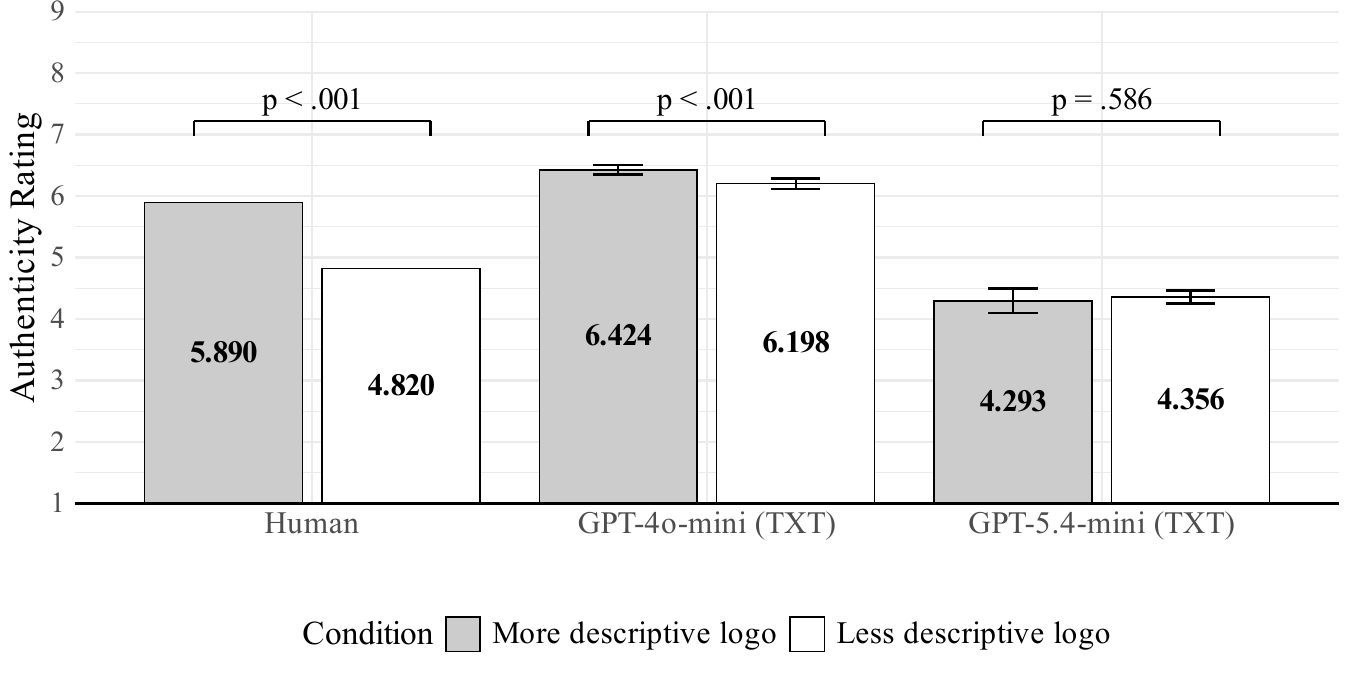}
\caption{\label{fig:s1-authenticity}Authenticity rating by logo descriptiveness.}
\par\smallskip{\footnotesize\setstretch{1}\raggedright \emph{Note}: Bars show zero-shot (no-intervention) condition means for the human benchmark, 4o-mini, and 5.4-mini. Error bars represent the 95\% confidence intervals. Error bars for the human sample are omitted because the original study does not report the corresponding standard deviations.\par}
\end{figure}

Humans perceive highly descriptive logos as more authentic in \citet{luffarelli2019a}. 4o-mini successfully reproduced this main effect (M\textsubscript{Less\_Desc} {[}SD\textsubscript{Less\_Desc}{]} = 6.198 {[}.405{]} vs. M\textsubscript{More\_Desc} = 6.424 {[}.354{]}, F(1, 167) = 14.828,
\emph{p \textless{}} .001, η² = .082; see \hyperref[fig:s1-authenticity]{Figure~\ref*{fig:s1-authenticity}}), with an effect size close to the human benchmark (η² = .091). However, while the original study found that ease of processing mediated this relationship, this indirect effect was not significant for 4o-mini (β = $-$.053, 95\% CI {[}$-$.134, .016{]}). Logo descriptiveness did not significantly predict ease of processing. Post-hoc textual analysis revealed that 4o-mini linked its authenticity ratings to ``professionalism,'' emphasizing polished design, rather than to ease of processing. This is an example of triggering an alternative psychological prior, despite the same outcome as the human benchmark.

5.4-mini, in contrast, showed no significant main effect (M\textsubscript{Less\_Desc} {[}SD\textsubscript{Less\_Desc}{]} = 4.356 {[}.507{]} vs. M\textsubscript{More\_Desc} = 4.293 {[}.970{]}, F(1, 178) = .298, \emph{p =} .586, η² = .002).  The post-hoc analysis shows that 5.4-mini focused on general design quality rather than logo descriptiveness, which explains the null result.

\FloatBarrier
\subsubsection{Study 2: Color saturation and temporal distance}\label{study-2-color-saturation-and-temporal-distance}

\begin{figure}[H]
\centering
\includegraphics[width=6.42574in,height=3.21287in]{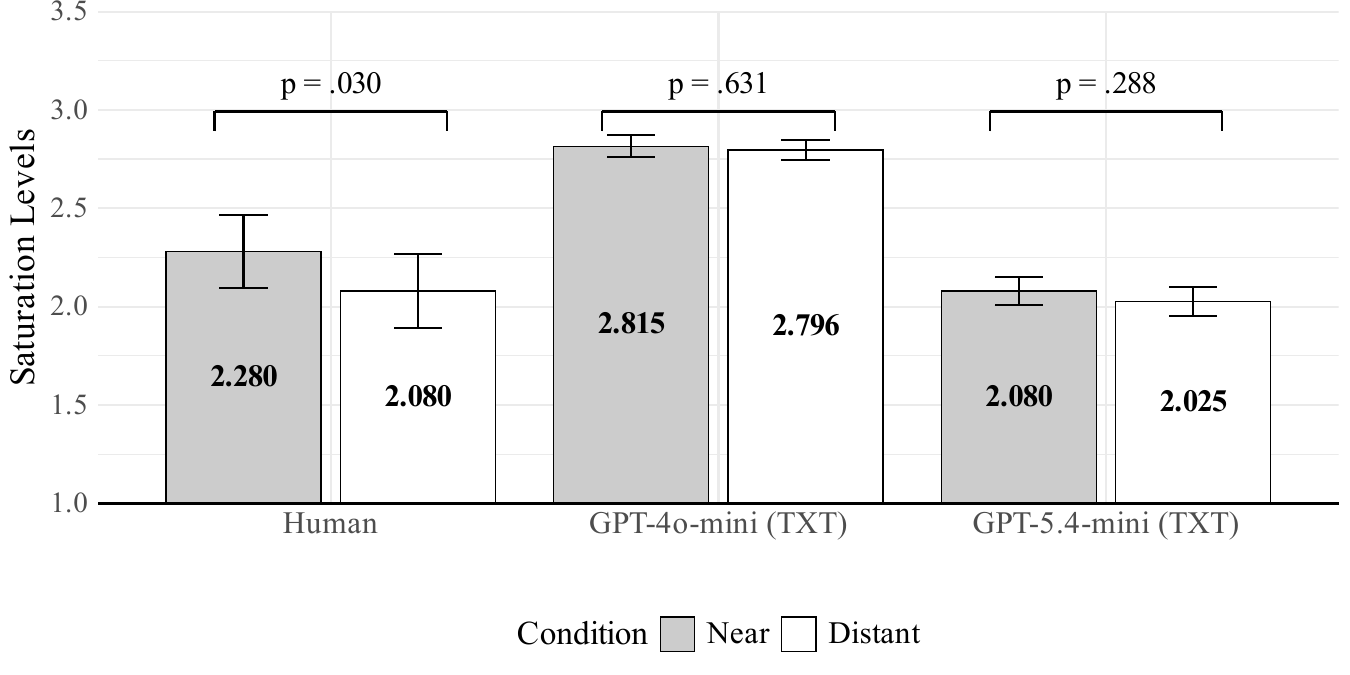}
\caption{\label{fig:s2-saturation}Selected saturation by temporal distance.}
\par\smallskip{\footnotesize\setstretch{1}\raggedright \emph{Note}: Bars show zero-shot (no-intervention) condition means for the human benchmark, 4o-mini, and 5.4-mini. Error bars represent the 95\% confidence intervals.\par}
\end{figure}

Humans associate less saturated images with distant future events (Lee et al., \citeyear{lee2017}). However, neither 4o-mini nor 5.4-mini reproduced this association. Neither model showed a significant main effect (4o-mini: M\textsubscript{Near} {[}SD\textsubscript{Near}{]} = 2.815 {[}.211{]} vs. M\textsubscript{Distant} = 2.796 {[}.188{]}; F(1, 106) = .231, \emph{p =} .631, η² = .002. 5.4-mini: M\textsubscript{Near}
{[}SD\textsubscript{Near}{]} = 2.080 {[}.266{]} vs. M\textsubscript{Distant} = 2.025 {[}.274{]}; F(1, 106) = 1.143, \emph{p =} .288, η² = .011; \hyperref[fig:s2-saturation]{Figure~\ref*{fig:s2-saturation}}), despite passing the manipulation checks (\hyperref[tab:zeroshot-main]{Table~\ref*{tab:zeroshot-main}}; the check measures are described in \hyperref[web-appendix-b-description-of-studies-and-visual-stimuli]{Web Appendix B}). The results suggest that while LLMs could recognize the visual difference and understand the temporal prompts, they registered saturation as an aesthetic property but did not connect it to temporal distance, the association that carries the human effect.

\FloatBarrier
\subsubsection{\texorpdfstring{Study 3: Logo shape and product attributes}{Study 3: Logo shape and product attributes}}\label{study-3-logo-shape-and-perceived-comfortableness}

\begin{figure}[H]
\centering
\includegraphics[width=6in,height=3in]{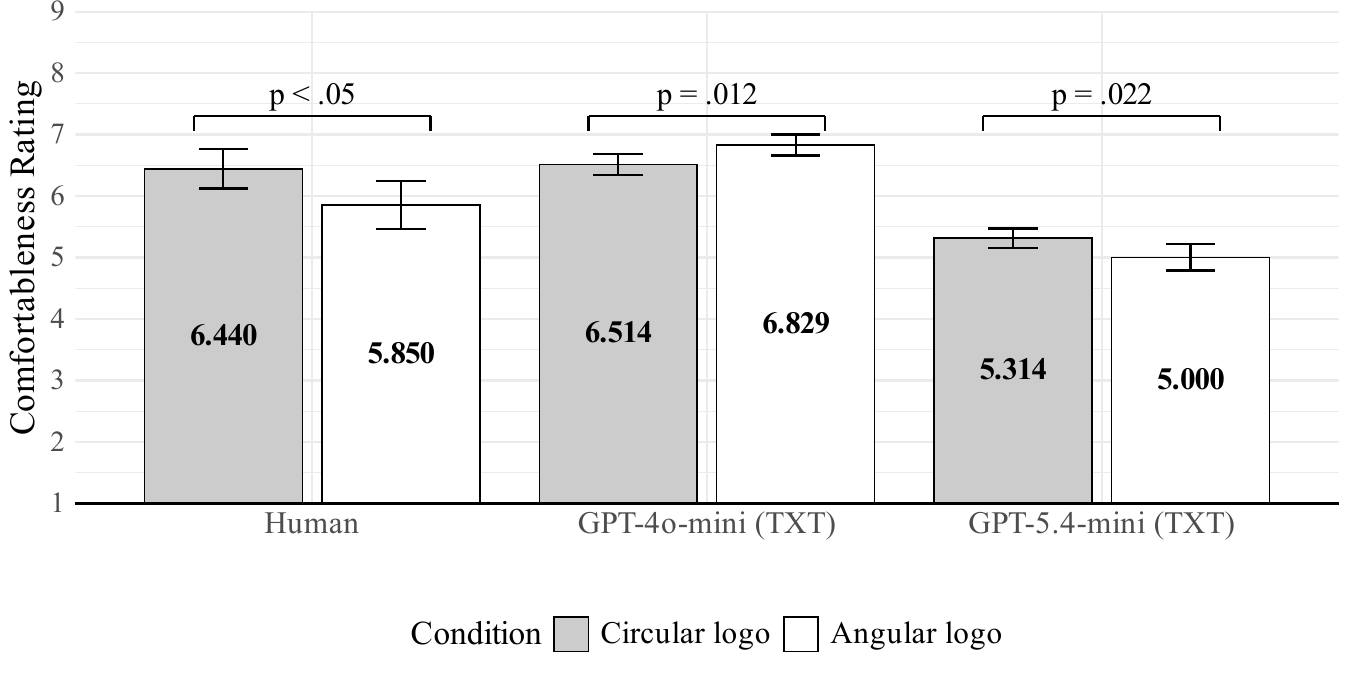}
\caption{\label{fig:s3-comfort}Perceived comfort by logo shape.}
\par\smallskip{\footnotesize\setstretch{1}\raggedright \emph{Note}: Bars show zero-shot (no-intervention) condition means for the human benchmark, 4o-mini, and 5.4-mini. Error bars represent the 95\% confidence intervals.\par}
\end{figure}

\begin{figure}[H]
\centering
\includegraphics[width=6in,height=3in]{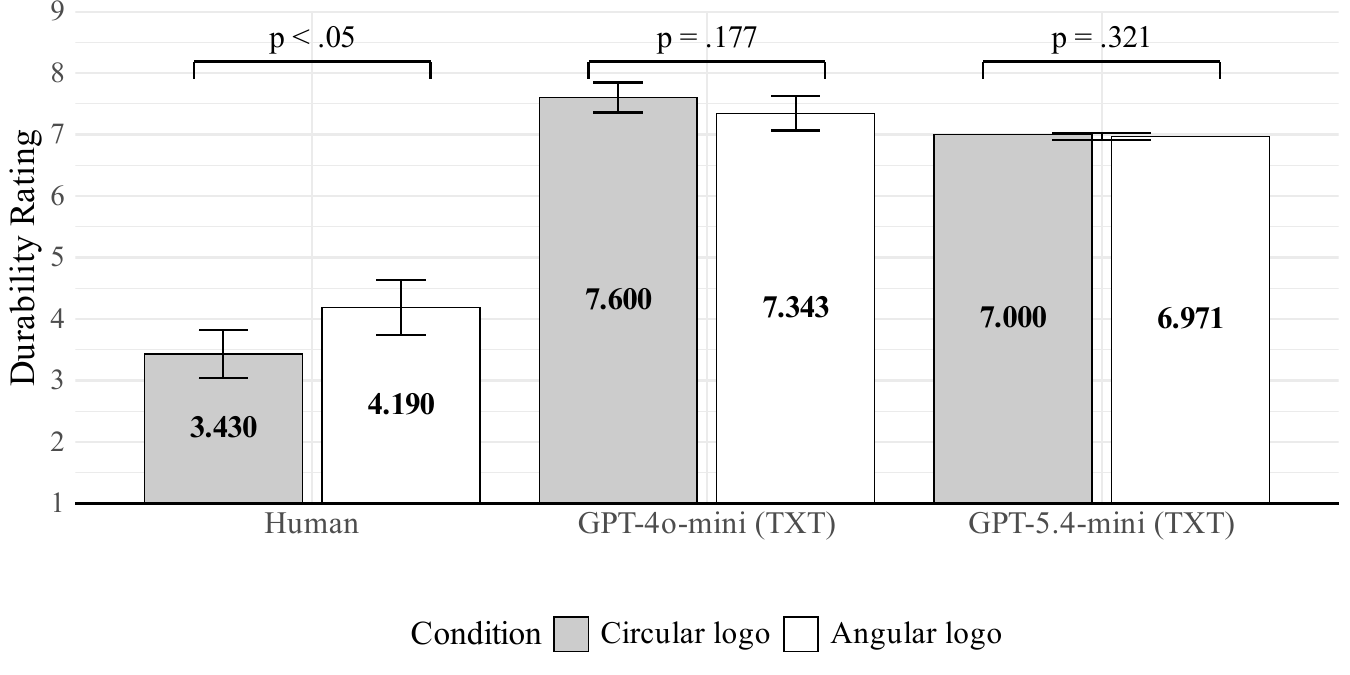}
\caption{\label{fig:s3-durability}Perceived durability by logo shape.}
\par\smallskip{\footnotesize\setstretch{1}\raggedright \emph{Note}: Bars show zero-shot (no-intervention) condition means for the human benchmark, 4o-mini, and 5.4-mini. Error bars represent the 95\% confidence intervals.\par}
\end{figure}

Study 3 provides the sharpest contrast between the two model generations. We used logo shape (circular vs. angular) as the independent variable to assess the impact on perceived comfort and durability, as in \citet{jiang2016}. 4o-mini rated products with angular logos more comfortable than those with circular logos, a reversal of the human findings (M\textsubscript{angular} {[}SD\textsubscript{angular}{]} = 6.829 {[}.514{]} vs. M\textsubscript{circular} = 6.514 {[}.507{]}, F(1, 68) = 6.635, \emph{p} = .012, η² = .089; \hyperref[fig:s3-comfort]{Figure~\ref*{fig:s3-comfort}}). Additionally, 4o-mini associated circular logos with higher perceived durability. This estimate, though not statistically significant, was directionally opposite to the human data (M\textsubscript{angular} {[}SD\textsubscript{angular}{]} = 7.343 {[}.838{]} vs. M\textsubscript{circular} = 7.600 {[}.736{]}; F(1, 68) = 1.861, \emph{p} = .177, η² = .027; \hyperref[fig:s3-durability]{Figure~\ref*{fig:s3-durability}}). In the post-hoc text analysis, 4o-mini provided an alternative explanation by associating an angular logo with ``better support'' and, therefore, a ``more comfortable'' sofa.

5.4-mini, in contrast, reproduced the human association on comfort, rating products with circular logos as more comfortable than those with angular logos (M\textsubscript{circular} {[}SD\textsubscript{circular}{]} = 5.314 {[}.471{]} vs. M\textsubscript{angular} = 5.000 {[}.642{]}, F(1, 68) = 5.456, \emph{p} = .022, η² = .074; \hyperref[fig:s3-comfort]{Figure~\ref*{fig:s3-comfort}}). The effect size is slightly larger than the human benchmark (human benchmark: η² = .071). However, 5.4-mini failed to reproduce the durability effect, showing no significant difference between shapes (M\textsubscript{circular} {[}SD\textsubscript{circular}{]} = 7.000 {[}.000{]} vs. M\textsubscript{angular} = 6.971 {[}.169{]}; F(1, 68) = 1.000, \emph{p} = .321, η² = .014). The circular cell shows zero variance, so we read this test only descriptively. We further find that the semantic association underlying the pathway showed the same split: For 4o-mini, logo shape did not predict perceived softness, whereas for 5.4-mini, the indirect effect through softness was significant for comfort, as in the human benchmark (\hyperref[e-study-3]{Table E3}, \hyperref[web-appendix-e-prompt-intervention-templates-and-results-after-interventions]{Web Appendix E}). In the rationales, 5.4-mini judged the sofas on ``softness'' and ``comfort,'' the benchmark mediator, whereas 4o-mini explained comfort through ``support.'' This difference matches their replication outcomes.

\FloatBarrier
\subsubsection{Study 4: Logo asymmetry and perceived excitement}\label{study-4-logo-asymmetry-and-perceived-excitement}

\begin{figure}[H]
\centering
\includegraphics[width=\linewidth]{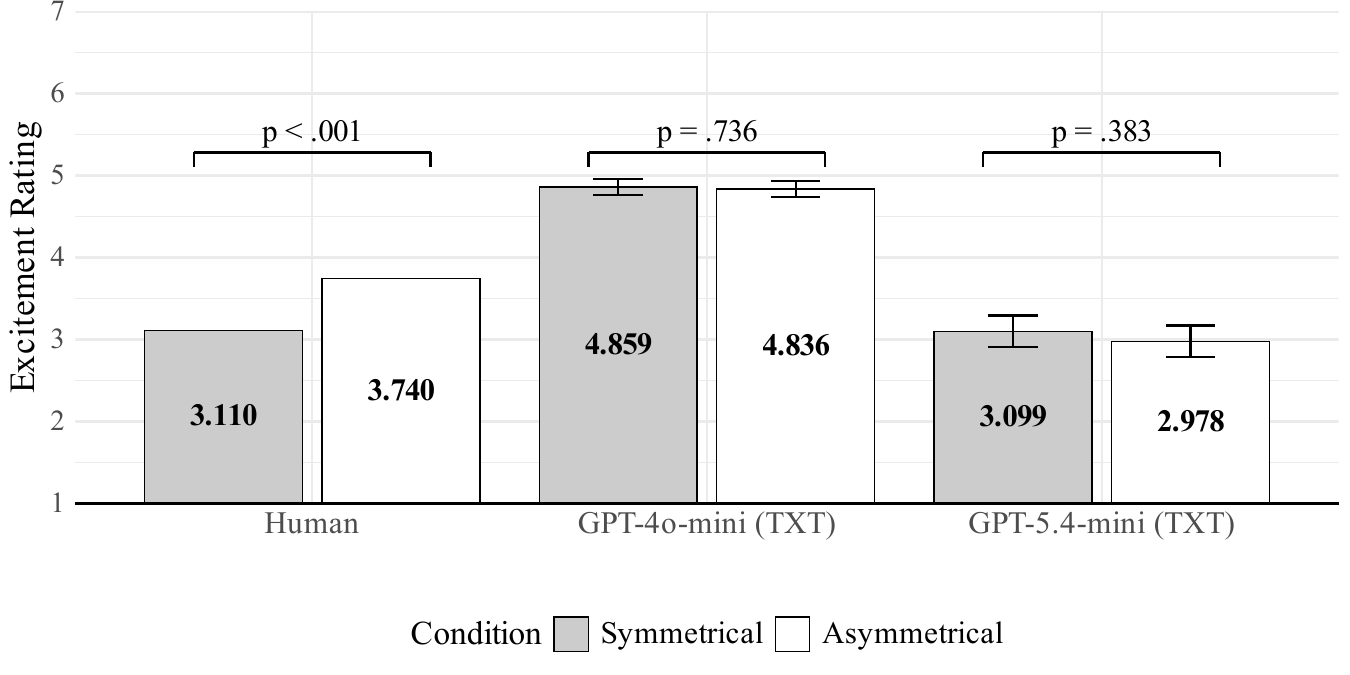}
\caption{\label{fig:s4-excitement}Excitement rating by logo symmetry.}
\par\smallskip{\footnotesize\setstretch{1}\raggedright \emph{Note}: Bars show zero-shot (no-intervention) condition means for the human benchmark, 4o-mini, and 5.4-mini. Error bars represent the 95\% confidence intervals. Error bars for the human sample are omitted because the original study does not report the corresponding standard deviations.\par}
\end{figure}

Humans associate asymmetrical images with perceived excitement \citep{luffarelli2019}. Neither model reproduced this association. 4o-mini did not associate logo asymmetry with excitement (M\textsubscript{Symm} {[}SD\textsubscript{Symm}{]} = 4.859 {[}.511{]} vs. M\textsubscript{Asym} = 4.836 {[}.514{]}, F(1, 217) = .114, \emph{p =} .736, η² = .001; see \hyperref[fig:s4-excitement]{Figure~\ref*{fig:s4-excitement}}). 5.4-mini showed no significant association either, and its estimate was directionally reversed (M\textsubscript{Symm} {[}SD\textsubscript{Symm}{]} = 3.099 {[}1.031{]} vs. M\textsubscript{Asym} = 2.978 {[}1.023{]}, F(1, 218) = .765, \emph{p =} .383, η² = .003). Neither model reproduced the main effect, and at zero-shot neither model met the association criterion for the benchmark's arousal pathway, which requires a replicated main effect first (see details in \hyperref[e-study-4]{Table E4}, \hyperref[web-appendix-e-prompt-intervention-templates-and-results-after-interventions]{Web Appendix E}).

\FloatBarrier
\subsubsection{Study 5: Color saturation and size estimation}\label{study-5-color-saturation-and-size-estimation}

\begin{figure}[H]
\centering
\includegraphics[width=6.5in,height=3.25in]{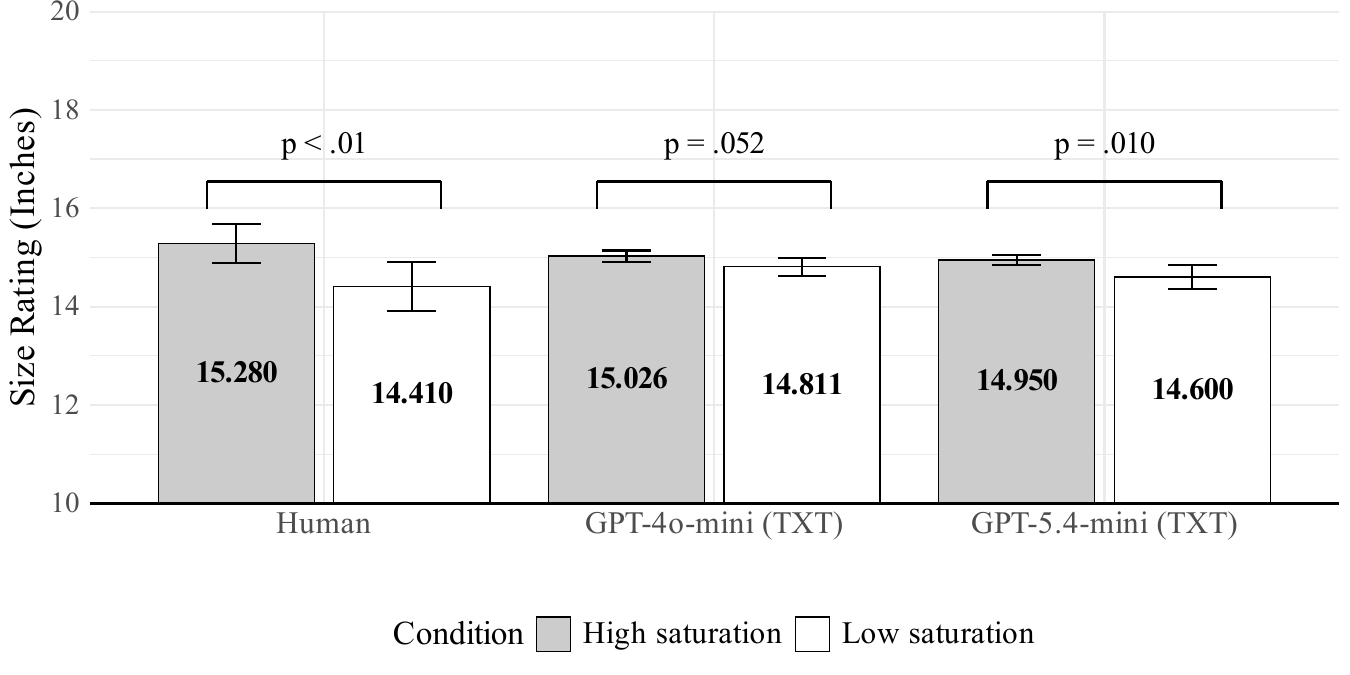}
\caption{\label{fig:s5-size}Perceived size by color saturation.}
\par\smallskip{\footnotesize\setstretch{1}\raggedright \emph{Note}: Bars show zero-shot (no-intervention) condition means for the human benchmark, 4o-mini, and 5.4-mini. Error bars represent the 95\% confidence intervals.\par}
\end{figure}

Humans associate a more color-saturated product with a larger estimated
size \citep{hagtvedt2017}. Despite not meeting the reproduction criterion, 4o-mini produced a directionally consistent effect with a borderline significant p-value and a smaller effect size than the human benchmark (M\textsubscript{High\_Saturation}
{[}SD\textsubscript{High\_Saturation}{]} = 15.026 {[}.362{]} vs. M\textsubscript{Low\_Saturation} = 14.811 {[}.569{]}, F(1, 74) = 3.895, \emph{p =} .052, η² = .050;
human benchmark: η² = .087; see \hyperref[fig:s5-size]{Figure~\ref*{fig:s5-size}}).
5.4-mini reproduced the association (M\textsubscript{High\_Saturation} {[}SD\textsubscript{High\_Saturation}{]} = 14.950 {[}.316{]} vs. M\textsubscript{Low\_Saturation} = 14.600 {[}.778{]}, F(1, 78) = 6.949, \emph{p =} .010, η² = .082), with an effect size closer to humans'.

We followed up by testing whether 5.4-mini, the model that reproduced the main effect, also reproduced the associations underlying the benchmark pathway. The indirect effect of saturation on size through attention, the benchmark pathway in \citet{hagtvedt2017}, was not significant (β = .022, 95\% CI {[}$-$.169, .226{]}). In a separate model for the upstream link, the indirect effect of saturation on attention through arousal was significant (β = .334, 95\% CI {[}.241, .433{]}; mediation outcomes are summarized in \hyperref[e-study-5]{Table E5}, \hyperref[web-appendix-e-prompt-intervention-templates-and-results-after-interventions]{Web Appendix E}). The model thus reproduced the upstream links, saturation heightening arousal and drawing attention, but attention did not carry through to a larger size estimate. This reinforces that even when an LLM reproduces a main effect, the semantic association it expresses can deviate from the one underlying the human effect.

\FloatBarrier
\subsubsection{Study 6: Logo stability and perceived utility}\label{study-6-logo-stability-and-perceived-utility}

Study 6 tests whether the effect of logo stability on perceived utility depends on the product category. \citet{rahinel2016} show that an unstable logo (e.g., a square rotated on its vertex) signals a need for safety, thereby increasing the perceived utility of safety-oriented products. We followed their mixed design, with logo stability and category type as between-subjects factors and product replicate as a within-subjects factor. Perceived utility is measured as the mean of the usefulness, importance, and usage-likelihood ratings, averaged over the six products within a category. First, as a background check consistent with the human benchmark, both models rated safety-oriented products as more useful than neutral products (4o-mini: M\textsubscript{Safety} = 8.059 vs. M\textsubscript{Neutral} = 6.977, F(1, 157) = 1373.659, \emph{p \textless{}} .001, η² = .897; 5.4-mini: M\textsubscript{Safety} = 7.159 vs. M\textsubscript{Neutral} = 6.184, F(1, 152) = 324.094, \emph{p \textless{}} .001, η² = .681). Neither model, however,
showed a main effect of logo stability on perceived utility (4o-mini: F(1, 157) = .020, \emph{p =} .888, η² = .000; 5.4-mini: F(1, 152) = .481, \emph{p =} .489, η² = .003).

\begin{figure}[H]
\centering
\includegraphics[width=\linewidth]{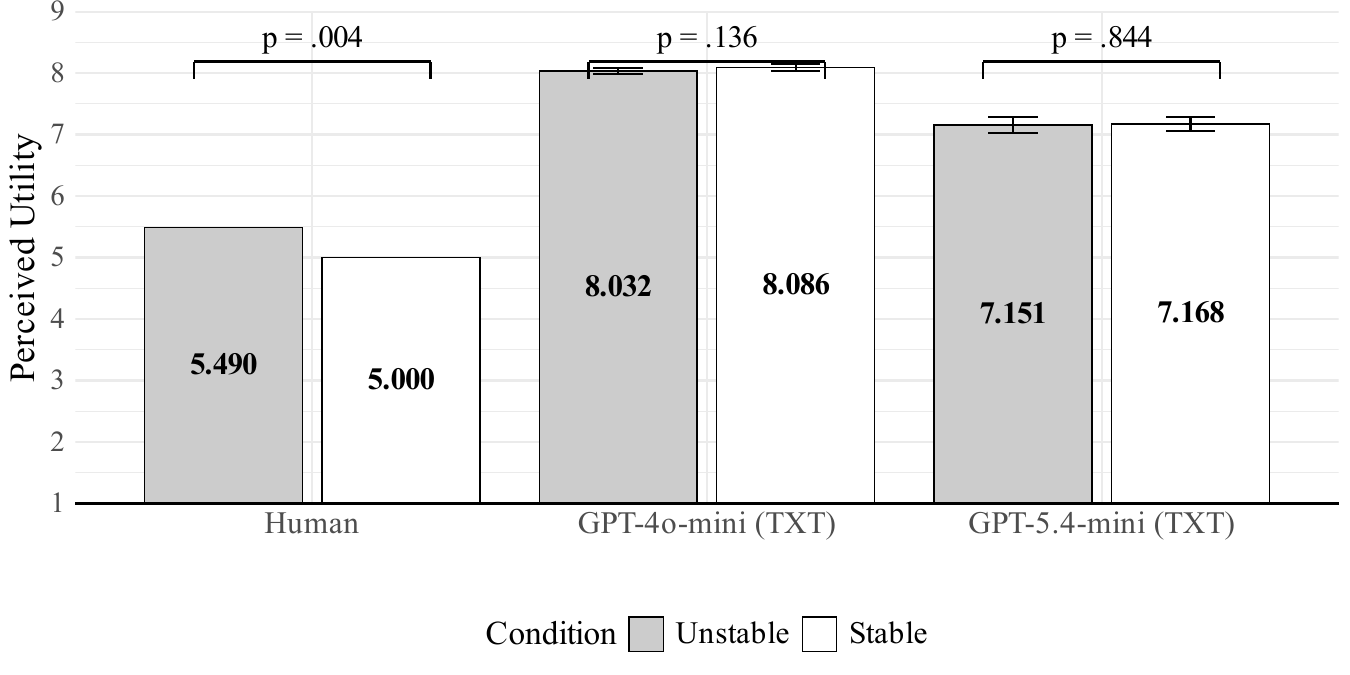}
\caption{\label{fig:s6-utility}Perceived utility by logo stability for safety-oriented products.}
\par\smallskip{\footnotesize\setstretch{1}\raggedright \emph{Note}: Bars show zero-shot (no-intervention) condition means for the human benchmark, 4o-mini, and 5.4-mini. Error bars represent the 95\% confidence intervals. Error bars for the human sample are omitted because the original study does not report the corresponding standard deviations.\par}
\end{figure}

Second, we examined the logo-stability $\times$ category-type interaction. For 4o-mini, the interaction was significant (F(1, 155) = 4.373, \emph{p =} .038, η² = .027), but its source differs from the human benchmark: The utility rating was lower under the unstable logo in the safety-oriented category (M\textsubscript{unstable} {[}SD\textsubscript{unstable}{]} = 8.032 {[}.141{]} vs. M\textsubscript{stable} = 8.086 {[}.179{]}, F(1, 78) = 2.269, \emph{p =} .136, η² = .028; see \hyperref[fig:s6-utility]{Figure~\ref*{fig:s6-utility}}) and higher in the neutral category (M\textsubscript{unstable} {[}SD\textsubscript{unstable}{]} = 7.011 {[}.188{]} vs. M\textsubscript{stable} = 6.944 {[}.215{]}, F(1, 77) = 2.164, \emph{p =} .145, η² = .027), with neither simple effect reaching significance. For 5.4-mini, the interaction was not significant (F(1, 150) = 2.366, \emph{p =} .126, η² = .016); the unstable logo did not increase utility for safety-oriented products (M\textsubscript{unstable} {[}SD\textsubscript{unstable}{]} = 7.151 {[}.407{]} vs. M\textsubscript{stable} = 7.168 {[}.364{]}, F(1, 74) = .039, \emph{p =} .844, η² = .001) but did so in the neutral category (M\textsubscript{unstable} {[}SD\textsubscript{unstable}{]} = 6.258 {[}.261{]} vs. M\textsubscript{stable} = 6.110 {[}.287{]}, F(1, 76) = 5.691, \emph{p =} .020, η² = .070), the category where humans show no effect. In sum, neither model reproduced the pattern in the human benchmark that an unstable logo raises utility for safety-oriented products but not neutral products. 4o-mini's interaction reflects a crossover pattern opposite to the benchmark, and 5.4-mini expresses the effect in the wrong category.

\subsection{Robustness of findings}\label{robustness-of-findings}

We conducted five robustness checks. First, to assess the stability of the results, we repeated each between-subjects experiment five times and each within-subjects experiment twice. The results were stable across these runs (\hyperref[table-f1-test-retest-reliability-for-5-different-runs-temperature-1]{Table F1}, \hyperref[web-appendix-f-robustness-of-findings]{Web Appendix F}, illustrates with Study 1), so each cell in the main text reports a single run and the full set of replicates is available on request.

Second, we assessed temperature sensitivity by re-running Study 1 across a range of temperatures (0.5, 1.0, and 1.5). The manipulation checks and main effects remained directionally and statistically consistent (\hyperref[table-f2-different-temperature-settings-in-gpt]{Table F2}, \hyperref[web-appendix-f-robustness-of-findings]{Web Appendix F}). The response variance stayed narrow across temperature settings.

Third, we investigated whether different image resolutions affected the LLMs' judgments. We found that the manipulation checks and the main effects remained stable (\hyperref[table-f3-resolution-of-the-visual-stimuli-256-512-1024-temperature-1]{Table F3}, \hyperref[web-appendix-f-robustness-of-findings]{Web Appendix F}), suggesting our results are not an artifact of the 512×512 standard.

Fourth, a methodological concern in programmatic experiments is serial correlation across consecutive API calls, which violates the independence assumption. We assessed the serial correlation by modeling the rating \(Y_{ri}\) of the \emph{i}-th AI participant in replicate \emph{r} as a function of the preceding participant's rating \(Y_{r,i - 1}\) in the same replicate:

\[Y_{ri} = \beta\ Condition_{ri} + \lambda\ Y_{r,i - 1} + \eta_{r} + \varepsilon_{ri}.\]

A statistically significant carryover coefficient ($\lambda$) would indicate a
violation of the independence assumption. We included the binary
condition indicator (\(Condition_{ri}\)) and replicate fixed effects
(\(\eta_{r}\)) to capture the potential difference across experiment
replicates. Standard errors are clustered at the replicate level.
\hyperref[table-f4-carryover]{Table F4}, \hyperref[web-appendix-f-robustness-of-findings]{Web Appendix F}, shows that the coefficients of
carryover ($\lambda$) are not significantly different from zero for any
dependent variable in Study 1.

Fifth, we investigated potential heterogeneous effects in personas by adding (de-meaned) age, gender, and their interactions with the treatment to each study's canonical specification (full specifications and coefficients in \hyperref[table-c2-combined-study1]{Tables C2--C7}, \hyperref[web-appendix-c-heterogeneity]{Web Appendix C}). These terms were almost all nonsignificant, and the few exceptions showed no consistent pattern across configurations. The only exception is Study 4, where older personas gave lower excitement ratings in all four configurations. In summary, we do not detect systematic heterogeneity in treatment effects across the assigned personas.

\section{Mitigating perceptual misalignment through in-context learning}\label{mitigating-the-misalignment-using-in-context-learning}

When a synthetic consumer fails to reproduce a human benchmark, we conceptualize this as a conflict between the two priors introduced in our theoretical framework. Without explicit guidance, a model may default to its technical prior (e.g., treating ``color saturation'' purely as an aesthetic attribute) or activate a competing psychological prior (e.g., linking an ``angular logo'' to ``support'' rather than to the intended ``softness''). We also showed that analyzing the model's textual rationales reveals which specific prior has been anchored onto.

To mitigate this misalignment between AI outputs and human perception, we
use ICL, an accessible intervention that practitioners can implement directly via prompting. Following Xie et al.\ (\citeyear{xie2022}), we theoretically ground ICL as a process of implicit Bayesian inference that supplies necessary contextual evidence to shift the model's probability distribution toward the target human-like psychological prior. In this section, we address RQ2 by examining whether injecting conceptual or empirical in-context evidence can successfully steer a misaligned synthetic consumer toward the human benchmark, and whether success depends on how the agent is configured (model generation and input format).

\subsection{Testing the effectiveness of prompts with conceptual vs.\ empirical evidence}\label{testing-the-effectiveness-of-prompts-with-conceptual-vs-empirical-evidence}

To steer the model from its technical default toward the intended psychological
target, we designed two forms of evidence-based prompts. We integrated the
intervention text directly into the system prompt during the initial
persona instantiation phase (Step 1, \hyperref[fig:api-example]{Figure~\ref*{fig:api-example}}), alongside the demographic profile and before the visual stimulus, priming the AI participant to evaluate the subsequent image through the lens of the
provided information (\hyperref[tab:icl-conditions]{Table~\ref*{tab:icl-conditions}}):

\begin{enumerate}
\def\labelenumi{(\arabic{enumi})}
\item
  Conceptual Evidence: We provided the models with the abstract section
  of the original research, which outlines the theoretical links to the
  psychological prior. The text provides the relevant constructs and the directional nature of the association, but deliberately omits quantitative statistics. This condition tests whether a purely qualitative, theoretical account is sufficient to recalibrate the model.
\item
  Empirical Evidence: We prompted the model with the specific statistical findings typically reported in the results section of the original research. In addition to the
  conceptual links, we supplied concrete distributional evidence featuring explicit
  input-output mappings (Xie et al., \citeyear{xie2022}). Because this prompt explicitly contains the benchmark statistics, this condition does not aim to assess spontaneous reproduction; rather, it measures the upper bound of how effectively concrete evidence can steer the model. Conceptually, this mirrors how a marketing manager might use historical campaign data or market research findings to construct a behavioral digital twin of a specific consumer segment.
\end{enumerate}

{
{\footnotesize\begin{longtable}[]{@{}
  >{\raggedright\arraybackslash}p{(\linewidth - 4\tabcolsep) * \real{0.18}}
  >{\raggedright\arraybackslash}p{(\linewidth - 4\tabcolsep) * \real{0.41}}
  >{\raggedright\arraybackslash}p{(\linewidth - 4\tabcolsep) * \real{0.41}}@{}}
\caption{\label{tab:icl-conditions}Comparison of in-context learning (ICL) intervention conditions}\tabularnewline
\toprule\noalign{}
\begin{minipage}[b]{\linewidth}\raggedright
\end{minipage} & \begin{minipage}[b]{\linewidth}\raggedright
\textbf{Conceptual Evidence}
\end{minipage} & \begin{minipage}[b]{\linewidth}\raggedright
\textbf{Empirical Evidence}
\end{minipage} \\
\midrule\noalign{}
\endfirsthead
\toprule\noalign{}
\begin{minipage}[b]{\linewidth}\raggedright
\end{minipage} & \begin{minipage}[b]{\linewidth}\raggedright
\textbf{Conceptual Evidence}
\end{minipage} & \begin{minipage}[b]{\linewidth}\raggedright
\textbf{Empirical Evidence}
\end{minipage} \\
\midrule\noalign{}
\endhead
\bottomrule\noalign{}
\endlastfoot
\textbf{What It Tests} & Evaluates whether a qualitative, directional account is sufficient to improve benchmark alignment. & Establishes the upper bound of calibration when the model receives precise distributional statistics. \\
\textbf{Information Provided} & Theoretical constructs, key terminology, and directional hypotheses. & Descriptive statistics (means, SDs), inferential statistics ($F$-tests, $p$-values), and exact input-output mappings. \\
\textbf{Operationalization} & Integration of the target study's published abstract into the system prompt. &
Integration of the target study's results section into the system prompt. \\
\textbf{Prompt Example (Study 1)} & \emph{Logos frequently include
textual and/or visual design elements that are descriptive of the type
of product/service that brands market... more (vs. less) descriptive
logos are easier to process and thus elicit stronger impressions of
authenticity, which consumers value.} & \emph{The more descriptive logos
(M = 5.89) elicited significantly stronger impressions of authenticity
than the less descriptive logos (M = 4.82; F(1, 176) = 17.69, p
\textless{} .001)... mediated moderation regression analysis with ease
of processing as the mediator...} \\
\end{longtable}}
}

As noted in our theoretical background, the input format can dictate how evidence is processed by the model. In a plain-text structure, statistical figures are embedded within narrative sentences, requiring the model to parse the text to extract them. On the other hand, the JSON format isolates individual statistics under pre-specified fields. This distinction is particularly important for the empirical evidence condition, given it is dense with quantitative data. Here, we defined the JSON schema (\hyperref[table-e8-tagmap]{Table E8}, \hyperref[web-appendix-e-prompt-intervention-templates-and-results-after-interventions]{Web Appendix E}) based on the shared structural components of the six studies. As a validation check, an independent LLM tasked with mapping the plain-text sentences back to this schema reproduced our categorizations with an accuracy rate of 80\% to 94\%.

\subsection{Mean-level alignment after intervention}\label{both-icl-approaches-mitigate-misalignments-in-main-effects}

We first assessed whether prompts with contextual evidence helped align the models' inferential conclusions (main effects) with human baselines across the six studies. \hyperref[tab:intervention-main]{Table~\ref*{tab:intervention-main}} shows outcomes of every format $\times$ model configuration. We find that both conceptual and empirical evidence generally steered the models toward the human benchmark. Notably, empirical evidence yielded the most consistent gains, though its effectiveness varied substantially across model generations.

Specifically, empirical evidence moved a majority of configurations to directional consistency with the human benchmark. In Study 6, for example, all configurations became directionally consistent, with both GPT-5.4-mini formats reaching statistical significance. Prompts leveraging conceptual evidence also recovered directionally consistent main effects in several previously null or reversed cells. Study 2 provides a clear illustration. All four configurations transitioned from null effects at zero-shot to significant effects aligned with human data ($\eta^{2} = .176$ for 4o-mini and $.049$ for 5.4-mini in the plain-text condition).

Despite these directional improvements, the recovered effect sizes were consistently attenuated compared to the original human studies. In Study 1, empirical evidence shifted 5.4-mini from a nominally reversed null to a significant positive effect; yet, the recovered effect sizes ($\eta^{2}=.048$ and $\eta^{2}=.034$) remained roughly half the magnitude of the human benchmark ($\eta^{2}=.091$).

Furthermore, the pattern of success in alignment was not uniform. Study 4 is the only instance where the newer model failed to recover the effect across all configurations, whereas the older model successfully recovered it when provided with empirical evidence. A notable anomaly occurred in Study 5, where 4o-mini never achieved significance, and the contextual interventions effectively attenuated the main effect rather than strengthening it (\hyperref[e-study-5]{Table E5}, \hyperref[web-appendix-e-prompt-intervention-templates-and-results-after-interventions]{Web Appendix E}). The models passed $100\%$ of the manipulation checks under both evidence conditions, indicating that the interventions recalibrated the models' psychological priors without impairing their capabilities in visual recognition.

\begin{landscape}
\thispagestyle{empty}%
\AddToShipoutPictureFG*{\put(\LenToUnit{\dimexpr\paperwidth-0.7in},\LenToUnit{0.5\paperheight}){\rotatebox{90}{\makebox[0pt]{\thepage}}}}%
\begin{center}\singlespacing\footnotesize
{
\setlength{\tabcolsep}{4pt}
\begin{longtable}[]{@{}lcccccccccccc@{}}
\caption{\label{tab:intervention-main}Main effects after prompt-based interventions by format $\times$ model}\\
\toprule
 & \multicolumn{3}{c}{GPT-4o-mini (TXT)} & \multicolumn{3}{c}{GPT-4o-mini (JSON)} & \multicolumn{3}{c}{GPT-5.4-mini (TXT)} & \multicolumn{3}{c}{GPT-5.4-mini (JSON)} \\
\cmidrule(lr){2-4}\cmidrule(lr){5-7}\cmidrule(lr){8-10}\cmidrule(lr){11-13}
Study & No Int. & Concep. & Empir. & No Int. & Concep. & Empir. & No Int. & Concep. & Empir. & No Int. & Concep. & Empir. \\
\midrule\endfirsthead
\toprule
 & \multicolumn{3}{c}{GPT-4o-mini (TXT)} & \multicolumn{3}{c}{GPT-4o-mini (JSON)} & \multicolumn{3}{c}{GPT-5.4-mini (TXT)} & \multicolumn{3}{c}{GPT-5.4-mini (JSON)} \\
\cmidrule(lr){2-4}\cmidrule(lr){5-7}\cmidrule(lr){8-10}\cmidrule(lr){11-13}
Study & No Int. & Concep. & Empir. & No Int. & Concep. & Empir. & No Int. & Concep. & Empir. & No Int. & Concep. & Empir. \\
\midrule\endhead
1 (Authenticity) & \textbf{$+$ ($<$.001)} & \textbf{$+$ ($<$.001)} & \textbf{$+$ ($<$.001)} & \textbf{$+$ ($<$.001)} & \textbf{$+$ ($<$.001)} & \textbf{$+$ ($<$.001)} & $-$ (.586) & $+$ (.420) & \textbf{$+$ (.003)} & $-$ (.793) & $+$ (.466) & \textbf{$+$ (.013)} \\
2 (Selected saturation) & $+$ (.631) & \textbf{$+$ ($<$.001)} & \textbf{$+$ (.006)} & $+$ (.202) & \textbf{$+$ ($<$.001)} & \textbf{$+$ (.008)} & $+$ (.288) & \textbf{$+$ (.022)} & \textbf{$+$ ($<$.001)} & $+$ (.373) & \textbf{$+$ (.035)} & \textbf{$+$ ($<$.001)} \\
3 (Comfort) & \textbf{$-$ (.012)} & $+$ (.291) & $+$ (.603) & $-$ (.110) & $+$ (.220) & $+$ (.133) & \textbf{$+$ (.022)} & \textbf{$+$ ($<$.001)} & \textbf{$+$ ($<$.001)} & \textbf{$+$ (.009)} & \textbf{$+$ ($<$.001)} & \textbf{$+$ ($<$.001)} \\
3 (Durability) & $-$ (.177) & $-$ (.258) & $-$ (.534) & $-$ (.229) & $-$ (.197) & $-$ (.512) & $-$ (.321) & \textbf{$+$ ($<$.001)} & \textbf{$+$ ($<$.001)} & $-$ (.321) & \textbf{$+$ ($<$.001)} & \textbf{$+$ ($<$.001)} \\
4 (Excitement) & $-$ (.736) & $+$ (.595) & \textbf{$+$ (.048)} & $-$ (.230) & $+$ (.714) & \textbf{$+$ (.014)} & $-$ (.383) & $-$ (.821) & $-$ (.833) & $-$ (.562) & $+$ (.826) & $-$ (.772) \\
5 (Size estimation) & $+$ (.052) & $+$ (.150) & $+$ (.146) & $+$ (.055) & $+$ (.189) & $+$ (.314) & \textbf{$+$ (.010)} & $+$ (.060) & \textbf{$+$ (.037)} & \textbf{$+$ (.002)} & $+$ (.060) & \textbf{$+$ (.029)} \\
6 (Utility) & $-$ (.136) & $+$ (.264) & $+$ (.111) & $-$ (.401) & $+$ (.581) & $+$ (.443) & $-$ (.844) & \textbf{$+$ ($<$.001)} & \textbf{$+$ (.005)} & $-$ (.874) & \textbf{$+$ (.006)} & \textbf{$+$ (.001)} \\
\bottomrule
\end{longtable}

\smallskip
{\footnotesize\setstretch{1}\raggedright\emph{Note}: ``No Int./Concep./Empir.'' $=$ No Intervention / prompt with conceptual evidence / prompt with empirical evidence. ``$+$/$-$'' indicates the main-effect direction consistent/inconsistent with the human finding (shown even when $n.s.$); parentheses contain p-values; bold marks significant effects ($p<.05$). Parentheses after each study number give the dependent variable. Study 6 reports utility for the safety-oriented product category, measured as the mean of the usefulness, importance, and usage-likelihood ratings. The JSON empirical cells use the fully tagged version of \hyperref[table-e8-tagmap]{Table E8}, \hyperref[web-appendix-e-prompt-intervention-templates-and-results-after-interventions]{Web Appendix E}. Complete results appear in \hyperref[e-study-1]{Tables E1--E6}, \hyperref[web-appendix-e-prompt-intervention-templates-and-results-after-interventions]{Web Appendix E}.\par}
}
\end{center}
\end{landscape}

\subsection{\texorpdfstring{The model generation matters more than the input format}{The model generation matters more than the input format}}\label{model-vs-format}

Regarding the configuration levers in RQ2, we found that in zero-shot settings, the two input formats yielded statistically indistinguishable results, with no single configuration superior for baseline alignment. Under intervention, however, model generation became a determining factor. When provided with empirical evidence, 5.4-mini successfully reproduced the human effect in six of the seven experimental cells (two of which were already aligned at zero-shot), whereas 4o-mini reproduced the human effect in three (one already aligned at zero-shot). These success rates were identical across both input formats within each respective model. Overall gains were most pronounced in the newer 5.4-mini, and 4o-mini showed robust improvement primarily under the empirical evidence condition. For conceptual evidence, the JSON input wraps the same abstract within a fixed prefix and suffix, so the structural difference between formats is just a few framing tokens. However, because empirical evidence contains significantly denser quantitative background text, we further investigated whether alignment was sensitive to the choice of JSON style.\footnote{We supplied the identical empirical evidence across three varying JSON styles: no tags, section headings only, and the fully granular schema detailed in \hyperref[table-e8-tagmap]{Table E8}, \hyperref[web-appendix-e-prompt-intervention-templates-and-results-after-interventions]{Web Appendix E}. Results by style are available upon request.} 5.4-mini successfully recovered the same studies regardless of the complexity of JSON style, whereas 4o-mini showed degraded performance when the evidence was tagged too granularly. Therefore, alignment is less contingent on the mere presence of JSON formatting than on the schema's granularity, a sensitivity unique to the older model. This pattern echoes \citet{he2024}, who find that sensitivity to input formatting declines as model capability increases.

\subsection{Reproduction of the underlying semantic associations}\label{results-prompt-further-steered-mediation-pathways}

We next posed a stricter test: can contextual evidence also lead the model to reproduce the semantic association underlying the original study's psychological pathway (i.e., its mediator), beyond merely reproducing the main effects? A pathway's semantic association was deemed successfully reproduced if (1) the main effect was reproduced in that cell, (2) the indirect effect was statistically significant, and (3) the mediator (or its semantic synonym) appeared among the keywords of that dependent variable's three
largest topics. We tested the five psychological pathways in each cell (i.e., ease of processing in Study 1, softness in Study 3 with two DVs, arousal in Study 4, and attention in Study 5). 5.4-mini steadily improved as we added evidence. In both formats, it went from recovering at most one of the five associations at baseline (zero-shot) to three out of five when given empirical evidence. In contrast, 4o-mini barely reacted to the interventions, never recovering more than one pathway.

As an additional check, we measured how prominently the models discussed the mediator in their rationales. Using BERTopic, which summarizes topics using keyword lists, we grouped words with the same root (e.g., \emph{soft}, \emph{softness}, \emph{softer}) into a single item. We then calculated the share of these keywords that matched the mediator. To prevent pathways with multiple dependent variables from dominating the results, we averaged these shares equally across the five pathways. For 5.4-mini, this mediator share steadily increased in both formats as evidence was added (e.g., rising from 12.4\% to 19.6\% in JSON). 4o-mini's share stayed flat in both formats (JSON: 14.7\% to 13.0\%; TXT: 15.0\% to 13.4\%), matching almost no real improvement in actual association recovery (0 to 1 out of 5) (\hyperref[tab:mediation]{Table~\ref*{tab:mediation}}).

{
{\footnotesize\setlength{\tabcolsep}{4pt}\setlength\LTleft{\fill}\setlength\LTright{\fill}\begin{longtable}[]{@{}lcccccc@{}}
\caption{\label{tab:mediation}Reproduction of the semantic associations underlying the human pathways, and mediator-keyword presence, by format $\times$ model}\\
\toprule
 & \multicolumn{3}{c}{Associations reproduced (out of 5)} & \multicolumn{3}{c}{Keyword-group share (\%)} \\
\cmidrule(lr){2-4}\cmidrule(lr){5-7}
Configuration & No Intervention & Conceptual & Empirical & No Intervention & Conceptual & Empirical \\
\midrule\endfirsthead
\toprule
 & \multicolumn{3}{c}{Associations reproduced (out of 5)} & \multicolumn{3}{c}{Keyword-group share (\%)} \\
\cmidrule(lr){2-4}\cmidrule(lr){5-7}
Configuration & No Intervention & Conceptual & Empirical & No Intervention & Conceptual & Empirical \\
\midrule\endhead
4o-mini TXT & 0/5 & 0/5 & 1/5 & 15.0 & 11.9 & 13.4 \\
4o-mini JSON & 1/5 & 0/5 & 1/5 & 14.7 & 11.5 & 13.0 \\
5.4-mini TXT & 1/5 & 2/5 & 3/5 & 13.2 & 19.6 & 21.8 \\
5.4-mini JSON & 0/5 & 2/5 & 3/5 & 12.4 & 16.0 & 19.6 \\
\bottomrule
\end{longtable}}
{\footnotesize\setstretch{1}\raggedright\emph{Note}: An association counts as reproduced when it meets three conditions: (1) The main effect is reproduced in that cell; (2) The indirect effect is significant; (3) The mediator or a synonym (e.g., soft or cushioned for softness) appears among the keywords of that DV's three largest topics. ``Keyword-group share'' is the unweighted mean of the five pathway shares. A pathway's share is the proportion of keyword groups in its DV's three largest topics that contain the mediator, counting words with the same root as one group. Averaging with equal weight keeps pathways with more dependent variables from dominating the figure. For example, 12.4\% for 5.4-mini JSON at zero-shot is the mean of its five pathway shares (13/93, 7/29, 4/54, 9/54, and 0/54 keyword groups containing the mediator, for Studies 1, 3-comfort, 3-durability, 4, and 5, respectively).\par}
}

\subsection{\texorpdfstring{Variance shrinkage persists in every configuration}{Variance shrinkage persists in every configuration}}\label{variance-shrinkage-persists-in-every-configuration}

\begin{figure}[H]
\centering
\includegraphics[width=5.6in]{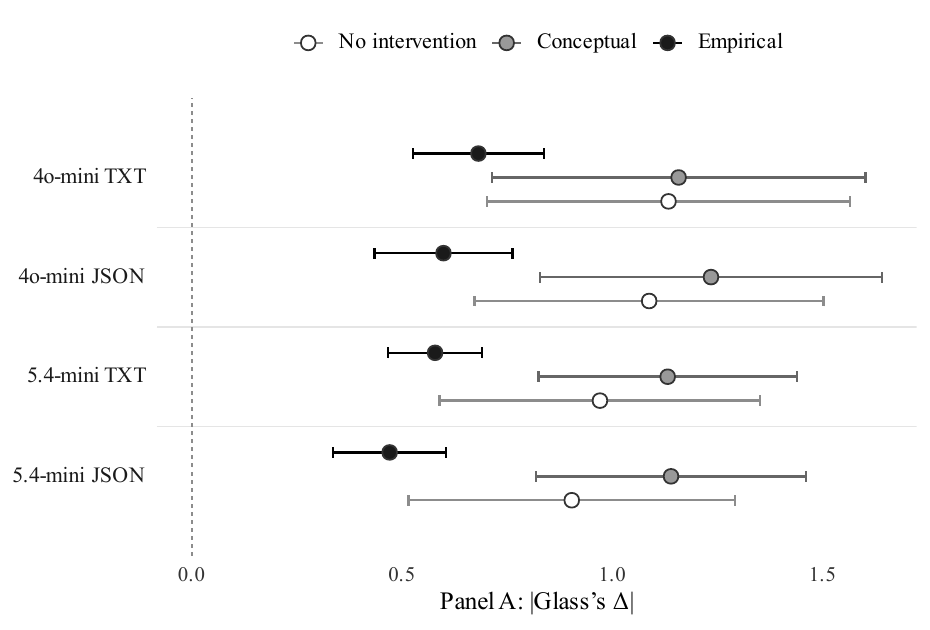}\par\vspace{0.3em}
\includegraphics[width=5.6in]{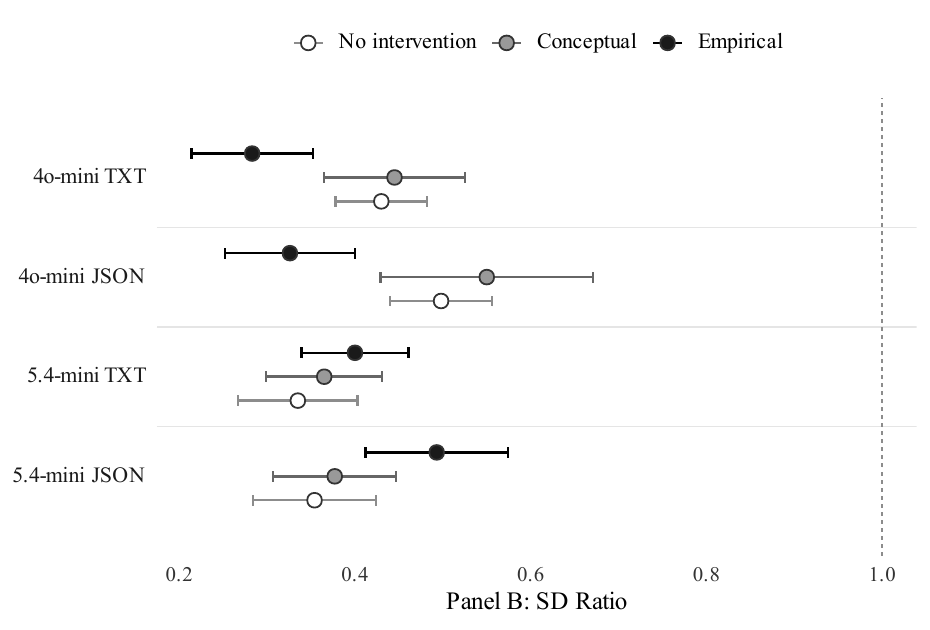}
\caption{\label{fig:trend-dumbbell}$|$Glass's $\Delta|$ (Panel A) and SD Ratio (Panel B) by format $\times$ model across the three interventions.}
\par\smallskip{\footnotesize\setstretch{1}\raggedright \emph{Note}: Each row represents a single GPT configuration. Each point denotes the mean across the eight condition cells ($\pm 1$ SE); the error bars describe between-cell heterogeneity, not sampling uncertainty. The dashed line marks the human benchmark ($|$Glass's $\Delta|=0$; SD Ratio $=1$).\par}
\end{figure}

Next, we checked whether any configuration reproduced the natural spread of human responses to address RQ3. We used two distribution-level metrics from \citet{peng2025}: Glass's $\Delta = (\overline{X}_{\mathrm{LLM}} - \overline{X}_{\mathrm{Human}})/\mathit{SD}_{\mathrm{Human}}$, which measures the standardized discrepancy of the mean, and the SD Ratio $= \mathit{SD}_{\mathrm{LLM}}/\mathit{SD}_{\mathrm{Human}}$, which measures how well the variances align. Both metrics require the human standard deviation, so we calculated them for each format $\times$ model configuration across the eight condition cells that report one: the two conditions in each of Studies 2, 3-comfort, 3-durability, and 5 (\hyperref[fig:trend-dumbbell]{Figure~\ref*{fig:trend-dumbbell}}).

We observed two patterns. First, adding empirical evidence moved every configuration closer to the human mean, shrinking $|$Glass's $\Delta|$ toward zero across all four setups. However, this improvement was not statistically significant within any single setup (e.g., 4o-mini (TXT): $F(2, 14) = 1.788$, $\emph{p} = .203$). Second, the narrow spread of responses is not a cost of our calibration process; it is a persistent flaw across the board. All twelve combinations of configuration and intervention produced an SD Ratio below $1.0$, and every zero-shot value fell below $.50$. This shows that the synthetic responses were already far too uniform before we introduced any evidence.\footnote{The GPT-4o-mini (JSON) average under conceptual evidence approaches $1.0$ only because of one outlier. Cell-level values are available upon request.}

The interventions did not consistently worsen this shrinkage. For 5.4-mini, the SD Ratio increased slightly toward the human baseline under empirical evidence (TXT: $.335 \to .400$; JSON: $.354 \to .493$). For 4o-mini, it decreased (TXT: $.430 \to .283$; JSON: $.498 \to .326$). Ultimately, synthetic consumers systematically fail to capture human heterogeneity across all configurations \citep{bisbee2024}, and neither the model, the format, the prompt, nor the sampling temperature resolved it (see \hyperref[table-f2-different-temperature-settings-in-gpt]{Table F2}, \hyperref[web-appendix-f-robustness-of-findings]{Web Appendix F} for the temperature check).

This shrinkage appears to be an inherent feature of these agents, even with variation in demographic personas—the standard way the industry generates synthetic samples. In-context learning can only soften this shrinkage instead of fixing it completely.

\section{General discussion}\label{general-discussion}

Our research investigates the boundary conditions of using LLMs as reliable proxies for human consumers. Our zero-shot benchmarks of canonical visual marketing effects showed that models reliably see visual features yet frequently fail to perceive the consumer meaning those features carry (RQ1). This misalignment, however, can often be steered. By injecting conceptual or empirical evidence via in-context learning (ICL), we can calibrate the AI toward human benchmarks, with gains that depend more on the underlying model generation than on the input format (RQ2). Yet, a critical limitation persists: While evidence-based prompting corrects the main effects, the simulated responses suffer from substantial variance shrinkage (RQ3).

Despite this limitation, a calibrated agent remains a useful tool for rapid, early-stage screening of visual cues. In the managerial implications, we translate these findings into a governance protocol that teams with basic analytics capacity can run, and we illustrate it with a case study.

\subsection{\texorpdfstring{Theoretical implications}{Theoretical implications}}\label{theoretical-implications}

First, we extend the silicon sampling literature to the visual domain, testing whether LLMs can simulate not only textual preferences but the multimodal interpretation of visual marketing assets. While text-based LLMs show promise in mimicking human responses \citep{ashokkumar2026}, this success does not carry over to these assets, where model judgments systematically depart from the human benchmark. The results provide empirical content to the distinction between a model's technical prior and its psychological prior. By integrating numerical ratings with textual rationales, we identify recurring sources of misalignment. In our manipulation checks, we observed no visual recognition errors. Instead, the misalignments stem from missing or competing semantic associations, in which the AI defaults to a technical or an alternative psychological interpretation. The broader implication is that recognizing a visual cue, reproducing the human outcome, and reproducing the semantic association that underlies the human psychological pathway are distinct tasks. Success in one does not guarantee success in the next.

Second, we evaluate how different types of in-context evidence activate the target psychological prior. We show that providing empirical evidence (e.g., statistical findings) steers the model toward the human benchmark more effectively than conceptual evidence. For the newer model, the empirical prompt also reproduced the semantic association underlying the human pathway in three of the five tests, from a maximum of one at zero-shot. The older model, conversely, never produced more than one successful reproduction. Importantly, as shown in Section~\ref{variance-shrinkage-persists-in-every-configuration}, these interventions do not restore response variance. Consequently, calibrated synthetic consumers are suitable for mean-level comparisons only.

\subsection{\texorpdfstring{Managerial implications}{Managerial implications}}\label{managerial-implications}

\begin{figure}[H]
\centering
\includegraphics[width=\linewidth]{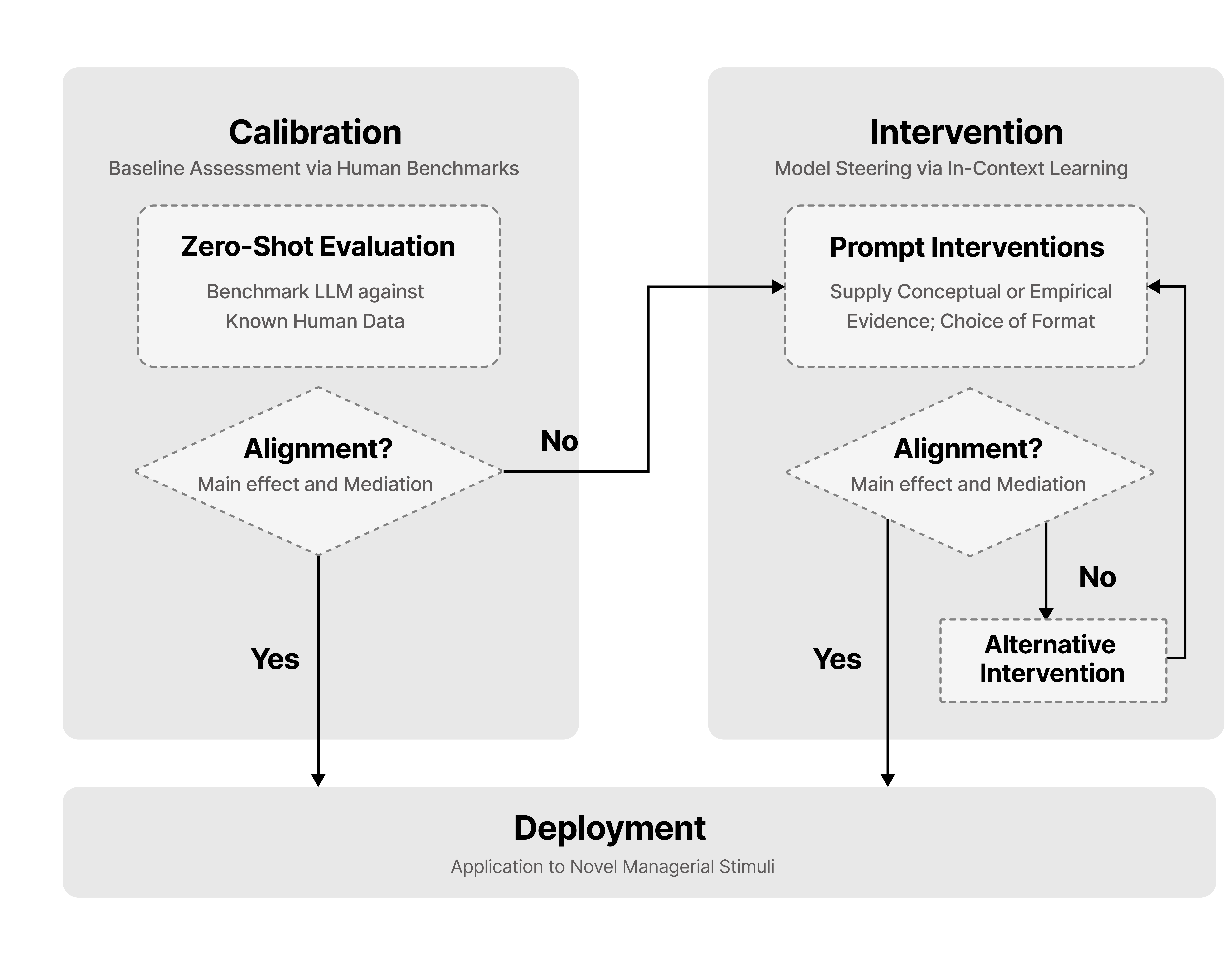}
\caption{\label{fig:framework}The three-step governance protocol (Calibrate, Intervene, Deploy) for deploying synthetic consumers.}
\end{figure}

Our six benchmark experiments show that the text-domain record behind synthetic consumers breaks down at visual judgment, where a model can describe an image accurately yet miss, or reverse, the meaning consumers assign to it. However, the practitioners most receptive to AI are precisely those least equipped to audit it \citep{tully2025}. We therefore develop a governance protocol paired with cautionary guidelines for deploying visual AI agents (\hyperref[fig:framework]{Figure~\ref*{fig:framework}}). It complements recent guidance on how to implement GenAI methods and when they add value \citep{joerling2026} by addressing the prior question of whether an agent's judgments can be trusted at all. This protocol establishes clear gating criteria to determine when a synthetic consumer is reliable and when human oversight is required. A configuration passes calibration only by reproducing a historical human benchmark, at the original study's sample size, in both direction and significance. Comparing the reproduced effect size against the benchmark, as the case study below illustrates, provides a further check. If it fails, the protocol triggers an ICL intervention. If that intervention does not resolve the misalignment, the task reverts to a human panel.

The protocol places calibration before intervention for two reasons. Analytically, intervention is blind without a benchmark. Until an agent's output is compared against known human data, a manager cannot know whether it needs steering or whether steering overcorrected. Even effects already validated with human subjects require this step; furthermore, visual effects are context-dependent, so re-testing is advised in each deployment context \citep{kim2023}. Our results show that some misalignments resist even the strongest evidence prompt, and that an uncalibrated agent can be confidently wrong. Economically, calibration is a one-time fixed cost: Once a given model generation and input format are validated for a stimulus domain and target judgment (e.g., logo authenticity), the agent screens subsequent creative variants at near-zero marginal cost. The investment thus directly relieves the evaluation bottleneck that motivates synthetic consumers in the first place. It recurs only when the configuration changes, such as when the vendor updates the model or the firm moves to a new stimulus domain or target judgment.

This governance is critical because practitioners cannot rely on the anthropomorphic assumption that AI perceives visual assets through a human psychological lens, a view that aligns with recent work warning against treating AI as a direct human substitute \citep{gao2025}. A newer model is no guarantee of alignment: the latest GPT generation was not uniformly better at zero-shot, though it responded far more to steering. Because behavior shifts from version to version, the practice of calibration is what survives model turnover, and firms must rerun it on whichever model they deploy.

This calibration should also cover the prompt architecture. Although plain text and JSON did not differ statistically in our studies, the older model degraded when the empirical evidence was tagged too finely, so the protocol treats the JSON schema as one more parameter to validate rather than a fixed default.

Finally, because response variance remains systematically below human levels in every configuration, strategic decisions that hinge on consumer heterogeneity should always return to a human panel.

\subsubsection{\texorpdfstring{A governance protocol case study}{A governance protocol case study}}\label{managerial-implication-a-governance-protocol-case-study}

\begin{figure}[H]
\centering
{\footnotesize\begin{tabular}{c@{\hspace{3em}}c}
Less Descriptive & More Descriptive \\[0.5ex]
\includegraphics[width=1.72277in,height=1.72277in]{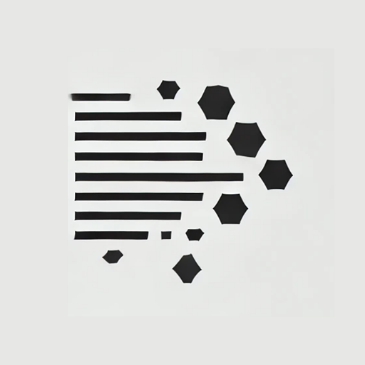} &
\includegraphics[width=1.6913in,height=1.6913in]{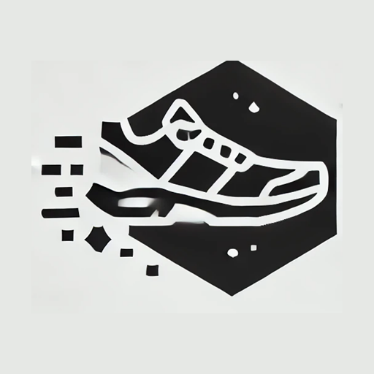} \\
\end{tabular}}
\caption{\label{fig:case-logos}AI-generated logos for a running shoe brand. Both images were created with ChatGPT (OpenAI) as stimuli for the case study.}
\end{figure}

Consider a manager using GenAI to design an ``authentic'' running shoe logo. She first generates two options: one less descriptive and one more descriptive (\hyperref[fig:case-logos]{Figure~\ref*{fig:case-logos}}). To validate
these designs, she plans to use a synthetic panel to predict consumer perceptions of authenticity. She reaches for a new model, GPT-5.4-mini, on the assumption that a state-of-the-art
system will read the logos as her customers would. However, aware of the potential for misalignment between human and model perceptions, she applies the three-step governance protocol to verify the agent before deployment.

\textbf{Step 1: Calibration.} The manager first calibrates the agent on a reference benchmark with known human data (e.g., historical in-house results or published research). She picks \citet{luffarelli2019a}, whose descriptiveness--authenticity effect ($\eta^2 = .091$) closely mirrors her screening question, and evaluates GPT-5.4-mini (TXT) on the study's original stimuli. At zero-shot, the model shows no significant main effect ($p =$ .586, $\eta^2 = .002$; \hyperref[tab:zeroshot-main]{Table~\ref*{tab:zeroshot-main}}), indicating a misalignment with the benchmark. She therefore advances to the intervention stage.

\textbf{Step 2: Intervention.} She re-runs the check on the original stimuli, this time supplying the model with evidence from the benchmark study. She has learned that empirical evidence is more helpful in mitigating this misalignment in human-AI perception than conceptual evidence (\hyperref[tab:intervention-main]{Table~\ref*{tab:intervention-main}}). On the one hand, the empirical prompt restores the human main effect ($p =$ .003) with an effect size of η² $=$ .048, roughly half the human benchmark. On the other hand, the conceptual prompt does not reach significance under this configuration ($p =$ .420). Based on this finding, she selects the empirical prompt for GPT-5.4-mini (TXT) and moves forward.

\textbf{Step 3: Deployment.} Having validated the prompt against a task with a known human benchmark, she now applies it to her own logo pair (\hyperref[fig:case-logos]{Figure~\ref*{fig:case-logos}}). She runs the same data-collection pipeline, simulating 160 AI participants (80 per logo) in each prompt condition (zero-shot versus the validated empirical prompt; full results in \hyperref[e-study-7]{Table E7}, \hyperref[web-appendix-e-prompt-intervention-templates-and-results-after-interventions]{Web Appendix E}).\footnote{For completeness, the manager also evaluates the conceptual prompt on the novel stimuli. Although this prompt restores the directional main effect, it amplifies the effect size more than fourfold ($\eta^2 = .408$). This overshoot underscores the instability of the intervention with only conceptual evidence.} At zero-shot, GPT-5.4-mini (TXT) again shows no significant main effect on her logos (M\textsubscript{Less\_Desc} {[}SD\textsubscript{Less\_Desc}{]} $=$ 4.567 {[}.493{]} vs. M\textsubscript{More\_Desc} $=$ 4.504 {[}.495{]}, F(1, 158) $=$ .641, $p =$ .425, η² $=$ .004), despite clearly distinguishing between the two descriptiveness levels (manipulation check: t(158) $=$ 33.353, $p <$ .001, η² $=$ .876). This outcome reproduces the misalignment we saw, but on new visual assets this time.

After the empirical prompt, the more descriptive logo is rated as more authentic (M\textsubscript{Less\_Desc} {[}SD\textsubscript{Less\_Desc}{]} $=$ 6.033 {[}.755{]} vs. M\textsubscript{More\_Desc} $=$ 6.442 {[}.491{]}, F(1, 158) $=$ 16.444, $p <$ .001, η² $=$ .094). She reads the output as a directional signal rather than a human effect size, since no human baseline exists for her proprietary logos. Because the agent's variance remains below human levels, she ranks designs on average authenticity and takes no segmentation decision from it.

\subsection{\texorpdfstring{Limitations and future research}{Limitations and future research}}\label{limitations-and-future-research}

First, our empirical scope focuses on generic demographic personas across two models from a single vendor's low-cost tier, evaluating numerical ratings of static images. It is unclear whether the reported alignment gaps and variance shrinkage would be of similar magnitude in flagship models, alternative platforms (e.g., Claude, Gemini), or agents conditioned on richer, high-dimensional profiles, as in \citet{toubia2025} and \citet{peng2025}. Extending this governance protocol to dynamic video creatives remains a natural next step for assessing its broader applicability.

Second, we operationalize alignment using published literature rather than absolute ground truth. Given the replication crisis in behavioral science \citep{simmons2011}, an AI's inability to reproduce a finding may reflect instability in the original human effect rather than a shortcoming of the model's perception capability. Furthermore, all six benchmarks predate the training cutoffs for both models. However, rather than undermining our findings, this potential prior data exposure strengthens our core argument: Even if the models possessed latent memory of the published results, they still failed to reproduce most of the human effects at zero-shot. Evaluating models on future, post-cutoff visual experiments could nonetheless distinguish true perceptual alignment from memorized retrieval.

Third, our protocol evaluates models purely on their behavioral inputs and outputs, without attempting to unpack the AI black box. We therefore do not claim that the AI processes visual information through the same internal mechanisms as humans, and because our empirical prompt supplies the benchmark statistics, the resulting alignment demonstrates steerability rather than spontaneous agreement. Future research can investigate whether calibrated agents can move beyond rating images to generating assets that operationalize these latent psychological principles.

\section*{Use of generative AI}
During the preparation of this manuscript, the authors used AI (Claude) for copyediting and stylistic refinement. After the use of this tool, the authors reviewed and revised the manuscript and take full responsibility for the final published article. This declaration concerns manuscript preparation only; the use of large language models as experimental subjects is part of the study's methodology and is reported in Section~\ref{methodology}.

\newpage
\bibliographystyle{apacite}
\bibliography{references}

\clearpage

\setcounter{secnumdepth}{0}
\titleformat{\section}{\normalfont\fontsize{14}{16}\selectfont\bfseries\raggedright}{}{0pt}{}
\titleformat{\subsection}{\normalfont\fontsize{12}{14}\selectfont\bfseries\raggedright}{}{0pt}{}
\titleformat{\subsubsection}{\normalfont\fontsize{12}{14}\selectfont\bfseries\raggedright}{}{0pt}{}
\captionsetup{labelformat=empty,font=normalsize,justification=centering}
\setlength\LTleft{\fill}\setlength\LTright{\fill}

\begin{center}
  {\large\bfseries Web Appendix\par}
  \medskip
  {\large Seeing Is Not Perceiving: When Synthetic Consumers Can and Cannot Pretest Visual Marketing\par}
\end{center}
\bigskip

\section{Web Appendix \texorpdfstring{A}{A}: Pre-registration for studies}\label{web-appendix-a-pre-registration-for-studies}

This section contains all the studies that were registered on
aspredicted.org. Links are listed below by study.

\textbf{Study 1}

\begin{enumerate}
\def\labelenumi{\Alph{enumi}.}
\item
  Without any interventions: \url{https://aspredicted.org/ghd5-p8kc.pdf}
\item
  Prompt with Conceptual Evidence:
  \url{https://aspredicted.org/7wys-msff.pdf}
\item
  Prompt with Empirical Evidence:
  \url{https://aspredicted.org/tdpx-28tg.pdf}
\end{enumerate}

\textbf{Robustness checks (Study 1)}

\begin{enumerate}
\def\labelenumi{\Alph{enumi}.}
\item
  Different temperatures: \url{https://aspredicted.org/8hkf-kfnw.pdf}
\item
  Different resolutions of the visual stimuli:
  \url{https://aspredicted.org/rsgz-pcmg.pdf}
\item
  Test with AI-generated stimuli:
  \url{https://aspredicted.org/qbfx-mnrv.pdf}
\end{enumerate}

\textbf{Study 2}

\begin{enumerate}
\def\labelenumi{\Alph{enumi}.}
\item
  Without any interventions: \url{https://aspredicted.org/bfdz-nc7f.pdf}
\item
  Prompt with Conceptual Evidence:
  \url{https://aspredicted.org/d455-q65j.pdf}
\item
  Prompt with Empirical Evidence:
  \url{https://aspredicted.org/pc4z-b2p3.pdf}
\end{enumerate}

\textbf{Study 3}

\begin{enumerate}
\def\labelenumi{\Alph{enumi}.}
\item
  Without any interventions:
  \url{https://aspredicted.org/v3pb-zpf8.pdf}
\item
  Prompt with Conceptual Evidence:
  \url{https://aspredicted.org/8r8w-972x.pdf}
\item
  Prompt with Empirical Evidence:
  \url{https://aspredicted.org/pjk7-xgpk.pdf}
\end{enumerate}

\textbf{Study 4}

\begin{enumerate}
\def\labelenumi{\Alph{enumi}.}
\item
  Without any interventions: \url{https://aspredicted.org/r7hf-f7xx.pdf}
\item
  Prompt with Conceptual Evidence:
  \url{https://aspredicted.org/qk4f-p3g2.pdf}
\item
  Prompt with Empirical Evidence:
  \url{https://aspredicted.org/ydkv-7ynq.pdf}
\end{enumerate}

\textbf{Study 5}

\begin{enumerate}
\def\labelenumi{\Alph{enumi}.}
\item
  Without any interventions: \url{https://aspredicted.org/8wp3-httb.pdf}
\item
  Prompt with Conceptual Evidence:
  \url{https://aspredicted.org/kydc-6jnz.pdf}
\item
  Prompt with Empirical Evidence:
  \url{https://aspredicted.org/d5dr-g8vm.pdf}
\end{enumerate}

\textbf{Study 6}

\begin{enumerate}
\def\labelenumi{\Alph{enumi}.}
\item
  Without any interventions: \url{https://aspredicted.org/w7qb-5zhq.pdf}
\item
  Prompt with Conceptual Evidence:
  \url{https://aspredicted.org/srzm-prxf.pdf}
\item
  Prompt with Empirical Evidence:
  \url{https://aspredicted.org/dfzj-7x3s.pdf}
\end{enumerate}

\clearpage
\section{Web Appendix \texorpdfstring{B}{B}: Description of studies and visual stimuli}\label{web-appendix-b-description-of-studies-and-visual-stimuli}

This section summarizes the objectives, sample sizes, research designs,
and important variables of each study we investigated, along with the
visual stimuli from the original papers.

\subsection{B.1 Study 1: Based on Study 1 in \texorpdfstring{\citet{luffarelli2019a}}{Luffarelli, Mukesh, and Mahmood (2019)}}\label{study-1-based-on-study-1-in-luffarelli-mukesh-and-mahmood-2019}

\emph{Main Questions/Hypotheses}

This study examined whether more (vs. less) descriptive logos elicit
stronger impressions of brand authenticity. It also investigated whether
this effect occurs because such logos are easier to process, testing
ease of processing as a mediator.

\emph{Methods}

We adopted the same stimuli and target sample size (N = 180)\footnote{Target sample sizes follow the original studies, rounded up where an odd total could not be split evenly across conditions (Study 3: 70 rather than 69; Study 6 for GPT-4o-mini (TXT): 160 rather than 154). Because GPT occasionally refuses to answer, the realized sample sizes deviate slightly from the target in some configurations.} as
\citet{luffarelli2019a}. The synthetic consumers mirrored
the demographics of the original human sample recruited via Amazon
Mechanical Turk. The study was a 2 (logo descriptiveness: less vs. more)
× 2 (brand scenarios: basketball equipment manufacturer vs. running shoe
brand) between-subjects design.

AI participants were randomly assigned to one of four conditions. After
being presented with the assigned logo
(\hyperref[figure-b1-visual-stimuli-from-study-1-in-luffarelli-mukesh-and-mahmood-2019]{Figure
B1}), they rated brand authenticity on three nine-point scales
(authentic, trustworthy, and credible; 1 = ``not at all,'' 9 = ``very'';
adapted from \citealp{morhart2015}, and \citealp{napoli2014}), which were
then averaged into a single measure. Next, AI participants assessed ease
of processing on two nine-point scales (1 = ``not at all fluent/not at
all eye-catching,'' 9 = ``very fluent/very eye-catching''; adapted from
\citealp{labroo2008}, and Lee \& Aaker, \citeyear{lee2004}), which were
averaged into a single measure. Next, AI participants provided textual
rationales for their ratings. Finally, they completed the manipulation
check by rating perceived descriptiveness (1 = ``not at all
descriptive,'' 9 = ``very descriptive'').

\subsubsection{\hyperref[figure-b1-visual-stimuli-from-study-1-in-luffarelli-mukesh-and-mahmood-2019]{Figure B1}. Visual stimuli from Study 1 in \texorpdfstring{\citet{luffarelli2019a}}{Luffarelli, Mukesh, and Mahmood (2019)}}\label{figure-b1-visual-stimuli-from-study-1-in-luffarelli-mukesh-and-mahmood-2019}

{
{\footnotesize\begin{longtable}[]{@{}
  >{\centering\arraybackslash}p{(\linewidth - 6\tabcolsep) * \real{0.2406}}
  >{\centering\arraybackslash}p{(\linewidth - 6\tabcolsep) * \real{0.2598}}
  >{\centering\arraybackslash}p{(\linewidth - 6\tabcolsep) * \real{0.2890}}
  >{\centering\arraybackslash}p{(\linewidth - 6\tabcolsep) * \real{0.2106}}@{}}
\toprule\noalign{}
\multicolumn{2}{@{}>{\centering\arraybackslash}p{(\linewidth - 6\tabcolsep) * \real{0.5004} + 2\tabcolsep}}{%
\begin{minipage}[b]{\linewidth}\centering
\textbf{Basketball equipment manufacturer}
\end{minipage}} &
\multicolumn{2}{>{\centering\arraybackslash}p{(\linewidth - 6\tabcolsep) * \real{0.4996} + 2\tabcolsep}@{}}{%
\begin{minipage}[b]{\linewidth}\centering
\textbf{Running shoe brand}
\end{minipage}} \\
\midrule\noalign{}
\endhead
\bottomrule\noalign{}
\endlastfoot
Less

descriptiveness & More

descriptiveness & Less

descriptiveness & More descriptiveness \\
\includegraphics[width=0.88235in,height=0.88235in]{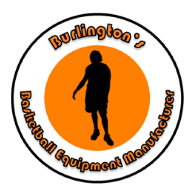}
&
\includegraphics[width=0.87153in,height=0.87153in]{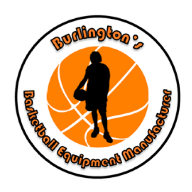}
&
\includegraphics[width=0.79144in,height=0.79144in]{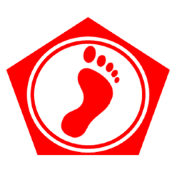}
&
\includegraphics[width=0.79679in,height=0.79679in]{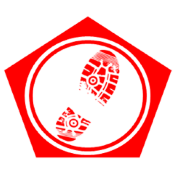} \\
\end{longtable}}
}

\subsection{B.2 Study 2: Based on Study 2A in \texorpdfstring{Lee et al. (\citeyear{lee2017})}{Lee et al. (2017)}}\label{study-2-based-on-study-2a-in-lee-et-al.-2017}

\protect\phantomsection\label{_Toc223972189}{}\emph{Main
Questions/Hypotheses}

This study examined whether temporal distance (distant vs. near future)
influences visual imagery. The specific hypothesis was that AI
participants imagining distant-future events would select images with
lower color saturation (i.e., more black-and-white) to represent their
mental images, compared to those imagining near-future events, who would
select more saturated images.

\emph{Methods}

We adopted the same stimuli and target sample size (N = 108) as Lee et al. (\citeyear{lee2017}). The study used a 2 (temporal distance: distant vs. near)
between-subjects design, with 3 (scenario types: hotel, theater, and
park) as a within-subjects factor.

AI participants were randomly assigned to either the ``distant'' (e.g.,
a year from now) or ``near'' (e.g., a week from now) condition. For each
of the three scenarios (e.g., staying in a hotel), they chose one of
three photo versions, which varied in color saturation (low, medium, or
high; see
\hyperref[figure-b2-visual-stimuli-from-lee-et-al.-2017]{Figure B2}),
presented in a random order. They were prompted to indicate which image
``most closely resembled what they had imagined for each scenario.''
These choices were coded as 1 (low saturation), 2 (medium saturation),
and 3 (high saturation), respectively, for analysis. After that, they
were prompted to provide the rationale for their selection of each
scenario's image. Finally, after responding to each of the three
scenarios, they were prompted to rate perceived temporal distance (1
= ``very near,'' 9 = ``very far'') on a nine-point scale as the
manipulation check.

\subsubsection{\hyperref[figure-b2-visual-stimuli-from-lee-et-al.-2017]{Figure B2}. Visual stimuli from \texorpdfstring{Lee et al. (\citeyear{lee2017})}{Lee et al. (2017)}}\label{figure-b2-visual-stimuli-from-lee-et-al.-2017}

{
{\footnotesize\begin{longtable}[]{@{}
  >{\raggedright\arraybackslash}p{(\linewidth - 4\tabcolsep) * \real{0.3386}}
  >{\raggedright\arraybackslash}p{(\linewidth - 4\tabcolsep) * \real{0.3320}}
  >{\raggedright\arraybackslash}p{(\linewidth - 4\tabcolsep) * \real{0.3293}}@{}}
\toprule\noalign{}
\multicolumn{3}{@{}>{\centering\arraybackslash}p{(\linewidth - 4\tabcolsep) * \real{1.0000} + 4\tabcolsep}@{}}{%
\begin{minipage}[b]{\linewidth}\centering
\textbf{Hotel}

(Saturation level: low  \(\rightarrow\)  medium  \(\rightarrow\)  high)
\end{minipage}} \\
\midrule\noalign{}
\endhead
\bottomrule\noalign{}
\endlastfoot
\includegraphics[width=1.7in,height=1.51111in]{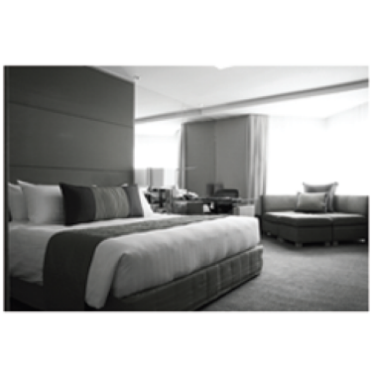}
&
\includegraphics[width=1.79097in,height=1.66875in]{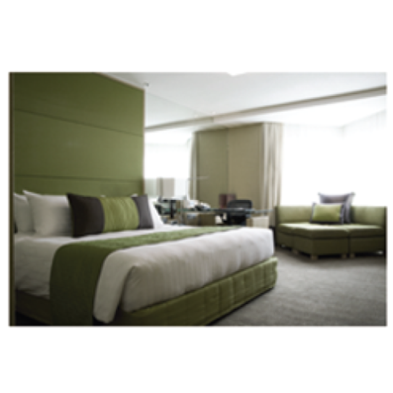}
&
\includegraphics[width=1.77986in,height=1.77986in]{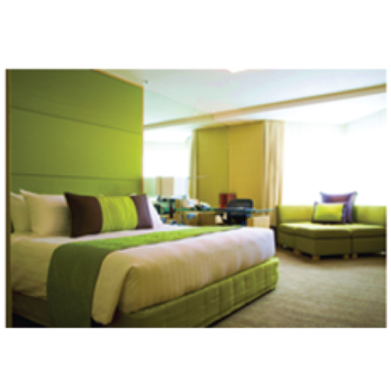} \\
\multicolumn{3}{@{}>{\centering\arraybackslash}p{(\linewidth - 4\tabcolsep) * \real{1.0000} + 4\tabcolsep}@{}}{%
\textbf{Theater}} \\
\includegraphics[width=1.89017in,height=1.89017in]{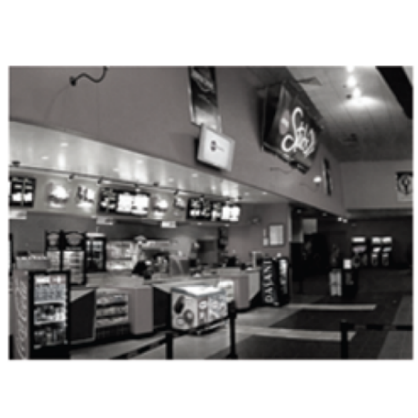}
&
\includegraphics[width=1.87861in,height=1.87861in]{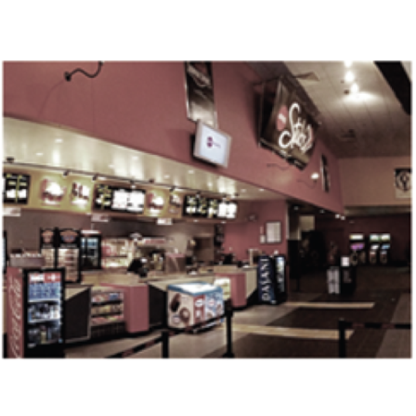}
&
\includegraphics[width=1.84971in,height=1.84971in]{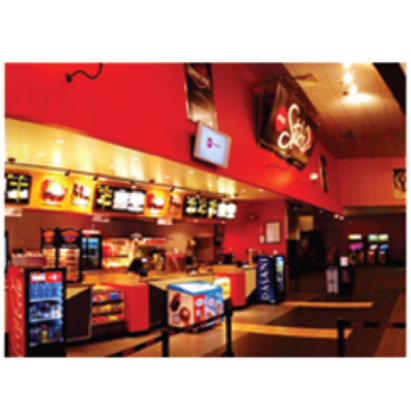} \\
\multicolumn{3}{@{}>{\centering\arraybackslash}p{(\linewidth - 4\tabcolsep) * \real{1.0000} + 4\tabcolsep}@{}}{%
\textbf{Park}} \\
\includegraphics[width=1.87342in,height=1.87342in]{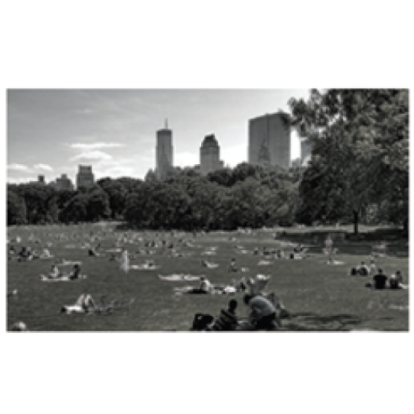}
&
\includegraphics[width=1.89873in,height=1.89873in]{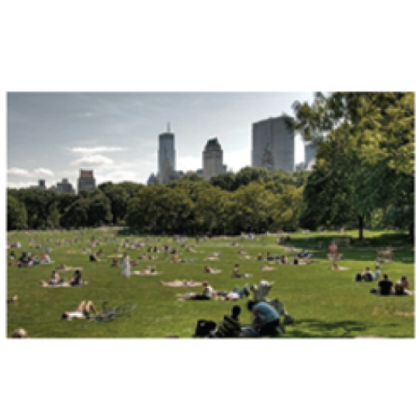}
&
\includegraphics[width=1.91139in,height=1.91139in]{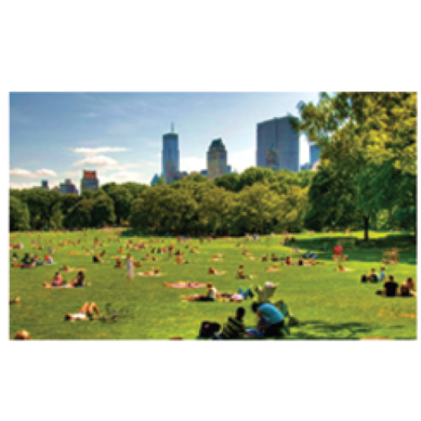} \\
\end{longtable}}
}

\subsection{B.3 Study 3: Based on Study 2 in \texorpdfstring{\citet{jiang2016}}{Jiang et al. (2016)}}\label{study-3-based-on-study-2-in-jiang-et-al.-2016}

\emph{Main Questions/Hypotheses}

This study examined the effect of logo shape (circular vs. angular) on
judgments of product comfort and durability. The original authors
hypothesized that circular logos increase perceived comfortableness,
whereas angular logos increase perceived durability, and that these
effects are mediated by activated associations of perceived softness or
hardness.

\emph{Methods}

We adopted the same visual stimuli and target sample size (N = 69) as in
Study 2 of \citet{jiang2016}. This study used a 2
(logo shape: circular vs. angular) between-subjects design.

AI participants were randomly assigned to view an advertisement for a
sofa featuring either a circular or angular logo
(\hyperref[figure-b3-visual-stimuli-from-jiang-et-al.-2016]{Figure B3}).
Following exposure to the ad, they rated the softness of the sofa's back
cushion, seat cushion, and arms on nine-point scales (1 = ``very hard,''
9 = ``very soft''). They then rated the product's comfortableness (1 =
``not at all comfortable/cozy/cushiony,'' 9 = ``very
comfortable/cozy/cushiony'') and durability (1 = ``not at all
durable/enduring/long-lasting,'' 9 = ``very
durable/enduring/long-lasting''). After that, they provided the textual
rationale for each rating. Finally, they completed a manipulation check
rating perceived circularity (1 = ``not at all circular,'' 9 = ``very
circular'').

\subsubsection{\hyperref[figure-b3-visual-stimuli-from-jiang-et-al.-2016]{Figure B3}. Visual stimuli from \texorpdfstring{\citet{jiang2016}}{Jiang et al. (2016)}}\label{figure-b3-visual-stimuli-from-jiang-et-al.-2016}

{
{\footnotesize\begin{longtable}[]{@{}
  >{\centering\arraybackslash}p{(\linewidth - 2\tabcolsep) * \real{0.4297}}
  >{\centering\arraybackslash}p{(\linewidth - 2\tabcolsep) * \real{0.4297}}@{}}
\toprule\noalign{}
\begin{minipage}[b]{\linewidth}\centering
\textbf{Angular logo}
\end{minipage} & \begin{minipage}[b]{\linewidth}\centering
\textbf{Circular logo}
\end{minipage} \\
\midrule\noalign{}
\endhead
\bottomrule\noalign{}
\endlastfoot
\includegraphics[width=1.97468in,height=1.97468in]{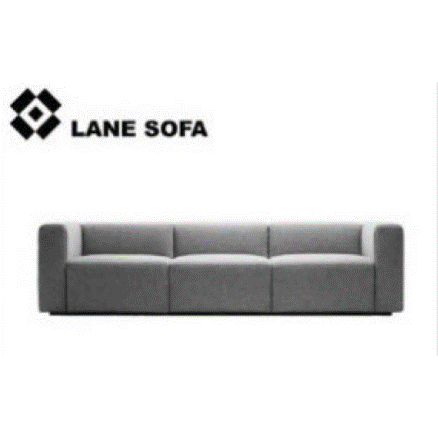}
&
\includegraphics[width=1.9557in,height=1.9557in]{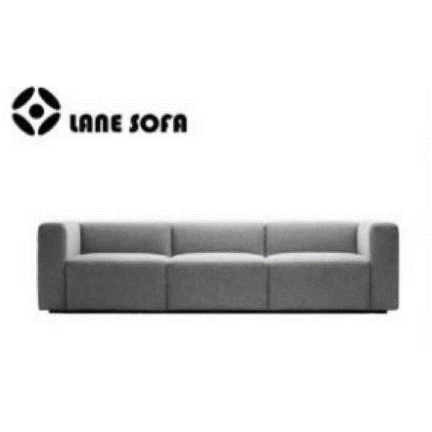} \\
\end{longtable}}
}

\subsection{B.4 Study 4: Based on Study 1B in \texorpdfstring{\citet{luffarelli2019}}{Luffarelli, Stamatogiannakis, and Yang (2019)}}\label{study-4-based-on-study-1b-in-luffarelli-stamatogiannakis-and-yang-2019}

\emph{Main Questions/Hypotheses}

This study examined whether asymmetrical logos are
perceived as more exciting than symmetrical logos. The authors
hypothesized that this effect is mediated by logo-evoked arousal.

\emph{Methods}

We adopted the experimental stimuli and target sample size (N = 220)
from Study 1B of \citet{luffarelli2019}. The study used a 2 (logo symmetry: symmetrical vs.
asymmetrical) × 4 (logo pair) between-subjects design
(\hyperref[figure-b4-visual-stimuli-from-luffarelli-stamatogiannakis-and-yang-2019]{Figure
B4}).

AI participants were randomly assigned to one of two conditions, with
four logo-pair replicates. After being presented with their assigned
logo, they were first prompted to rate the logo on five
excitement-related adjectives using a seven-point scale (trendy, cool,
daring, imaginative, and exciting; 1 = ``not at all,'' 7 = ``very'';
\citealp{aaker1997}). These ratings were averaged
into a measure of logo-evoked perception of excitement. Next, they were
prompted to indicate how they felt when viewing the logo using
seven-point scales (``aroused/unaroused,'' ``stimulated/relaxed,''
``frenzied/sluggish,'' ``jittery/dull,'' ``wide-awake/sleepy'';
\citealp{mehrabian1974}). These
responses were averaged into a single measure of logo-evoked arousal.
After that, they were prompted to provide the rationale for each rating.
Finally, they were prompted to rate perceived asymmetry on a nine-point
scale (1 = ``not at all asymmetrical,'' 9 = ``very asymmetrical'') as the
manipulation check.

\newpage
\subsubsection{\hyperref[figure-b4-visual-stimuli-from-luffarelli-stamatogiannakis-and-yang-2019]{Figure B4}. Visual stimuli from \texorpdfstring{\citet{luffarelli2019}}{Luffarelli, Stamatogiannakis, and Yang (2019)}}\label{figure-b4-visual-stimuli-from-luffarelli-stamatogiannakis-and-yang-2019}

{
{\footnotesize\begin{longtable}[]{@{}
  >{\centering\arraybackslash}p{(\linewidth - 6\tabcolsep) * \real{0.2236}}
  >{\centering\arraybackslash}p{(\linewidth - 6\tabcolsep) * \real{0.2234}}
  >{\centering\arraybackslash}p{(\linewidth - 6\tabcolsep) * \real{0.2196}}
  >{\centering\arraybackslash}p{(\linewidth - 6\tabcolsep) * \real{0.2198}}@{}}
\toprule\noalign{}
\multicolumn{2}{@{}>{\centering\arraybackslash}p{(\linewidth - 6\tabcolsep) * \real{0.4469} + 2\tabcolsep}}{%
\begin{minipage}[b]{\linewidth}\centering
\textbf{Logo pair A}
\end{minipage}} &
\multicolumn{2}{>{\centering\arraybackslash}p{(\linewidth - 6\tabcolsep) * \real{0.4394} + 2\tabcolsep}@{}}{%
\begin{minipage}[b]{\linewidth}\centering
\textbf{Logo pair B}
\end{minipage}} \\
\midrule\noalign{}
\endfirsthead
\bottomrule\noalign{}
\endlastfoot
Symmetrical & Asymmetrical & Symmetrical & Asymmetrical \\*
\includegraphics[width=0.97in,height=0.97in]{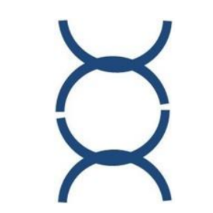}
&
\includegraphics[width=0.96in,height=0.96in]{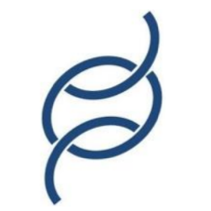}
&
\includegraphics[width=0.89017in,height=0.89017in]{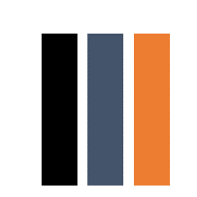}
&
\includegraphics[width=0.94798in,height=0.94798in]{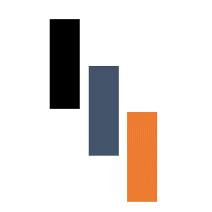} \\*
\multicolumn{2}{@{}>{\centering\arraybackslash}p{(\linewidth - 6\tabcolsep) * \real{0.4469} + 2\tabcolsep}}{%
\textbf{Logo pair C}} &
\multicolumn{2}{>{\centering\arraybackslash}p{(\linewidth - 6\tabcolsep) * \real{0.4394} + 2\tabcolsep}@{}}{%
\textbf{Logo pair D}} \\*
Symmetrical & Asymmetrical & Symmetrical & Asymmetrical \\*
\includegraphics[width=0.98844in,height=0.98844in]{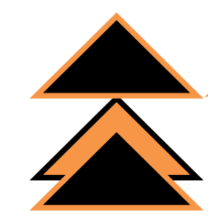}
&
\includegraphics[width=1.03468in,height=1.03468in]{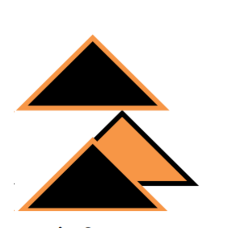}
&
\includegraphics[width=1.14451in,height=1.14451in]{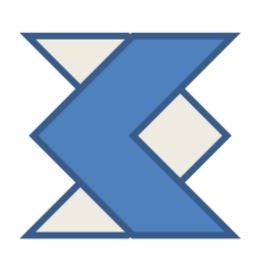}
&
\includegraphics[width=1.12139in,height=1.12139in]{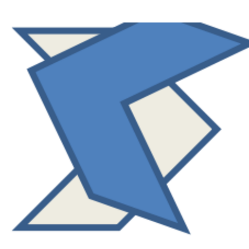} \\
\end{longtable}}
}

\subsection{B.5 Study 5: Based on Study 3 in \texorpdfstring{\citet{hagtvedt2017}}{Hagtvedt and Brasel (2017)}}\label{study-5-based-on-study-3-in-hagtvedt-and-brasel-2017}

\emph{Main Questions/Hypotheses}

This study examined how a product's color saturation (high vs. low)
influences visual estimations of its size. The authors hypothesized that
highly saturated objects increase attention, leading to larger size
estimates. The study also measured arousal and attention as mediating
variables.

\emph{Methods}

We adopted the experimental stimuli and target sample size (N = 80)
from Study 3 of \citet{hagtvedt2017}.
The study used a 2 (color saturation: high vs. low)
between-subjects design.

AI participants were randomly assigned to view an image of a red laptop
with either high or low saturation (see
\hyperref[figure-b5-visual-stimuli-from-hagtvedt-and-brasel-2017]{Figure
B5}). First, AI participants estimated the laptop's screen size on a
scale with 1-inch increments (10 to 20 inches). Next, they rated the
extent to which the laptop was ``attention-getting'' and ``captured
their attention'' (1 = ``not at all,'' 7 = ``definitely''); these items
were averaged into an attention scale. AI participants also reported
arousal and mood using the affect grid \citep{kupor2015,russell1989}. Following these measures, they provided
textual rationales. Finally, they completed a manipulation check rating
perceived saturation (1 = ``very low,'' 7 = ``very high'').

\subsubsection{\hyperref[figure-b5-visual-stimuli-from-hagtvedt-and-brasel-2017]{Figure B5}. Visual stimuli from \texorpdfstring{\citet{hagtvedt2017}}{Hagtvedt and Brasel (2017)}}\label{figure-b5-visual-stimuli-from-hagtvedt-and-brasel-2017}

{
{\footnotesize\begin{longtable}[]{@{}
  >{\raggedright\arraybackslash}p{(\linewidth - 2\tabcolsep) * \real{0.4297}}
  >{\raggedright\arraybackslash}p{(\linewidth - 2\tabcolsep) * \real{0.4326}}@{}}
\toprule\noalign{}
\begin{minipage}[b]{\linewidth}\centering
\textbf{Low saturation}
\end{minipage} & \begin{minipage}[b]{\linewidth}\centering
\textbf{High saturation}
\end{minipage} \\
\midrule\noalign{}
\endhead
\bottomrule\noalign{}
\endlastfoot
\includegraphics[width=1.91in,height=1.91in]{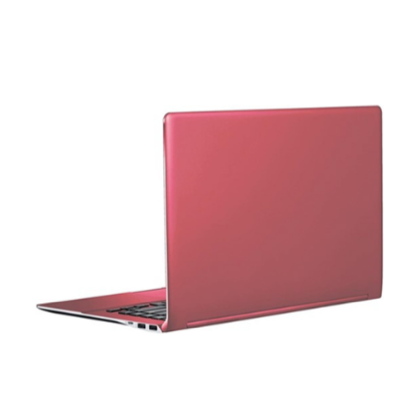}
&
\includegraphics[width=1.86096in,height=1.86096in]{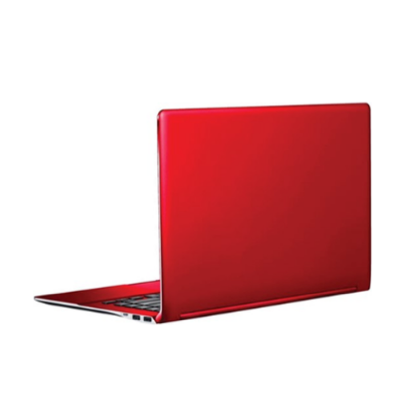} \\
\end{longtable}}
}

\subsection{B.6 Study 6: Based on Study 2A in \texorpdfstring{\citet{rahinel2016}}{Rahinel and Nelson (2016)}}\label{study-6-based-on-study-2a-in-rahinel-and-nelson-2016}

\emph{Main Questions/Hypotheses}

This study examined the interactive effect of logo stability (unstable
vs. stable) and product category on perceived utility. \citet{rahinel2016} hypothesized that an unstable logo (e.g., a square rotated
on its vertex) signals a need for safety, thereby increasing the
perceived utility of safety-oriented products. This effect was expected
to be attenuated or absent in neutral product categories.

\emph{Methods}

We adopted the experimental stimuli and target sample size (N = 154)
from Study 2A of \citet{rahinel2016}. The study used a 2 (logo
stability: unstable vs. stable) × 2 (category type: safety-oriented vs.
neutral) × 6 (product replicate) mixed design.

AI participants were randomly assigned to view a logo for an online
company, manipulated to appear either stable (square on base) or
unstable (square on vertex; see
\hyperref[figure-b6-visual-stimuli-from-rahinel-and-nelson-2016]{Figure
B6}). AI participants then rated the perceived utility of six products
on three nine-point items (``how important the product is,'' ``how
useful the product is,'' ``how likely the product is to be used''; 1 =
``not at all,'' 9 = ``very''), which were averaged into a composite
index. Depending on the condition, the products were either
safety-oriented (e.g., ``security system'') or neutral (e.g., ``lawn
care''). Following these ratings, AI participants provided textual
rationales. Finally, they completed a manipulation check rating
perceived instability (1 = ``not at all,'' 9 = ``very'').

\subsubsection{\hyperref[figure-b6-visual-stimuli-from-rahinel-and-nelson-2016]{Figure B6}. Visual stimuli from \texorpdfstring{\citet{rahinel2016}}{Rahinel and Nelson (2016)}}\label{figure-b6-visual-stimuli-from-rahinel-and-nelson-2016}

{
{\footnotesize\begin{longtable}[]{@{}
  >{\centering\arraybackslash}p{(\linewidth - 2\tabcolsep) * \real{0.4297}}
  >{\centering\arraybackslash}p{(\linewidth - 2\tabcolsep) * \real{0.4326}}@{}}
\toprule\noalign{}
\begin{minipage}[b]{\linewidth}\centering
\textbf{Stable logo}
\end{minipage} & \begin{minipage}[b]{\linewidth}\centering
\textbf{Unstable logo}
\end{minipage} \\
\midrule\noalign{}
\endhead
\bottomrule\noalign{}
\endlastfoot
\includegraphics[width=1.9in,height=1.84028in]{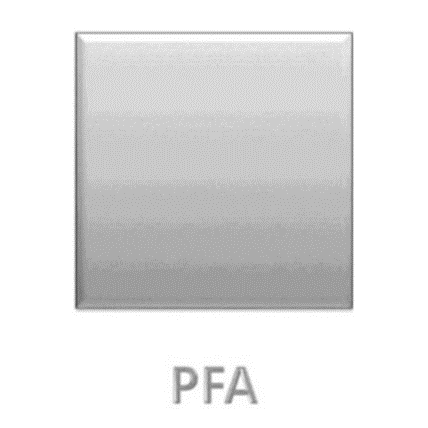}
&
\includegraphics[width=1.93056in,height=1.84091in]{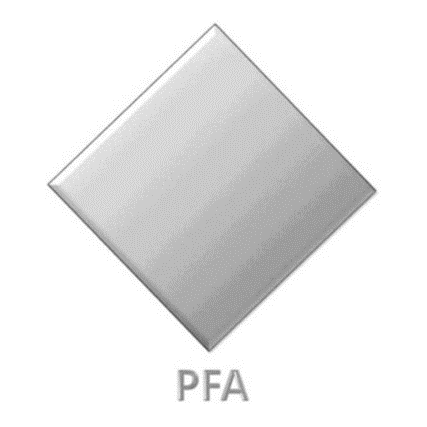} \\
\end{longtable}}
}

\clearpage
\section{Web Appendix \texorpdfstring{C}{C}: Heterogeneity analysis by personas}\label{web-appendix-c-heterogeneity}

{This appendix reports whether the synthetic consumers' judgments vary with their assigned personas. Because the four GPT format $\times$ model configurations (4o-T, 4o-J, 5.4-T, 5.4-J) are the focus of the main text, all tables report these four configurations. C.1 reports the realized demographic composition. C.2 fits regression models of the outcome variable(s) from each study with age, gender in assigned personas, and their interactions with the treatment added to the study's specification, and reports the four demographic coefficients $\gamma_1$--$\gamma_4$. Across studies, the age and gender main effects and their treatment interactions are almost all non-significant, and the few that reach $p<.05$ show no consistent pattern across configurations, except a small negative age effect on excitement that appears in all four configurations in Study 4 (\hyperref[table-c5-combined-study4]{Table C5}). The treatment effects are therefore essentially homogeneous across the assigned demographics.\par}

\subsection{C.1 Demographic composition}\label{c.1-demographic-composition}

\subsubsection{\hyperref[table-c1-demographics]{Table C1}. Demographic distribution of each study}\label{table-c1-demographics}

{
{\footnotesize\begin{longtable}[]{@{}llccc@{}}
\toprule
Study & Config & Age $M$ (SD) & Age range & \% female \\
\midrule\endhead
\bottomrule\endlastfoot
1 & 4o-T & 41.5 (13.9) & 18--65 & 49.3 \\
1 & 4o-J & 41.3 (13.8) & 18--65 & 50.2 \\
1 & 5.4-T & 42.2 (14.3) & 18--65 & 47.8 \\
1 & 5.4-J & 41.2 (14.2) & 18--65 & 51.3 \\
2 & 4o-T & 40.4 (13.7) & 18--65 & 47.2 \\
2 & 4o-J & 40.8 (14.1) & 18--65 & 50.6 \\
2 & 5.4-T & 41.9 (13.7) & 18--65 & 50.6 \\
2 & 5.4-J & 42.1 (14.0) & 18--65 & 48.5 \\
3 & 4o-T & 19.9 (1.4) & 18--22 & 53.3 \\
3 & 4o-J & 20.1 (1.4) & 18--22 & 49.0 \\
3 & 5.4-T & 20.1 (1.4) & 18--22 & 46.2 \\
3 & 5.4-J & 20.0 (1.4) & 18--22 & 47.1 \\
4 & 4o-T & 42.2 (13.8) & 18--65 & 48.5 \\
4 & 4o-J & 41.2 (14.1) & 18--65 & 52.0 \\
4 & 5.4-T & 41.9 (13.8) & 18--65 & 50.5 \\
4 & 5.4-J & 43.3 (13.8) & 18--65 & 52.4 \\
5 & 4o-T & 21.4 (2.3) & 18--25 & 51.7 \\
5 & 4o-J & 21.5 (2.3) & 18--25 & 43.8 \\
5 & 5.4-T & 21.4 (2.4) & 18--25 & 49.6 \\
5 & 5.4-J & 21.3 (2.3) & 18--25 & 49.6 \\
6 & 4o-T & 20.4 (1.8) & 18--23 & 46.8 \\
6 & 4o-J & 20.5 (1.7) & 18--23 & 46.5 \\
6 & 5.4-T & 20.5 (1.7) & 18--23 & 48.9 \\
6 & 5.4-J & 20.5 (1.7) & 18--23 & 46.5 \\
\end{longtable}}
{\setstretch{1}\footnotesize\raggedright\emph{Note}: Config labels: 5.4-J/5.4-T $=$ GPT-5.4-mini with JSON/TXT input; 4o-J/4o-T $=$ GPT-4o-mini with JSON/TXT input. The samples are about 50\% female by design. Distributions are pooled across the three interventions, which drew new personas from the same sampling rule.\par}
}

\subsection{C.2 Demographic effects and treatment-effect heterogeneity}\label{c.2-demographic-heterogeneity}

{\hyperref[table-c2-combined-study1]{Tables C2--C7} report, per study, the combined-model demographic coefficients $\gamma_1$--$\gamma_4$ (age main, treatment$\times$age, gender main, treatment$\times$gender) at zero-shot, with significant effects in bold; (DV) marks the study's dependent variable. Subscript $i$ indexes AI participants. Studies 2 and 6 include within-subjects factors, so those models add a participant random intercept $u_i$.\par}

\subsubsection{\hyperref[table-c2-combined-study1]{Table C2}. Study 1: Logo descriptiveness and brand authenticity (DV)}\label{table-c2-combined-study1}

{
{\footnotesize\setlength{\tabcolsep}{3pt}\begin{longtable}[]{@{}lcccccccc@{}}
\toprule
& \multicolumn{2}{c}{$\gamma_1$ Age} & \multicolumn{2}{c}{$\gamma_2$ Age$\times$T} & \multicolumn{2}{c}{$\gamma_3$ Gender} & \multicolumn{2}{c}{$\gamma_4$ Gen$\times$T} \\
\cmidrule(lr){2-3}\cmidrule(lr){4-5}\cmidrule(lr){6-7}\cmidrule(lr){8-9}
Config & $\gamma$ (SE) & $\eta^2$ & $\gamma$ (SE) & $\eta^2$ & $\gamma$ (SE) & $\eta^2$ & $\gamma$ (SE) & $\eta^2$ \\
\midrule\endhead
\bottomrule\endlastfoot
4o-T & $-$.005 (.002) & .044 & .001 (.004) & .000 & $-$.004 (.051) & .000 & $-$.007 (.102) & .000 \\
4o-J & .001 (.002) & .002 & $-$.002 (.004) & .001 & .032 (.053) & .002 & $-$.036 (.106) & .001 \\
5.4-T & .005 (.003) & .024 & .006 (.005) & .008 & .024 (.078) & .001 & $-$.074 (.155) & .001 \\
5.4-J & .005 (.003) & .027 & .002 (.005) & .001 & $-$.080 (.070) & .008 & $-$.134 (.140) & .005 \\
\end{longtable}}
{\setstretch{1}\footnotesize\raggedright\emph{Note}: For age, $\gamma$ is the change per extra year from the mean; for gender, $\gamma$ is the female-minus-male difference (Male $=$ 0, Female $=$ 1). \emph{Replicate} is a binary indicator that denotes the two stimulus groups as in the original study. $\text{Authenticity}_i=\beta_0+\beta_1\text{Desc}_i+\beta_2\text{Replicate}_i+\beta_3\,\text{Desc}_i{\times}\text{Replicate}_i+\gamma_1\text{Age}_i+\gamma_2\,\text{Desc}_i{\times}\text{Age}_i+\gamma_3\text{Gender}_i+\gamma_4\,\text{Desc}_i{\times}\text{Gender}_i+\varepsilon_i$.\par}
}

\subsubsection{\hyperref[table-c3-combined-study2]{Table C3}. Study 2: Temporal distance and color saturation (DV)}\label{table-c3-combined-study2}

{
{\footnotesize\setlength{\tabcolsep}{3pt}\begin{longtable}[]{@{}lcccccccc@{}}
\toprule
& \multicolumn{2}{c}{$\gamma_1$ Age} & \multicolumn{2}{c}{$\gamma_2$ Age$\times$T} & \multicolumn{2}{c}{$\gamma_3$ Gender} & \multicolumn{2}{c}{$\gamma_4$ Gen$\times$T} \\
\cmidrule(lr){2-3}\cmidrule(lr){4-5}\cmidrule(lr){6-7}\cmidrule(lr){8-9}
Config & $\gamma$ (SE) & $\eta^2$ & $\gamma$ (SE) & $\eta^2$ & $\gamma$ (SE) & $\eta^2$ & $\gamma$ (SE) & $\eta^2$ \\
\midrule\endhead
\bottomrule\endlastfoot
4o-T & $-$.000 (.001) & .000 & $-$.004 (.003) & .006 & .016 (.041) & .001 & $-$.039 (.081) & .001 \\
4o-J & $-$.002 (.001) & .020 & .001 (.003) & .001 & .068 (.041) & .026 & $-$.012 (.082) & .000 \\
5.4-T & .001 (.002) & .001 & .004 (.004) & .003 & .091 (.054) & .009 & $-$.017 (.107) & .000 \\
5.4-J & $-$.002 (.002) & .003 & \textbf{.008 (.004)} & .014 & $-$.043 (.051) & .002 & $-$.010 (.101) & .000 \\
\end{longtable}}
{\setstretch{1}\footnotesize\raggedright\emph{Note}: For age, $\gamma$ is the change per extra year from the mean; for gender, $\gamma$ is the female-minus-male difference (Male $=$ 0, Female $=$ 1). \emph{Scene}$_s$ is the image scene (hotel, theater, or park), and $u_i$ is a participant random intercept. Partial $\eta^2$ for these mixed models is $F\,\mathrm{df}_1/(F\,\mathrm{df}_1+\mathrm{df}_2)$ from the Type-III $F$-test with Satterthwaite degrees of freedom. $\text{SelSat}_{is}=\beta_0+\beta_1\text{Temporal}_i+\beta_2\text{Scene}_s+\beta_3\,\text{Temporal}_i{\times}\text{Scene}_s+\gamma_1\text{Age}_i+\gamma_2\,\text{Temporal}_i{\times}\text{Age}_i+\gamma_3\text{Gender}_i+\gamma_4\,\text{Temporal}_i{\times}\text{Gender}_i+u_i+\varepsilon_{is}$.\par}
}

\subsubsection{\hyperref[table-c4-combined-study3]{Table C4}. Study 3: Logo shape and product attributes (DV)}\label{table-c4-combined-study3}

{
{\footnotesize\setlength{\tabcolsep}{3pt}\begin{longtable}[]{@{}llcccccccc@{}}
\toprule
& & \multicolumn{2}{c}{$\gamma_1$ Age} & \multicolumn{2}{c}{$\gamma_2$ Age$\times$T} & \multicolumn{2}{c}{$\gamma_3$ Gender} & \multicolumn{2}{c}{$\gamma_4$ Gen$\times$T} \\
\cmidrule(lr){3-4}\cmidrule(lr){5-6}\cmidrule(lr){7-8}\cmidrule(lr){9-10}
Config & DV & $\gamma$ (SE) & $\eta^2$ & $\gamma$ (SE) & $\eta^2$ & $\gamma$ (SE) & $\eta^2$ & $\gamma$ (SE) & $\eta^2$ \\
\midrule\endhead
\bottomrule\endlastfoot
4o-T & Comfort & .007 (.044) & .000 & $-$.076 (.087) & .012 & .150 (.125) & .022 & .368 (.251) & .033 \\
4o-T & Durability & $-$.046 (.068) & .007 & $-$.068 (.137) & .004 & $-$.211 (.196) & .018 & .348 (.392) & .012 \\
4o-J & Comfort & $-$.037 (.051) & .008 & .019 (.103) & .001 & \textbf{.440 (.139)} & .136 & .451 (.278) & .039 \\
4o-J & Durability & .130 (.072) & .048 & $-$.186 (.144) & .025 & $-$.375 (.195) & .054 & .465 (.391) & .022 \\
5.4-T & Comfort & $-$.046 (.059) & .009 & .059 (.119) & .004 & .053 (.145) & .002 & .061 (.290) & .001 \\
5.4-T & Durability & .016 (.012) & .026 & $-$.032 (.024) & .026 & .016 (.030) & .005 & $-$.032 (.060) & .005 \\
5.4-J & Comfort & .003 (.050) & .000 & $-$.009 (.100) & .000 & \textbf{.288 (.141)} & .061 & $-$.385 (.283) & .028 \\
5.4-J & Durability & $-$.007 (.010) & .007 & $-$.014 (.021) & .007 & .036 (.030) & .023 & .072 (.059) & .023 \\
\end{longtable}}
{\setstretch{1}\footnotesize\raggedright\emph{Note}: For age, $\gamma$ is the change per extra year from the mean; for gender, $\gamma$ is the female-minus-male difference (Male $=$ 0, Female $=$ 1). $\text{DV}_i=\beta_0+\beta_1\text{Shape}_i+\gamma_1\text{Age}_i+\gamma_2\,\text{Shape}_i{\times}\text{Age}_i+\gamma_3\text{Gender}_i+\gamma_4\,\text{Shape}_i{\times}\text{Gender}_i+\varepsilon_i$ (DV $=$ Comfort or Durability).\par}
}

\subsubsection{\hyperref[table-c5-combined-study4]{Table C5}. Study 4: Logo asymmetry and brand excitement (DV)}\label{table-c5-combined-study4}

{
{\footnotesize\setlength{\tabcolsep}{3pt}\begin{longtable}[]{@{}lcccccccc@{}}
\toprule
& \multicolumn{2}{c}{$\gamma_1$ Age} & \multicolumn{2}{c}{$\gamma_2$ Age$\times$T} & \multicolumn{2}{c}{$\gamma_3$ Gender} & \multicolumn{2}{c}{$\gamma_4$ Gen$\times$T} \\
\cmidrule(lr){2-3}\cmidrule(lr){4-5}\cmidrule(lr){6-7}\cmidrule(lr){8-9}
Config & $\gamma$ (SE) & $\eta^2$ & $\gamma$ (SE) & $\eta^2$ & $\gamma$ (SE) & $\eta^2$ & $\gamma$ (SE) & $\eta^2$ \\
\midrule\endhead
\bottomrule\endlastfoot
4o-T & \textbf{$-$.006 (.001)} & .088 & .003 (.003) & .007 & .017 (.038) & .001 & $-$.072 (.076) & .004 \\
4o-J & \textbf{$-$.004 (.001)} & .049 & .005 (.003) & .017 & $-$.013 (.038) & .001 & $-$.091 (.077) & .007 \\
5.4-T & \textbf{$-$.008 (.002)} & .061 & .002 (.004) & .001 & $-$.034 (.057) & .002 & $-$.022 (.114) & .000 \\
5.4-J & \textbf{$-$.007 (.002)} & .069 & $-$.001 (.004) & .000 & .065 (.052) & .008 & $-$.031 (.103) & .000 \\
\end{longtable}}
{\setstretch{1}\footnotesize\raggedright\emph{Note}: For age, $\gamma$ is the change per extra year from the mean; for gender, $\gamma$ is the female-minus-male difference (Male $=$ 0, Female $=$ 1). \emph{Pair} is the logo pair. $\text{Excitement}_i=\beta_0+\beta_1\text{Asym}_i+\beta_2\text{Pair}_i+\beta_3\,\text{Asym}_i{\times}\text{Pair}_i+\gamma_1\text{Age}_i+\gamma_2\,\text{Asym}_i{\times}\text{Age}_i+\gamma_3\text{Gender}_i+\gamma_4\,\text{Asym}_i{\times}\text{Gender}_i+\varepsilon_i$.\par}
}

\subsubsection{\hyperref[table-c6-combined-study5]{Table C6}. Study 5: Color saturation and estimated size (DV)}\label{table-c6-combined-study5}

{
{\footnotesize\setlength{\tabcolsep}{3pt}\begin{longtable}[]{@{}lcccccccc@{}}
\toprule
& \multicolumn{2}{c}{$\gamma_1$ Age} & \multicolumn{2}{c}{$\gamma_2$ Age$\times$T} & \multicolumn{2}{c}{$\gamma_3$ Gender} & \multicolumn{2}{c}{$\gamma_4$ Gen$\times$T} \\
\cmidrule(lr){2-3}\cmidrule(lr){4-5}\cmidrule(lr){6-7}\cmidrule(lr){8-9}
Config & $\gamma$ (SE) & $\eta^2$ & $\gamma$ (SE) & $\eta^2$ & $\gamma$ (SE) & $\eta^2$ & $\gamma$ (SE) & $\eta^2$ \\
\midrule\endhead
\bottomrule\endlastfoot
4o-T & .022 (.024) & .012 & .012 (.049) & .001 & $-$.198 (.109) & .045 & .384 (.218) & .042 \\
4o-J & $-$.000 (.039) & .000 & .063 (.079) & .010 & .093 (.179) & .004 & .229 (.359) & .006 \\
5.4-T & .022 (.030) & .007 & $-$.029 (.059) & .003 & $-$.048 (.138) & .002 & .331 (.276) & .019 \\
5.4-J & $-$.050 (.036) & .026 & .001 (.072) & .000 & $-$.129 (.160) & .009 & $-$.099 (.319) & .001 \\
\end{longtable}}
{\setstretch{1}\footnotesize\raggedright\emph{Note}: For age, $\gamma$ is the change per extra year from the mean; for gender, $\gamma$ is the female-minus-male difference (Male $=$ 0, Female $=$ 1). $\text{Size}_i=\beta_0+\beta_1\text{Saturation}_i+\gamma_1\text{Age}_i+\gamma_2\,\text{Saturation}_i{\times}\text{Age}_i+\gamma_3\text{Gender}_i+\gamma_4\,\text{Saturation}_i{\times}\text{Gender}_i+\varepsilon_i$.\par}
}

\subsubsection{\hyperref[table-c7-combined-study6]{Table C7}. Study 6: Logo stability and utility (DV)}\label{table-c7-combined-study6}

{
{\footnotesize\setlength{\tabcolsep}{3pt}\begin{longtable}[]{@{}lcccccccc@{}}
\toprule
& \multicolumn{2}{c}{$\gamma_1$ Age} & \multicolumn{2}{c}{$\gamma_2$ Age$\times$T} & \multicolumn{2}{c}{$\gamma_3$ Gender} & \multicolumn{2}{c}{$\gamma_4$ Gen$\times$T} \\
\cmidrule(lr){2-3}\cmidrule(lr){4-5}\cmidrule(lr){6-7}\cmidrule(lr){8-9}
Config & $\gamma$ (SE) & $\eta^2$ & $\gamma$ (SE) & $\eta^2$ & $\gamma$ (SE) & $\eta^2$ & $\gamma$ (SE) & $\eta^2$ \\
\midrule\endhead
\bottomrule\endlastfoot
4o-T & .006 (.008) & .003 & $-$.000 (.016) & .000 & .025 (.030) & .005 & .086 (.059) & .014 \\
4o-J & $-$.003 (.009) & .001 & $-$.018 (.019) & .006 & .037 (.031) & .009 & .032 (.062) & .002 \\
5.4-T & $-$.000 (.016) & .000 & .060 (.032) & .023 & .061 (.055) & .008 & $-$.033 (.111) & .001 \\
5.4-J & .003 (.020) & .000 & $-$.027 (.040) & .003 & .026 (.070) & .001 & $-$.133 (.140) & .006 \\
\end{longtable}}
{\setstretch{1}\footnotesize\raggedright\emph{Note}: For age, $\gamma$ is the change per extra year from the mean; for gender, $\gamma$ is the female-minus-male difference (Male $=$ 0, Female $=$ 1). \emph{Replicate} is the repeated stimulus set ($j$ indexes the rated products), and $u_i$ is a participant random intercept. Partial $\eta^2$ for these mixed models is $F\,\mathrm{df}_1/(F\,\mathrm{df}_1+\mathrm{df}_2)$ from the Type-III $F$-test with Satterthwaite degrees of freedom. $\text{Utility}_{ij}=\beta_0+\text{Stability}_i{\times}\text{ProductType}_j{\times}\text{Replicate}_j\text{ (full factorial)}+\gamma_1\text{Age}_i+\gamma_2\,\text{Stability}_i{\times}\text{Age}_i+\gamma_3\text{Gender}_i+\gamma_4\,\text{Stability}_i{\times}\text{Gender}_i+u_i+\varepsilon_{ij}$.\par}
}

\clearpage
\section{Web Appendix \texorpdfstring{D}{D}: BERTopic and LLM-assisted topic models}\label{web-appendix-d-bertopic-and-llm-assisted-topic-models}

\subsection{D.1 Natural language pipeline}\label{natural-language-pipeline}

\begin{enumerate}
\def\labelenumi{\arabic{enumi}.}
\item
  Semantic Embedding: We first converted each rationale (pre-processed
  to remove conversational artifacts) into a high-dimensional vector
  using a sentence-transformer. This embedding captures the
  rationale's contextual meaning, not just its keywords.
  For instance, in Study 3, this model understands that ``circular,''
  ``curved,'' and ``smoothly contoured'' are semantically similar
  concepts, clustering them even if they do not share words.
\item
  Topic Clustering: We followed the standard BERTopic pipeline \citep{grootendorst2022}, which
  applies UMAP for dimensionality reduction and HDBSCAN for
  clustering.\footnote{HDBSCAN (Hierarchical Density-Based Spatial
    Clustering of Applications with Noise) is the default clustering
    algorithm in BERTopic. Unlike algorithms like K-Means, it does not
    require the number of topics (clusters) to be pre-specified. It
    determines the number of clusters (topics) by identifying dense
    regions of documents in the embedding space and classifies the
    sparser data points as outliers (i.e., not belonging to a specific
    topic).} This process groups rationales into semantically coherent
  and conceptually distinct topics. Consistent with best practices for a
  smaller corpus, we set the minimum topic size to 5 and the number of
  neighbors to 15 to capture granular topics. We also varied these key
  hyperparameters for robustness checks.
\item
  Topic Representation: We identified keywords for each topic using a
  class-based metric that selects words that are both frequent within a
  topic and rare across other topics.\footnote{The metric, c-TF-IDF
    (Class-based Term Frequency-Inverse Document Frequency), is a
    version of TF-IDF adapted for topic modeling. It treats all
    documents belonging to one topic as one combined ``class.'' Then it
    calculates a score based on how frequent a word is in its parent
    topic, while also penalizing words that appear frequently in other
    topics. A high score thus indicates a word is both highly
    descriptive of and unique to the topic.} For example, Topic 0 in
  one model's Study 1 was represented by: ``context (.621), professional
  (.583), credibility (.581), logo (.579)....'' In addition to these
  keywords, a byproduct of BERTopic's clustering is the identification
  of the most representative documents for each topic (i.e., those
  closest to the topic's cluster centroid). We used this feature to
  enhance the objectivity and scalability of the labeling procedure, using
  an LLM to assist with topic labeling. For each topic, we prompted
  GPT-4o-mini with the three most representative documents and asked it
  to generate a concise, descriptive label. For example, the
  LLM-generated label for the topic above highlighted a lack of
  ``additional context or branding elements'' as a factor influencing
  ``credibility,'' an insight that emerged directly from the data-driven
  clustering.
\end{enumerate}

\subsection{D.2 Example of prompt for GPT-assisted topics}\label{example-of-prompt-for-gpt-assisted-topics}

We used the following prompt with GPT-4o-mini to assist with topic
description based on the representative documents selected by BERTopic. ``I
have a dataset of logo design evaluations, containing rationales about
different features. These evaluations consist of free-text responses,
ranging from short comments to detailed analyses. Please identify the
single main topic mentioned in these comments. Return a topic name and
topic description. The topic name should be short, but descriptive. The
topic description should be a complete sentence. A good topic
description looks like this: \textquotesingle There are concerns about
authenticity because the logo designs are overly
minimal.\textquotesingle{} Return the topic name and description as a
Python dictionary with a single key-value pair like this:
\{\{\textquotesingle topic\_name\textquotesingle:
\textquotesingle\textless topicName\textgreater\textquotesingle,
\textquotesingle topic\_description\textquotesingle:
\textquotesingle\textless topicDescription\textgreater\textquotesingle\}\}
If you cannot find a good topic label, just say, \textquotesingle No
topic identified.\textquotesingle{}''

\clearpage
\section{Web Appendix \texorpdfstring{E}{E}: Prompt intervention templates and results after interventions}\label{web-appendix-e-prompt-intervention-templates-and-results-after-interventions}

\subsection{E.1 Prompt templates}\label{prompt-templates}

\emph{Prompt with Conceptual Evidence}

``You are a
\{demographics{[}\textquotesingle age\textquotesingle{]}\}-year-old
\{demographics{[}\textquotesingle gender\textquotesingle{]}\}
participant who lives in the
\{demographics{[}\textquotesingle location\textquotesingle{]}\}. Please
take a moment to look at the image carefully and wait for my prompt.
Before answering the survey questions below, please read and internalize
the following background: {[}The abstract of the study is inserted
here.{]} Once you have fully processed that background, please proceed
to answer the following survey questions based on it.''

\emph{Prompt with Empirical Evidence}

``You are a
\{demographics{[}\textquotesingle age\textquotesingle{]}\}-year-old
\{demographics{[}\textquotesingle gender\textquotesingle{]}\}
participant who lives in the
\{demographics{[}\textquotesingle location\textquotesingle{]}\}. Please
take a moment to look at the image carefully and wait for my prompt.
Before answering the survey questions below, please read and internalize
the following background: {[}Results and statistics from the relevant
study are inserted here.{]} Once you have fully processed that
background, please proceed to answer the following survey questions
based on it.''

\subsection{E.2 Detailed results by study after prompt interventions}\label{detailed-results-by-study-after-prompt-interventions}

This section reports the main-effect results after the prompt interventions for all four GPT format $\times$ model configurations (4o-T, 4o-J, 5.4-T, 5.4-J; labels as in \hyperref[table-c1-demographics]{Table C1}, \hyperref[web-appendix-c-heterogeneity]{Web Appendix C}), under No Intervention, the prompt with conceptual evidence, and the prompt with empirical evidence. Dir($p$) gives the main-effect direction and its $p$-value ($+$/$-$ $=$ direction consistent/inconsistent with the human finding), with bold marking $p<.05$; Med.\ gives the mediation verdict (full or partial) for the four studies with a hypothesized mediator (\hyperref[e-study-1]{Tables E1}, E3, E4, and E5); a dot means that no mediation is established, either because the main effect is not significant, so the test is not interpreted, or because the indirect effect is not significant; and (DV) marks the study's dependent variable. Unless noted otherwise, each Dir($p$) comes from a one-way ANOVA on the study's dependent variable, $\eta^2$ is the effect size from that test, and each configuration's sample size matches the original study's (\hyperref[web-appendix-b-description-of-studies-and-visual-stimuli]{Web Appendix B}). \hyperref[e-study-1]{Tables E1--E6} report all four configurations; \hyperref[e-study-7]{Table E7} reports the case-study deployment, which uses GPT-5.4-mini (TXT) only. The tables in \hyperref[web-appendix-f-robustness-of-findings]{Web Appendix F} follow the same conventions.

\begin{landscape}
\subsubsection{\hyperref[e-study-1]{Table E1}. Study 1: Logo descriptiveness and brand authenticity (DV)}\label{e-study-1}

{
{\footnotesize\setlength{\tabcolsep}{2pt}\begin{longtable}[]{@{}lccccccccccccccc@{}}
\toprule
 & \multicolumn{5}{c}{No Intervention} & \multicolumn{5}{c}{Conceptual} & \multicolumn{5}{c}{Empirical} \\
\cmidrule(lr){2-6}\cmidrule(lr){7-11}\cmidrule(lr){12-16}
Config & More $M$(SD) & Less $M$(SD) & Dir($p$) & $\eta^2$ & Med. & More $M$(SD) & Less $M$(SD) & Dir($p$) & $\eta^2$ & Med. & More $M$(SD) & Less $M$(SD) & Dir($p$) & $\eta^2$ & Med. \\
\midrule\endhead
4o-T & 6.424(.354) & 6.198(.405) & \textbf{$+$ ($<$.001)} & .082 & . & 6.622(.334) & 6.447(.303) & \textbf{$+$ ($<$.001)} & .071 & . & 6.562(.269) & 6.295(.504) & \textbf{$+$ ($<$.001)} & .100 & . \\
4o-J & 6.470(.227) & 6.209(.431) & \textbf{$+$ ($<$.001)} & .125 & partial & 6.674(.299) & 6.457(.162) & \textbf{$+$ ($<$.001)} & .172 & . & 6.610(.196) & 6.345(.359) & \textbf{$+$ ($<$.001)} & .178 & . \\
5.4-T & 4.293(.970) & 4.356(.507) & $-$ (.586) & .002 & . & 6.485(.618) & 6.419(.478) & $+$ (.420) & .004 & . & 6.856(.669) & 6.519(.839) & \textbf{$+$ (.003)} & .048 & partial \\
5.4-J & 4.370(1.101) & 4.404(.483) & $-$ (.793) & .000 & . & 6.648(.660) & 6.585(.483) & $+$ (.466) & .003 & . & 6.630(.709) & 6.333(.865) & \textbf{$+$ (.013)} & .034 & partial \\
\bottomrule
\end{longtable}}
}
{\footnotesize\setstretch{1}\raggedright\emph{Note}: Study 1 (DV: brand authenticity); More/Less $=$ more- vs.\ less-descriptive logo condition. Rows are the four GPT configurations; cells report condition means (SD).\par}
\vspace{2em}

\subsubsection{\hyperref[e-study-2]{Table E2}. Study 2: Temporal distance and color saturation (DV)}\label{e-study-2}

{
{\footnotesize\setlength{\tabcolsep}{2pt}\begin{longtable}[]{@{}lcccccccccccc@{}}
\toprule
 & \multicolumn{4}{c}{No Intervention} & \multicolumn{4}{c}{Conceptual} & \multicolumn{4}{c}{Empirical} \\
\cmidrule(lr){2-5}\cmidrule(lr){6-9}\cmidrule(lr){10-13}
Config & Near $M$(SD) & Dist. $M$(SD) & Dir($p$) & $\eta^2$ & Near $M$(SD) & Dist. $M$(SD) & Dir($p$) & $\eta^2$ & Near $M$(SD) & Dist. $M$(SD) & Dir($p$) & $\eta^2$ \\
\midrule\endhead
4o-T & 2.815(.211) & 2.796(.188) & $+$ (.631) & .002 & 1.642(.488) & 1.259(.334) & \textbf{$+$ ($<$.001)} & .176 & 2.975(.088) & 2.877(.245) & \textbf{$+$ (.006)} & .068 \\
4o-J & 2.852(.211) & 2.799(.213) & $+$ (.202) & .015 & 1.420(.338) & 1.117(.196) & \textbf{$+$ ($<$.001)} & .234 & 2.932(.150) & 2.815(.280) & \textbf{$+$ (.008)} & .065 \\
5.4-T & 2.080(.266) & 2.025(.274) & $+$ (.288) & .011 & 1.167(.222) & 1.080(.158) & \textbf{$+$ (.022)} & .049 & 1.852(.280) & 1.469(.255) & \textbf{$+$ ($<$.001)} & .342 \\
5.4-J & 2.043(.243) & 2.000(.259) & $+$ (.373) & .007 & 1.136(.229) & 1.056(.155) & \textbf{$+$ (.035)} & .041 & 1.901(.392) & 1.321(.310) & \textbf{$+$ ($<$.001)} & .407 \\
\bottomrule
\end{longtable}}
}
{\footnotesize\setstretch{1}\raggedright\emph{Note}: Study 2 (DV: selected color saturation); Near/Dist.\ $=$ temporally near vs.\ distant event condition.\par}
\end{landscape}

\begin{landscape}
\subsubsection{\hyperref[e-study-3]{Table E3}. Study 3: Logo shape and product attributes (DV)}\label{e-study-3}

{
{\footnotesize\setlength{\tabcolsep}{2pt}\begin{longtable}[]{@{}lccccccccccccccc@{}}
\toprule
 & \multicolumn{5}{c}{No Intervention} & \multicolumn{5}{c}{Conceptual} & \multicolumn{5}{c}{Empirical} \\
\cmidrule(lr){2-6}\cmidrule(lr){7-11}\cmidrule(lr){12-16}
Config & Circ. $M$(SD) & Ang. $M$(SD) & Dir($p$) & $\eta^2$ & Med. & Circ. $M$(SD) & Ang. $M$(SD) & Dir($p$) & $\eta^2$ & Med. & Circ. $M$(SD) & Ang. $M$(SD) & Dir($p$) & $\eta^2$ & Med. \\
\midrule\endhead
4o-T Comf. & 6.514(.507) & 6.829(.514) & \textbf{$-$ (.012)} & .089 & . & 6.543(.611) & 6.371(.731) & $+$ (.291) & .016 & . & 6.629(.598) & 6.559(.504) & $+$ (.603) & .004 & . \\
4o-T Dur. & 7.600(.736) & 7.343(.838) & $-$ (.177) & .027 & . & 7.657(.482) & 7.514(.562) & $-$ (.258) & .019 & . & 5.029(.296) & 4.971(.460) & $-$ (.534) & .006 & . \\
4o-J Comf. & 5.743(.611) & 5.971(.568) & $-$ (.110) & .037 & . & 6.818(.528) & 6.531(1.218) & $+$ (.220) & .024 & . & 6.429(.608) & 6.235(.431) & $+$ (.133) & .033 & . \\
4o-J Dur. & 7.429(.739) & 7.200(.833) & $-$ (.229) & .021 & . & 7.545(.794) & 7.188(1.355) & $-$ (.197) & .026 & . & 5.000(.594) & 4.912(.514) & $-$ (.512) & .006 & . \\
5.4-T Comf. & 5.314(.471) & 5.000(.642) & \textbf{$+$ (.022)} & .074 & full & 6.571(.698) & 4.686(.631) & \textbf{$+$ ($<$.001)} & .674 & full & 6.857(.355) & 5.114(.676) & \textbf{$+$ ($<$.001)} & .728 & partial \\
5.4-T Dur. & 7.000(.000) & 6.971(.169) & $-$ (.321) & .014 & . & 6.286(.458) & 6.800(.406) & \textbf{$+$ ($<$.001)} & .266 & full & 3.971(.169) & 5.686(.900) & \textbf{$+$ ($<$.001)} & .643 & partial \\
5.4-J Comf. & 5.657(.482) & 5.286(.667) & \textbf{$+$ (.009)} & .095 & . & 6.457(.741) & 4.657(.539) & \textbf{$+$ ($<$.001)} & .665 & full & 6.457(.505) & 5.257(.852) & \textbf{$+$ ($<$.001)} & .430 & full \\
5.4-J Dur. & 7.029(.169) & 7.000(.000) & $-$ (.321) & .014 & . & 6.371(.598) & 6.800(.406) & \textbf{$+$ ($<$.001)} & .153 & partial & 3.686(.471) & 5.429(1.145) & \textbf{$+$ ($<$.001)} & .505 & partial \\
\bottomrule
\end{longtable}}
}
{\footnotesize\setstretch{1}\raggedright\emph{Note}: Study 3; Circ./Ang.\ $=$ circular vs.\ angular logo condition; Comf./Dur.\ $=$ perceived comfort and perceived durability (DVs), reported as separate rows per configuration.\par}
\vspace{2em}

\subsubsection{\hyperref[e-study-4]{Table E4}. Study 4: Logo asymmetry and brand excitement (DV)}\label{e-study-4}

{
{\footnotesize\setlength{\tabcolsep}{2pt}\begin{longtable}[]{@{}lccccccccccccccc@{}}
\toprule
 & \multicolumn{5}{c}{No Intervention} & \multicolumn{5}{c}{Conceptual} & \multicolumn{5}{c}{Empirical} \\
\cmidrule(lr){2-6}\cmidrule(lr){7-11}\cmidrule(lr){12-16}
Config & Asym. $M$(SD) & Symm. $M$(SD) & Dir($p$) & $\eta^2$ & Med. & Asym. $M$(SD) & Symm. $M$(SD) & Dir($p$) & $\eta^2$ & Med. & Asym. $M$(SD) & Symm. $M$(SD) & Dir($p$) & $\eta^2$ & Med. \\
\midrule\endhead
4o-T & 4.836(.514) & 4.859(.511) & $-$ (.736) & .001 & . & 4.731(.579) & 4.687(.645) & $+$ (.595) & .001 & . & 5.027(.370) & 4.901(.552) & \textbf{$+$ (.048)} & .018 & full \\
4o-J & 4.860(.492) & 4.943(.515) & $-$ (.230) & .007 & . & 5.229(.271) & 5.213(.362) & $+$ (.714) & .001 & . & 5.022(.472) & 4.856(.514) & \textbf{$+$ (.014)} & .028 & full \\
5.4-T & 2.978(1.023) & 3.099(1.031) & $-$ (.383) & .003 & . & 3.865(1.192) & 3.899(1.039) & $-$ (.821) & .000 & . & 3.753(1.178) & 3.785(1.125) & $-$ (.833) & .000 & . \\
5.4-J & 3.135(1.091) & 3.218(1.047) & $-$ (.562) & .002 & . & 4.041(1.093) & 4.009(1.082) & $+$ (.826) & .000 & . & 3.996(1.209) & 4.040(1.018) & $-$ (.772) & .000 & . \\
\bottomrule
\end{longtable}}
}
{\footnotesize\setstretch{1}\raggedright\emph{Note}: Study 4 (DV: brand excitement); Asym./Symm.\ $=$ asymmetrical vs.\ symmetrical logo condition.\par}
\end{landscape}

\begin{landscape}
\subsubsection{\hyperref[e-study-5]{Table E5}. Study 5: Color saturation and estimated size (DV)}\label{e-study-5}

{
{\footnotesize\setlength{\tabcolsep}{2pt}\begin{longtable}[]{@{}lccccccccccccccc@{}}
\toprule
 & \multicolumn{5}{c}{No Intervention} & \multicolumn{5}{c}{Conceptual} & \multicolumn{5}{c}{Empirical} \\
\cmidrule(lr){2-6}\cmidrule(lr){7-11}\cmidrule(lr){12-16}
Config & High $M$(SD) & Low $M$(SD) & Dir($p$) & $\eta^2$ & Med. & High $M$(SD) & Low $M$(SD) & Dir($p$) & $\eta^2$ & Med. & High $M$(SD) & Low $M$(SD) & Dir($p$) & $\eta^2$ & Med. \\
\midrule\endhead
4o-T & 15.026(.362) & 14.811(.569) & $+$ (.052) & .050 & . & 15.000(.000) & 14.892(.458) & $+$ (.150) & .028 & . & 15.000(.000) & 14.949(.223) & $+$ (.146) & .027 & . \\
4o-J & 15.179(.854) & 14.844(.515) & $+$ (.055) & .052 & . & 15.051(.320) & 14.972(.167) & $+$ (.189) & .023 & . & 15.000(.000) & 14.974(.160) & $+$ (.314) & .013 & . \\
5.4-T & 14.950(.316) & 14.600(.778) & \textbf{$+$ (.010)} & .082 & . & 15.075(.350) & 14.950(.221) & $+$ (.060) & .045 & . & 15.350(.580) & 15.125(.335) & \textbf{$+$ (.037)} & .055 & . \\
5.4-J & 14.900(.441) & 14.400(.871) & \textbf{$+$ (.002)} & .118 & . & 15.050(.316) & 14.925(.267) & $+$ (.060) & .045 & . & 15.150(.362) & 15.000(.226) & \textbf{$+$ (.029)} & .060 & full \\
\bottomrule
\end{longtable}}
}
{\footnotesize\setstretch{1}\raggedright\emph{Note}: Study 5 (DV: estimated product size); High/Low $=$ high vs.\ low color-saturation condition.\par}
\vspace{2em}

\subsubsection{\hyperref[e-study-6]{Table E6}. Study 6: Logo stability and utility (DV)}\label{e-study-6}

{
{\footnotesize\setlength{\tabcolsep}{2pt}\begin{longtable}[]{@{}lcccccccccccc@{}}
\toprule
 & \multicolumn{4}{c}{No Intervention} & \multicolumn{4}{c}{Conceptual} & \multicolumn{4}{c}{Empirical} \\
\cmidrule(lr){2-5}\cmidrule(lr){6-9}\cmidrule(lr){10-13}
Config & Unst. $M$(SD) & St. $M$(SD) & Dir($p$) & $\eta^2$ & Unst. $M$(SD) & St. $M$(SD) & Dir($p$) & $\eta^2$ & Unst. $M$(SD) & St. $M$(SD) & Dir($p$) & $\eta^2$ \\
\midrule\endhead
4o-T Safety & 8.032(.141) & 8.086(.179) & $-$ (.136) & .028 & 7.799(.206) & 7.744(.214) & $+$ (.264) & .017 & 7.977(.151) & 7.910(.185) & $+$ (.111) & .039 \\
4o-J Safety & 8.136(.089) & 8.160(.148) & $-$ (.401) & .009 & 7.849(.631) & 7.762(.747) & $+$ (.581) & .004 & 7.905(.780) & 7.679(1.459) & $+$ (.443) & .010 \\
5.4-T Safety & 7.151(.407) & 7.168(.364) & $-$ (.844) & .001 & 7.952(.240) & 7.679(.286) & \textbf{$+$ ($<$.001)} & .214 & 7.568(.404) & 7.279(.474) & \textbf{$+$ (.005)} & .100 \\
5.4-J Safety & 7.922(.347) & 7.933(.284) & $-$ (.874) & .000 & 7.098(.351) & 6.810(.524) & \textbf{$+$ (.006)} & .097 & 7.611(.425) & 7.150(.741) & \textbf{$+$ (.001)} & .130 \\
\bottomrule
\end{longtable}}
}
{\footnotesize\setstretch{1}\raggedright\emph{Note}: Study 6; Unst./St.\ $=$ unstable vs.\ stable logo condition. Rows report the logo-stability simple effect within the safety-oriented product category (Safety), based on the three-item utility composite averaged over the six products in the category; the interaction and neutral-category results for the no-intervention runs appear in the main text.\par}
\end{landscape}

\begin{landscape}
\subsubsection{\hyperref[e-study-7]{Table E7}. Case study: Running-shoe logo descriptiveness and brand authenticity (DV)}\label{e-study-7}

{
{\footnotesize\setlength{\tabcolsep}{2pt}\begin{longtable}[]{@{}lcccccccccccc@{}}
\toprule
 & \multicolumn{4}{c}{No Intervention} & \multicolumn{4}{c}{Conceptual} & \multicolumn{4}{c}{Empirical} \\
\cmidrule(lr){2-5}\cmidrule(lr){6-9}\cmidrule(lr){10-13}
Config & More $M$(SD) & Less $M$(SD) & Dir($p$) & $\eta^2$ & More $M$(SD) & Less $M$(SD) & Dir($p$) & $\eta^2$ & More $M$(SD) & Less $M$(SD) & Dir($p$) & $\eta^2$ \\
\midrule\endhead
5.4-T & 4.504(.495) & 4.567(.493) & $-$ (.425) & .004 & 6.404(.366) & 5.650(.533) & \textbf{$+$ ($<$.001)} & .408 & 6.442(.491) & 6.033(.755) & \textbf{$+$ ($<$.001)} & .094 \\
\bottomrule
\end{longtable}}
}
{\footnotesize\setstretch{1}\raggedright\emph{Note}: Case-study deployment (main text, managerial implications): the stimuli are the two AI-generated running-shoe logos (More/Less $=$ more- vs.\ less-descriptive design), not stimuli from a published study. GPT-5.4-mini (TXT) is the configuration validated in the case study; $N = 160$ per intervention column (80 per logo). Because no human baseline exists for these logos, $+$ marks a direction consistent with the Study 1 human benchmark (more descriptive $\rightarrow$ more authentic).\par}
\end{landscape}

\subsubsection{\hyperref[table-e8-tagmap]{Table E8}. Tag definitions and examples for the empirical evidence (statistics) prompt}\label{table-e8-tagmap}

{
{
{\footnotesize\begin{longtable}[]{@{}
  >{\raggedright\arraybackslash}p{(\linewidth - 8\tabcolsep) * \real{0.15}}
  >{\raggedright\arraybackslash}p{(\linewidth - 8\tabcolsep) * \real{0.07}}
  >{\raggedright\arraybackslash}p{(\linewidth - 8\tabcolsep) * \real{0.19}}
  >{\raggedright\arraybackslash}p{(\linewidth - 8\tabcolsep) * \real{0.26}}
  >{\raggedright\arraybackslash}p{(\linewidth - 8\tabcolsep) * \real{0.33}}@{}}
\toprule
Tag & Section & Definition & Additional fields & Example \\
\midrule\endfirsthead
\toprule
Tag & Section & Definition & Additional fields & Example \\
\midrule\endhead
\texttt{label} & both & Section heading & none & ``1. Method'' / ``2. Results'' \\
\texttt{sample} & method & Participant demographics and sample size & none & ``We recruited 180 individuals ($M_{age} = 34$; 46\% female) on MTurk'' (S1) \\
\texttt{design} & method & Experimental design structure & none & ``a 2 (descriptiveness) $\times$ 2 (replicate) between-participant experiment'' (S1) \\
\texttt{measures} & method & Dependent measures and corresponding reliabilities & \texttt{construct}, \texttt{reliability} & \texttt{construct}: authenticity; \texttt{reliability}: $\alpha=.94$ (S1) \\
\texttt{procedure} & method & Experimental task and cover story shown to participants & none & imagine three scenarios, then pick the matching image (S2) \\
\texttt{manipulation} & method & Operationalization of the independent variable & none & stable $=$ square on its base; unstable $=$ diamond (S6) \\
\texttt{analysis} & results & Statistical modeling approach & none & a 2 $\times$ 2 ANOVA with authenticity as the dependent variable (S1) \\
\texttt{effects} & results & Focal statistical findings & \texttt{dv}, \texttt{iv}, \texttt{comparison}, \texttt{statistic}, \texttt{p}, \texttt{effect\_size} & \texttt{dv}: size estimate; \texttt{iv}: color saturation; \texttt{comparison}: high ($M=15.28$) vs.\ low ($M=14.41$); \texttt{statistic}: $F(1,75)=7.13$; \texttt{p}: $<.01$; \texttt{effect\_size}: $\eta^{2}=.09$ (S5) \\
\texttt{mediation} & results & Mediation analysis results & \texttt{model}, \texttt{mediator}, \texttt{iv}, \texttt{dv}, \texttt{path\_a}, \texttt{path\_b}, \texttt{indirect}, \texttt{direct} & \texttt{model}: PROCESS Model 4 \citep{hayes2022}; \texttt{mediator}: arousal; \texttt{iv}: logo asymmetry; \texttt{dv}: excitement; \texttt{path\_a}: $b=.41$; \texttt{path\_b}: $b=.72$; \texttt{indirect}: 95\% CI [.07, .54]; \texttt{direct}: $b=.33$ (S4) \\
\texttt{auxiliary} & both & Supporting analyses & none & ``six participants were excluded for failing the attention check'' (S4) \\
\bottomrule
\end{longtable}}
{\footnotesize\setstretch{1}\raggedright\emph{Note}: The JSON schema comprises two primary sections: \texttt{method} and \texttt{results}. Every tag preserves its source sentence verbatim within a base \texttt{text} field. The ``Additional fields'' column denotes nested key-value pairs used to extract specific parameters (e.g., statistics, variable names) from that sentence. ``None'' indicates the tag retains only the base \texttt{text} field (e.g., \texttt{auxiliary} lacks nested fields due to high structural variance across studies). The \texttt{construct} field under \texttt{measures} defines the latent variable and its reliability, whereas the \texttt{dv} field under \texttt{effects} references that measure during hypothesis testing. If a single source sentence satisfies multiple tags (e.g., simultaneously reporting sample size and design), it is assigned to the primary tag, with the secondary tag pointing to it via reference to prevent data duplication. S1 (Study 1) through S6 (Study 6) denote the specific study from which the example is drawn.\par}
}
}

\clearpage

\clearpage
\section{Web Appendix \texorpdfstring{F}{F}: Robustness of findings}\label{web-appendix-f-robustness-of-findings}

The robustness checks below re-run the Study 1 replication (logo descriptiveness $\rightarrow$ brand authenticity) with GPT-4o-mini (TXT), the configuration reported in the main text; \hyperref[table-f1-test-retest-reliability-for-5-different-runs-temperature-1]{Tables F1} and F3 were run at temperature $=$ 1. MC/ME $=$ manipulation check (perceived descriptiveness)/main effect (authenticity). Dir($p$) and $\eta^2$ follow the conventions of \hyperref[web-appendix-e-prompt-intervention-templates-and-results-after-interventions]{Web Appendix E}; sample sizes match the original study's (\hyperref[web-appendix-b-description-of-studies-and-visual-stimuli]{Web Appendix B}).

\subsubsection{\hyperref[table-f1-test-retest-reliability-for-5-different-runs-temperature-1]{Table F1}. Test-retest reliability for 5 different runs}\label{table-f1-test-retest-reliability-for-5-different-runs-temperature-1}

{
{\footnotesize\setlength{\tabcolsep}{3pt}\begin{longtable}[]{@{}ccccccccc@{}}
\toprule
 & \multicolumn{4}{c}{Main Effect} & \multicolumn{4}{c}{Manipulation Check} \\
\cmidrule(lr){2-5}\cmidrule(lr){6-9}
Run & More $M$(SD) & Less $M$(SD) & Dir($p$) & $\eta^2$ & More $M$(SD) & Less $M$(SD) & Dir($p$) & $\eta^2$ \\
\midrule\endhead
\bottomrule\endlastfoot
1 & 6.424(.369) & 6.211(.417) & \textbf{$+$ (.001)} & .068 & 7.532(.502) & 7.085(.422) & \textbf{$+$ ($<$.001)} & .191 \\
2 & 6.424(.354) & 6.198(.405) & \textbf{$+$ ($<$.001)} & .082 & 7.553(.500) & 7.131(.485) & \textbf{$+$ ($<$.001)} & .156 \\
3 & 6.464(.316) & 6.263(.434) & \textbf{$+$ (.001)} & .065 & 7.568(.499) & 7.173(.495) & \textbf{$+$ ($<$.001)} & .138 \\
4 & 6.457(.309) & 6.267(.414) & \textbf{$+$ (.001)} & .064 & 7.513(.503) & 7.106(.464) & \textbf{$+$ ($<$.001)} & .152 \\
5 & 6.427(.355) & 6.294(.356) & \textbf{$+$ (.015)} & .035 & 7.565(.522) & 7.083(.445) & \textbf{$+$ ($<$.001)} & .199 \\
\end{longtable}}
}
{\footnotesize\setstretch{1}\raggedright\emph{Note}: Five independent reruns of the Study 1 replication under identical settings (GPT-4o-mini (TXT), temperature $=$ 1, 512-pixel stimuli); the main text reports a single run from this set.\par}

\subsubsection{\hyperref[table-f2-different-temperature-settings-in-gpt]{Table F2}. Different temperature settings}\label{table-f2-different-temperature-settings-in-gpt}

{
{\footnotesize\setlength{\tabcolsep}{3pt}\begin{longtable}[]{@{}ccccccccc@{}}
\toprule
 & \multicolumn{4}{c}{Main Effect} & \multicolumn{4}{c}{Manipulation Check} \\
\cmidrule(lr){2-5}\cmidrule(lr){6-9}
Temperature & More $M$(SD) & Less $M$(SD) & Dir($p$) & $\eta^2$ & More $M$(SD) & Less $M$(SD) & Dir($p$) & $\eta^2$ \\
\midrule\endhead
\bottomrule\endlastfoot
.5 & 6.422(.291) & 6.238(.454) & \textbf{$+$ (.002)} & .056 & 7.547(.501) & 7.046(.504) & \textbf{$+$ ($<$.001)} & .201 \\
1.0 & 6.424(.354) & 6.198(.405) & \textbf{$+$ ($<$.001)} & .082 & 7.553(.500) & 7.131(.485) & \textbf{$+$ ($<$.001)} & .156 \\
1.5 & 6.468(.286) & 6.263(.426) & \textbf{$+$ (.001)} & .075 & 7.486(.602) & 7.013(.554) & \textbf{$+$ ($<$.001)} & .145 \\
\end{longtable}}
}
{\footnotesize\setstretch{1}\raggedright\emph{Note}: The Study 1 replication re-run at three temperature settings (GPT-4o-mini (TXT), 512-pixel stimuli); 1.0 is the API default used in the main text.\par}

\subsubsection{\hyperref[table-f3-resolution-of-the-visual-stimuli-256-512-1024-temperature-1]{Table F3}. Resolution of the visual stimuli: 256, 512, 1024}\label{table-f3-resolution-of-the-visual-stimuli-256-512-1024-temperature-1}

{
{\footnotesize\setlength{\tabcolsep}{3pt}\begin{longtable}[]{@{}ccccccccc@{}}
\toprule
 & \multicolumn{4}{c}{Main Effect} & \multicolumn{4}{c}{Manipulation Check} \\
\cmidrule(lr){2-5}\cmidrule(lr){6-9}
Pixel & More $M$(SD) & Less $M$(SD) & Dir($p$) & $\eta^2$ & More $M$(SD) & Less $M$(SD) & Dir($p$) & $\eta^2$ \\
\midrule\endhead
\bottomrule\endlastfoot
256 & 6.461(.326) & 6.228(.422) & \textbf{$+$ ($<$.001)} & .088 & 7.573(.498) & 7.105(.450) & \textbf{$+$ ($<$.001)} & .198 \\
512 & 6.424(.354) & 6.198(.405) & \textbf{$+$ ($<$.001)} & .082 & 7.553(.500) & 7.131(.485) & \textbf{$+$ ($<$.001)} & .156 \\
1024 & 6.528(.209) & 6.318(.396) & \textbf{$+$ ($<$.001)} & .099 & 7.646(.481) & 7.267(.518) & \textbf{$+$ ($<$.001)} & .127 \\
\end{longtable}}
}
{\footnotesize\setstretch{1}\raggedright\emph{Note}: The Study 1 replication re-run at three stimulus resolutions (GPT-4o-mini (TXT), temperature $=$ 1); 512 pixels is the standard used in the main text.\par}

\subsubsection{\hyperref[table-f4-carryover]{Table F4}. Testing the carryover of ratings among consecutive API calls (Study 1: 4o-mini (TXT))}\label{table-f4-carryover}

{
{\footnotesize\setlength\LTleft{\fill}\setlength\LTright{\fill}\setlength{\tabcolsep}{6pt}\begin{longtable}[]{@{}lcccc@{}}
\toprule
 & Authenticity & Trustworthiness & Credibility & Descriptiveness \\
\midrule\endfirsthead
\toprule
 & Authenticity & Trustworthiness & Credibility & Descriptiveness \\
\midrule\endhead
More descriptive condition &  &  &  &  \\
\quad $\beta$ & .082 & .114 & .379 & .430 \\
\quad Standard error & .025 & .016 & .026 & .016 \\
\quad \emph{p}-value & .030 & .002 & $<$.001 & $<$.001 \\
Carryover &  &  &  &  \\
\quad $\lambda$ & .016 & .019 & $-$.000 & .007 \\
\quad Standard error & .050 & .040 & .029 & .040 \\
\quad \emph{p}-value & .763 & .659 & .988 & .866 \\
Replicate Fixed Effect & Yes & Yes & Yes & Yes \\
N & 814 & 814 & 814 & 814 \\
R-squared & .016 & .025 & .101 & .168 \\
\bottomrule
\end{longtable}}
{\footnotesize\setstretch{1}\raggedright\emph{Note}: OLS estimates for Study 1, 4o-mini (TXT), pooled across the five experiment replicates (N = 814). Each column regresses the rating $Y_{ri}$ of participant $i$ in replicate $r$ on the binary condition indicator ($\beta$; more vs.\ less descriptive logo), the preceding participant's rating on the same measure ($\lambda$, the carryover coefficient), and replicate fixed effects. Standard errors are clustered at the replicate level. A $\lambda$ indistinguishable from zero indicates no carryover between consecutive API calls.\par}
}

\end{document}